\documentclass{article}
\usepackage{xcolor}

 \usepackage[preprint]{neurips_2026}

\usepackage[utf8]{inputenc} % allow utf-8 input
\usepackage[T1]{fontenc}    % use 8-bit T1 fonts
\usepackage{hyperref}       % hyperlinks
\usepackage{url}            % simple URL typesetting
\usepackage{booktabs}       % professional-quality tables
\usepackage{amsfonts}       % blackboard math symbols
\usepackage{nicefrac}       % compact symbols for 1/2, etc.
\usepackage{microtype}      % microtypography
\usepackage{xcolor}         % colors
\usepackage{cleveref}

\usepackage{graphicx}
\usepackage{subcaption}
\usepackage{wrapfig}
\usepackage[inline]{enumitem}

\title{Physically Viable World Models: A Case for Query-Conditioned Embodied AI}

\author{%
\textbf{Adam J. Thorpe}\thanks{Equal contribution.} \quad
\textbf{Stepan Tretiakov}\footnotemark[1] \quad
\textbf{Cheng-Hsi Hsiao} \quad
\textbf{Su Ann Low} \quad
\textbf{Xingjian Li} \\
\textbf{Hassan Iqbal} \quad
\textbf{Neel P. Bhatt} \quad
\textbf{Ufuk Topcu} \quad
\textbf{Krishna Kumar} \\
\normalfont The University of Texas at Austin \\
\normalfont\texttt{adam.thorpe@austin.utexas.edu} \quad
\normalfont\texttt{stepan@utexas.edu} \quad
\normalfont\texttt{chhsiao@utexas.edu} \\
\normalfont\texttt{suann@utexas.edu} \quad
\normalfont\texttt{xingjian.li@austin.utexas.edu} \quad
\normalfont\texttt{hassan.iqbal@utexas.edu} \\
\normalfont\texttt{npbhatt@utexas.edu} \quad
\normalfont\texttt{utopcu@utexas.edu} \quad
\normalfont\texttt{krishnak@utexas.edu}
}

\begin{document}

\maketitle

\begin{abstract}
World models for embodied AI must be physically viable: constructed to answer intervention queries by representing the underlying physical structure that governs the outcomes of an agent's action, rather than merely serving as generic predictors of future observations. Existing world models, trained to predict observations, can produce visually plausible rollouts that are physically incorrect. This failure is structural; different physical systems can produce identical observations yet behave differently under intervention. We expose this problem through controlled benchmarks that keep the visible scene fixed while varying the underlying physics. We demonstrate that observation-predictive models may recommend infeasible actions, mispredict interaction outcomes, or certify behaviors that would otherwise be unsafe in the real world. We argue that embodied AI requires world models that identify the simplest physical abstraction sufficient to answer a given intervention query. Such a physically viable world model is composed from modular components, including environment representation, latent state and parameter estimation, action specification, dynamics under intervention, and response to queries. An autonomous orchestrator should identify the relevant abstraction and compose the world model from compatible learned and structured components for each query. The transition model may be analytic, simulated, learned, or hybrid when closed-form physics is unavailable, uncertain, or computationally expensive, but it must preserve the physical structure that determines the outcome of interventions. The resulting modular decomposition makes the model interpretable, its components verifiable, and its outputs auditable against the query. It also provides both a design principle for new world models and a feasibility test for existing ones: the right abstraction is not the most detailed model of the world, but the simplest model that preserves the distinctions relevant to the query. We demonstrate how this approach could work in practice on intervention queries that existing systems fail to answer correctly, and outline how an orchestrator can dynamically assemble and adapt physically viable models for planning, control, and verification.
\end{abstract}

% There are some principled steps we can take. 
% We show an instantiation toward some of these goals.

\begin{figure}[!th]
    \centering
    \includegraphics[width=\linewidth, keepaspectratio]{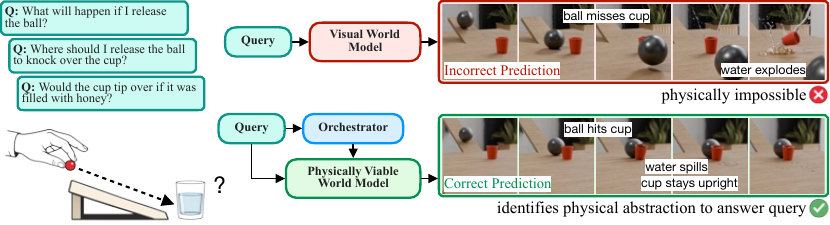}
    \caption{Visual world models can produce visually plausible but physically impossible predictions. We argue that embodied AI therefore requires world models that identify the simplest physical abstraction sufficient to answer a given intervention query.}
    \label{fig: title}
\end{figure}

\section{Introduction}

A world model for embodied AI is physically viable when it supports correct reasoning about how a physical system evolves under intervention. For embodied AI, the relevant future is not just a sequence of observations, but the evolution of a physical system under intervention. A model used for planning, control, counterfactual reasoning, or safety analysis must therefore preserve the physical distinctions that affect those decisions. Visual plausibility does not guarantee this property. A rollout may look realistic while using invalid dynamics or omitting variables that the action depends on. Embodied world modeling therefore requires query-conditioned physically viable world models: models whose state variables, dynamics, parameters, and constraints are sufficient for the intervention being evaluated. Scaling observation prediction alone does not guarantee this requirement, especially when the variables that determine the outcome are latent, unobserved, or revealed only through action.
As illustrated in Figure~\ref{fig: title}, the same intervention query may require reasoning over contact, mass, fluid response, or stability constraints that are not identifiable from visual appearance alone.

A query specifies the intervention, the task, and the standard of correctness the prediction must satisfy. These elements determine the abstraction the model must construct: which variables, dynamics, parameters, constraints, and level of fidelity are required to answer the query. Physical viability does not require the most detailed model of the world. It requires the simplest abstraction that preserves the distinctions relevant to the query. For example, a grasping task may require contact and friction, a pouring task may require volume transfer and conservation laws, and a safety query may require reachability or barrier constraints without requiring photorealistic rendering. The right world model is therefore not the most realistic model in general, but the model whose variables, equations, parameters, and constraints are sufficient for the intervention being considered.

Current observation-predictive world models fail when their training signal does not identify the dynamics needed for action. Vision-language models, video generators, and latent predictive models can match perceptual regularities while failing on the latent physical variables that determine intervention outcomes~\cite{chow2025physbench,kang2024far,meng2024towards,guo2025t2vphysbench,gu2025phyworldbench,motamed2026generative,zhang2025morpheus,zhang2026physion}. The issue is structural rather than architectural. The same observations can fit multiple physical systems that behave differently when acted on. More passive data may improve visual realism and short-horizon prediction without resolving the distinctions that determine the intervention outcome.

\textbf{We argue that world modeling for embodied AI must shift from observation extrapolation to query-conditioned construction of physically viable models.} This position does not require maximal physical detail, but explicit selection of the abstraction needed to answer the query. A physically viable model must represent the variables the intervention acts on, use compatible dynamics and constraints, and return the form of answer the query requires. This view separates roles that end-to-end predictors often conflate, including perception, abstraction, parameter estimation, dynamics, and query-level response. World models should therefore be judged not strictly by perceptual realism, but by whether their abstractions support downstream decisions under intervention. We support this position with controlled examples that expose the failure modes of visual world models and with illustrative constructions that show how query-conditioned physical abstractions could address them.

\textbf{Contributions:}
This paper makes three contributions.
\begin{enumerate*}[label=(\arabic*)]
    \item 
    We argue observation-predictive world models are structurally inadequate for embodied intervention, and embodied world models should be constructed per query, around the physical distinctions the intervention depends on.
    \item 
    % We support this positio Xn with controlled examples that expose the failure modes of visual world models under controlled physical variation.
    We demonstrate the resulting failures in three model families (vision-language, video diffusion, and action-conditioned latent prediction) under controlled physical variation.
    \item
    We define a modular design framework for query-conditioned construction of a physically viable world model, and specify the operations an orchestrator must perform to compose it.
    % We propose a modular design framework for physically viable world modeling, separating perception, abstraction, action specification, parameter estimation, dynamics, constraints, and query-level response, and illustrate it through controlled embodied examples.
\end{enumerate*}

\section{How Current World Models Fail to Represent Physics}

Current world models are often optimized to predict future observations from past observations, producing visually coherent predictions that fail under intervention. We use the simulations as an evaluation suite: static and counterfactual VLM prediction, diffusion video continuation, and action-conditioned latent control. Across these tests, appearance or action is held nearly fixed while latent physics varies; models often give plausible explanations, videos, or actions without preserving the variables that determine the outcome. Full protocols and results are in Appendix~\ref{subsec:vlm_details}, Appendix~\ref{subsec:diffusion_results}, and Appendix~\ref{subsec:vjepa_control_results}.

\begin{figure}[!t]
    \centering
    \includegraphics[width=\linewidth]{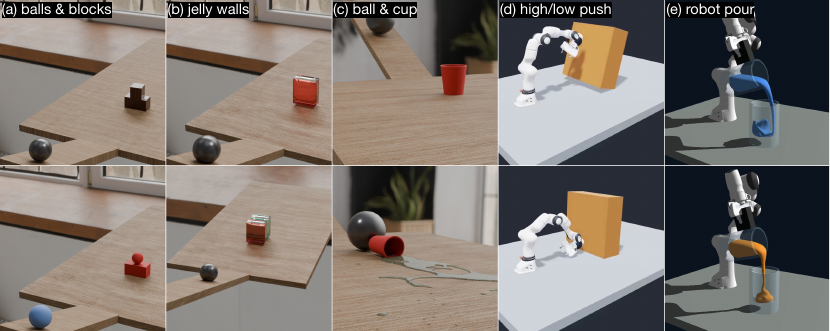}
    \caption{
    Controlled evaluation scenes used to expose failures of visual world models under latent physical variation, including rigid-body collision, deformable interaction, rigid--fluid coupling, contact-dependent pushing, and viscosity-dependent pouring.
    }
    \label{fig: simulation suite}
\end{figure}

\subsection{Simulation suite and tests}
\label{sec:simulation-suite}

We evaluate models on controlled simulations with specified physical parameters, interventions, and reference rollouts (Figure~\ref{fig: simulation suite}). Each scene family isolates a query-relevant latent variable while holding appearance or action nearly fixed. The suite supports the three tests summarized above and detailed in Appendix~\ref{subsec:vlm_details}, Appendix~\ref{subsec:diffusion_results}, and Appendix~\ref{subsec:vjepa_control_results}.

\textbf{Ramp-to-tower rigid-body interactions.}
These scenes test whether a model tracks latent rigid-body properties rather than predicting a generic collision outcome from visual appearance alone. A ball rolls down a ramp and collides with a small block tower. We use two closely related setups. In the density-variation setup, the tower geometry remains fixed while the block material changes, for example from wood to steel, so that the same apparent collision produces different momentum transfer and collapse behavior. In the restitution setup, we introduce high-restitution objects, including a bouncy projectile and composite towers that may contain a bouncy element above rigid lower blocks. These variants are shown in \Cref{fig:ramp_tower_variants}. They probe whether predictions account for mass, inertia, restitution, frictional dissipation, and contact ordering.

\textbf{Deformable-jelly-wall interactions.}
We replace the rigid block tower with one or two deformable jelly walls and vary the release distance. Visually similar collisions can produce deformation, sliding, tipping, or momentum transfer through multiple deformable bodies depending on impact energy and material response. Correct prediction therefore requires tracking deformation, energy dissipation, and sequential contact dynamics. These scenes test whether predictions capture compliance and sequential deformable-body interaction. The static VLM benchmark includes the single-jelly-wall and two-jelly-wall settings in \Cref{fig: vlm tests}, while \Cref{fig:diffusion_jelly_2} compares the physical rollout against a diffusion-generated continuation for the two-jelly-wall case.

\textbf{Ramp-to-liquid-filled-cup impact.}
These scenes test whether models capture coupled rigid--fluid interaction. A ball strikes a partially or fully liquid-filled cup while geometry and camera position remain fixed. We vary the ball material, release height, and fill level across trials. The outcome depends on ball momentum, cup motion, liquid inertia, sloshing, spillage, and the combined cup--fluid center of mass. The liquid-filled-cup setting appears in the static VLM benchmark in \Cref{fig: vlm tests}; \Cref{fig:diffusion_cup} further compares the physically simulated rigid--fluid interaction against a diffusion-generated video continuation.

\textbf{Robot-wall pushing.}
A Franka Panda end-effector follows the same horizontal pushing trajectory into a freestanding wall while contact height or floor friction is varied. High and low contact points separate tipping torque from translation, while friction variation separates sliding from overturning. This tests whether action prediction is conditioned on the physical contact regime. Representative high-push, low-push, and material-variation rollouts are shown in \Cref{fig:vjepa_wall_comparison,fig:vjepa_low_push_comparison}, where visually selected trajectories are compared against simulator-grounded executions.

\textbf{Robot-arm pouring and viscosity variants.}
These scenes test whether models infer latent fluid properties that alter the correct action. A robot pours from one glass into another under fixed geometry and controlled motion. We compare water-like, honey-like, and synthetic-viscosity liquids. Viscosity changes flow rate, transfer timing, retained liquid, and spill behavior, so the correct action may require a different hold duration or parameter-identification step. Candidate viscosities are evaluated against query-relevant quantities such as receiver fill, residual liquid, and spillage. The viscosity-dependent pouring variants and best-match parameter-estimation result are shown in \Cref{fig:appendix_pouring_viscosity_variants}.

Together, these scenes cover rigid impact, restitution, deformable interaction, rigid--fluid coupling, contact-rich pushing, and viscosity-dependent pouring. They instantiate the VLM, diffusion, latent-control, and viscosity-estimation tests reported in the appendix. The simulation code is available at \url{https://github.com/pvwm/physically-viable-world-models}, and supplementary videos of simulator rollouts, diffusion-generated video continuations, and V-JEPA action-conditioned rollouts are available on the project website: \url{https://pvwm.github.io/}.

\subsection{Why the failures are structural}
\label{subsec:structural_failures}

The failures above are not specific to one architecture. The tests in \Cref{subsec:vlm_details,subsec:diffusion_results,subsec:vjepa_control_results} show the same pattern in different forms: VLMs identify relevant effects but miss thresholded outcomes; diffusion rollouts remain visually coherent while violating contact, fluid, or deformable dynamics; and latent-control plans can be visually plausible but physically infeasible. VLMs mediate physical knowledge through image-text priors and token-level reasoning, without an explicit state that evolves under action. Video diffusion models represent state as image sequences, so physical variables exist only insofar as they are recoverable from pixel statistics. Latent world models learn transition functions in representations optimized for prediction~\cite{ha2018world,ha2018recurrent,hafner2019learning,hafner2019dream,hafner2020dreamerv2,chen2022transdreamer,hafner2023mastering,deng2023s4wm,hafner2025mastering,schrittwieser2020muzero,micheli2022transformers,hansen2023tdmpc2}, but these latents need not correspond to physical state variables. 
% The shared property is that physical dynamics are inferred rather than enforced.

This limitation follows from the training objective. Prediction over observation sequences asks the model to infer dynamics from observed data, but this inverse problem is not unique. Distinct physical systems can induce identical or nearly identical observations when the variables that determine evolution, such as mass, friction, compliance, inertia, restitution, viscosity, or contact state, are latent or indirectly observed. This is not a claim that dynamics can never be identified. With full state access, rich interventions, targeted system-identification data, or strong physical priors, the relevant system may be recoverable. The issue is that current training regimes often lack this information and optimize for observed outcomes rather than recovering the variables needed under intervention.

Our tests make the ambiguity explicit: appearance and actions stay nearly fixed while the correct response changes with density, restitution, deformability, contact height, friction, rigid--fluid coupling, or viscosity. Without representing, estimating, or probing these quantities, models follow visual similarity rather than intervention-relevant dynamics. Large-scale studies show the same pattern: in controlled mechanics environments, out-of-distribution error can remain dominated by visual similarity rather than dynamical variables~\cite{kang2024far}, and video generation models can violate physical laws despite large-scale training and diverse evaluation settings~\cite{meng2024towards,guo2025t2vphysbench,gu2025phyworldbench}.

Action-conditioned models reduce ambiguity only over the distribution of actions and regimes they observe. They can still fail when extrapolating to unseen contact modes, material properties, or control settings~\cite{hafner2019learning,hafner2025mastering,gupta2022maskvit,janner2022planning,ding2024diffusion,rigter2024avid,huang2025vid2world}. Likewise, existing approaches mitigate the problem without eliminating it. Physics-informed losses and architectural priors constrain predictions but are often soft regularizers~\cite{raissi2019physics,greydanus2019hnn,cranmer2020lnn,cranmer2020discovering,beucler2021enforcing}; object-centric models recover useful structure but still learn interaction dynamics from data~\cite{kipf2020,locatello2020,liu2023ifactor}; scientific surrogates and learned simulators can be accurate within fixed regimes but typically assume predefined variables or governing dynamics~\cite{fourcastnet,graphcast,panguweather,chang2016compositional,battaglia2016interaction,li2018learning,chen2018neural,sanchezgonzalez2020learning,pfaff2020learning}; and differentiable or hybrid simulators shift difficulty to representation, parameter estimation, and model selection~\cite{difftaichi,jatavallabhula2021gradsim,brax,physgaussian,abouchakra2024,zhang2024physdreamer,huang2024dreamphysics,liu2024physflow,chen2025physgen3d,zhou2025chronodreamer}. These tools are valuable, but they must be composed around the intervention query.

The consequence appears under action. Two systems that look the same over the training distribution can respond differently when intervened upon. Reliable planning, control, and counterfactual reasoning therefore require physical structure to enter as represented, estimated, or enforced components, rather than only as an emergent property of observation prediction.

\section{Physically viable world modeling}
% embodied AI should use query-conditioned, physically viable world models that construct the simplest physical abstraction sufficient for a specific intervention query, rather than treating world models as universal observation predictors. 

We now describe a constructive view of physically viable world models. The central idea is that an embodied agent should begin with an intervention query, determine which physical distinctions the query depends on, and construct the simplest abstraction sufficient to preserve those distinctions. We separate this process into two parts: an orchestrator that selects the relevant abstractions and compatible components for the query, and a world model that composes those choices into a model of the system under intervention.
Figure~\ref{fig:pipeline} gives a representative construction for counterfactual, planning, verification, and parameter-identification queries. We then use three examples—a ramp-and-cup interaction, robot pouring with latent viscosity, and flooded-road traversal—to show how the query determines variable selection, parameter estimation, action selection, dynamics, and constraints.

\subsection{Constructing the simplest abstraction sufficient for a query}

\begin{figure}[!t]
    \centering
    \includegraphics[width=\linewidth]{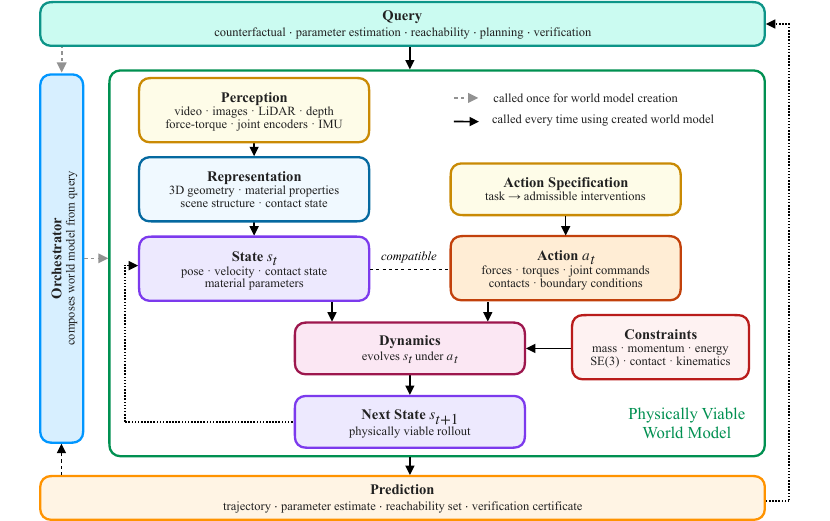}
    \caption{Representative construction of a physically viable world model. A user-specified intervention query determines the physical abstraction to be constructed. Perception and representation recover query-relevant scene variables, action specification defines admissible interventions, dynamics and constraints evolve the selected state under action, and prediction returns the response required by the query. The compatibility relation emphasizes that the selected representation, action interface, dynamics, constraints, and output must support the same abstraction.}
    \label{fig:pipeline}
\end{figure}

\textbf{Intervention queries.} A physically viable world model is constructed around the physical requirements determined by an intervention query. In embodied settings, the query is a request about interaction with the world, such as what will happen if a particular action is taken, which action will produce a desired outcome, or whether an action remains safe under uncertainty. 
% Such questions are central to control, planning, and verification because they concern the consequences of an action, not the likelihood of future visual observations. 
% 
The query defines the modeling problem because it specifies the intervention being considered, the outcome that matters, the physical uncertainty that must be resolved, and the form of the answer needed for decision-making. 
These elements determine the abstraction---the variables to represent, the parameters to estimate or bound, the actions to admit, the dynamics to use, the constraints to enforce, and the response the model should return.
A model that answers the query does not need to represent every physical detail of the scene. 
This query-conditioned construction is a design principle rather than a fixed architecture. The appropriate abstraction is the one that preserves the physical distinctions that can change the answer. 
If the answer depends on mass, friction, contact geometry, fill level, or viscosity, then those quantities must be represented, estimated, or bounded.
A richer model may be useful when it improves reuse, robustness, or safety margins, but additional detail also creates estimation and validation obligations.

\textbf{Variable selection.} 
Variable selection determines the state, latent parameters, and admissible actions required to answer the query. In Figure~\ref{fig:pipeline}, perception provides evidence from available sensors, representation maps that evidence into physical variables, and action specification defines interventions in the same variable space. These components must be compatible: an action is only meaningful if it acts on variables represented in the state, and the state is only useful if it exposes the quantities the action changes. For example, a force applied at a contact point requires a representation containing geometry, pose, contact location, mass, and velocity, rather than raw pixel observations alone. Different queries therefore require different abstractions. Navigation may require free space and robot pose, grasping may require contact geometry and friction, and pouring may require fill level, container geometry, and fluid properties. Variable selection must also account for latent quantities that cannot be identified from passive observation alone, including mass, friction, viscosity, compliance, or contact state. When such quantities affect the intervention outcome, the model should estimate them, maintain uncertainty over possible values, gather additional information, or return a conditional response rather than collapsing unresolved uncertainty into a single unsupported prediction.

\textbf{State evolution and constraints.} After the state and action variables are selected, the model must specify how the state evolves under the admissible intervention. This corresponds to the dynamics, constraints, and next-state blocks in Figure~\ref{fig:pipeline}. 
A simple query may only require a closed-form relation, such as a kinematic equation, conservation law, or stability condition. A query involving contact, liquid transfer, deformation, or coupled rigid-body motion may require a numerical solver. A learned simulator or constrained surrogate may be appropriate when analytic equations are unavailable, numerical solvers are too expensive, or the relevant physical process is only partially specified. A composite model may combine these pieces, for example by using learned perception to recover geometry, an analytic calculation for a stability margin, a numerical solver for contact, and a verifier for safety.
The goal is to select dynamics with the fidelity required by the query, not to build a universal simulator of the scene. 

% The transition model can therefore take many forms. It may be analytic, numerical, learned, or hybrid. Analytic models are useful when the query depends on a small number of known physical relations. Numerical solvers are useful when the query requires explicit contact, fluid, deformation, or multi-body evolution. Learned simulators or constrained surrogate models are useful when analytic equations are unavailable, numerical solvers are too expensive, or the relevant physical process is only partially specified. Composite models can combine these pieces, for example by using perception to estimate geometry, an analytic model to compute a stability margin, a learned surrogate to approximate local fluid response, and a verifier to check safety constraints.

The important requirement is that the selected dynamics preserve the physical structure that determines the query response. A learned simulator may be physically viable if it respects the relevant regimes, invariants, and constraints. A numerical simulator may fail if it uses the wrong physical regime or omits the parameter that the query depends on. A reduced analytic model may be preferable when it captures the decisive quantity without introducing unnecessary state. 

\textbf{Query-level response.} The output of a physically viable world model is the query-level response required for decision-making. This corresponds to the prediction block in Figure~\ref{fig:pipeline}. 
For example, the answer may be a trajectory, a parameter estimate, a feasible action set, a reachability set, a stability boundary, or a verification certificate. The model is judged by whether its response preserves the physical distinctions the query depends on. A visually plausible rollout can still fail if it omits the variable that determines the intervention outcome.

\textbf{Orchestrator.} This construction is useful only if the abstraction and components can be selected automatically, which remains the central open problem. An early orchestrator does not need to solve open-ended physical reasoning. It can instead operate conservatively by starting from a library of physical abstractions, selecting candidate variables from the query and scene context, testing whether latent parameters are identifiable from available evidence, routing between analytic, numerical, learned, or hybrid models, and returning uncertainty when the construction is underdetermined. Existing tools already provide partial mechanisms for this role, including model selection in system identification, program synthesis, differentiable simulation, tool-using language models, model predictive control, active information gathering, and verification over constrained dynamics.

\textit{The orchestrator should therefore be understood not as a fixed module, but as an adaptive, closed-loop selection and checking process that routes between analytic models, numerical simulators, learned surrogates, estimators, controllers, and verification tools while revising the construction when the query changes, new evidence arrives, or a selected component fails a compatibility check.} It may also determine that the available information is insufficient and that additional sensing or interaction is required before the query can be answered. Its role is to construct the simplest compatible world model for the query and ensure that the composed model supports the required reasoning under intervention. 
The orchestrator does not need to be a single learned module, but it does require the selection problem itself to be explicit. A physically viable world model should therefore not only produce an answer to the query, but also expose why the selected abstraction is adequate, which assumptions it depends on, which latent quantities remain uncertain, and which constraints were checked, making the model useful both for planning and control and for diagnosing when the query cannot be answered from the available information.

\subsection{Illustrative demonstrations}

In this section, we support our position with constructions that show how query-conditioned physical abstractions could address the failure modes we have identified.

\begin{figure}
    \centering
    \includegraphics[width=\linewidth]{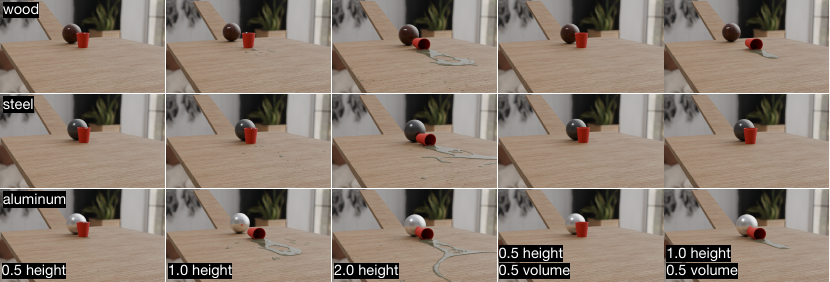}
    \caption{Counterfactual evaluation of a ramp-to-cup intervention query under varying release heights, ball materials, and fill conditions.}
    \label{fig: ball sims}
\end{figure}

\textbf{Placing a ball on a ramp:}
The first demonstration considers a ball released on a ramp toward a fluid-filled cup. The intervention query asks where the ball should be released so that the cup tips over, or where it should be released so that the cup remains upright. The query-level response is the set of release conditions that produce tipping or non-tipping behavior.

Figure~\ref{fig: ball sims} shows world model evaluations of this query under different release heights, ball materials, and fill conditions. Each rollout keeps the scene geometry nearly fixed while changing physical parameters that affect the outcome. The same visible setup can tip or remain upright depending on the ball mass, restitution, release height, fluid volume, and fluid response. These parameters shift the release conditions separating tipping from non-tipping outcomes.
The important point is that these differences are physical rather than visual. A steel ball, a wooden ball, and an aluminum ball can look nearly identical while transferring different momentum at impact. Changing the release height changes the impact speed, while changing the fill level changes the cup–fluid center of mass and the stability margin. These quantities determine the outcome of the interaction, but they may not be recoverable from visual appearance alone.

This example illustrates the central claim of the paper. The intervention outcome depends on latent physical quantities that observation-predictive models can miss when visually similar scenes behave differently under action. For this query, the physically viable abstraction is the minimal set of variables, parameters, and dynamics needed to determine whether the cup tips under intervention. A physically viable world model must therefore represent, estimate, or condition on these quantities rather than rely only on visual similarity.

\textbf{Pouring fluid into a jar:}
% \label{subsubsec:pouring_fluid_jar}
The second demonstration considers a robot arm pouring liquid from a source glass into a receiving jar. The intervention query is not whether a rollout visually resembles pouring, but which robot motion transfers a target volume, such as half a glass, without underfilling, overshooting, or spilling. The query-level response is therefore an action, policy, or termination condition rather than a rendered video.

Let $u_{0:T}$ denote the robot motion over horizon $T$, including tilt trajectory, angular velocity, hold duration, and stopping rule. We consider three related queries. First, a planning query asks for a motion such that the received volume reaches a target $V^\star$ within tolerance,
$V_{\mathrm{recv}}(T)\in[V^\star-\epsilon,V^\star+\epsilon]$, while the spilled volume satisfies $V_{\mathrm{spill}}(T)\leq\delta$. For a water-like liquid, a nominal tilt and hold duration may be sufficient. The required abstraction includes container geometry, robot pose, initial fill, liquid volume, gravity, and the fluid parameters governing transfer rate. Second, a counterfactual action query asks how the motion should change when liquid properties change while geometry and target remain fixed. In our longer-pour experiment, replacing a water-like liquid with a honey-like, high-viscosity liquid changes the correct action: the source glass must remain inclined longer to transfer the same half-glass volume. This difference is physical rather than visual. Higher viscosity slows the transfer curve and changes the residual--spill tradeoff, so a model that treats pouring as a generic visual event may underfill slow-flowing liquids or overshoot faster-flowing ones. Third, a parameter-identification query asks which latent fluid parameter best explains an observed fixed-motion pour. In our viscosity-estimation experiment, the robot executes the same prescribed trajectory for the same duration, and candidate simulations with different viscosities are scored against query-relevant quantities: receiver fill curve, residual liquid in the source glass, spilled volume, and first-arrival time in the receiver. Bayesian optimization then searches over viscosity values to identify a parameter estimate, or uncertainty set, that best explains the observed pour and can be passed to the planner for a subsequent target volume or tolerance.

This example illustrates the role of the orchestrator. If the liquid is known, it may directly search over tilt angle, duration, and stopping rule. If the liquid is unknown, it should insert a probing action, estimate viscosity from observed fill, residual, spill, and arrival behavior, and then re-plan under the inferred parameter. If the estimate remains ambiguous near the decision boundary, the response should remain conditional or use feedback termination based on receiver fill. Thus, viscosity is not a visual attribute but a decision-relevant latent variable. A physically viable world model must expose, estimate, and use this variable when the intervention query depends on it.

% \subsection{Simulation suite}
% \label{sec:simulation-suite}

\begin{figure}[t]
    \centering
    \includegraphics[width=\linewidth]{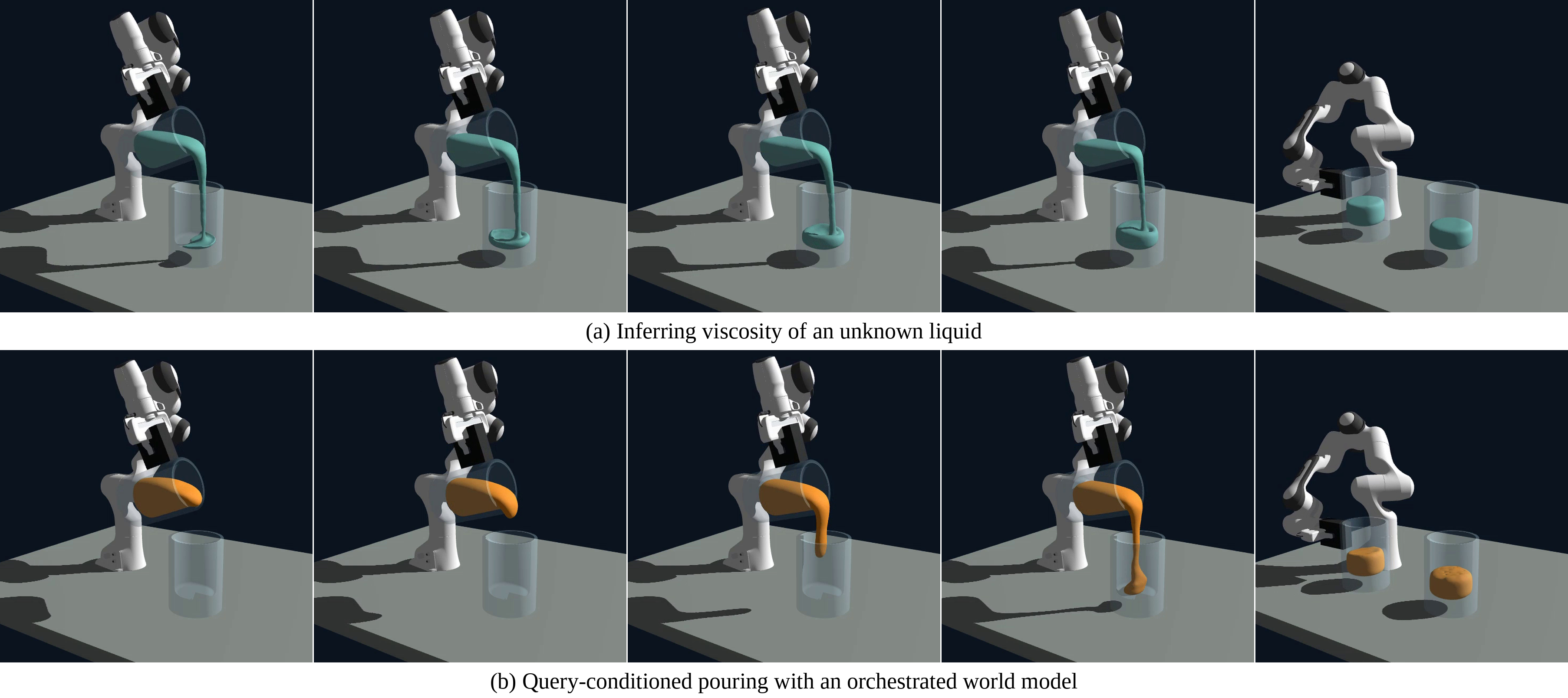}
    \caption{
    Viscosity-dependent pouring under a query-conditioned world model.
    (a) A fixed probing pour is used to infer the viscosity of an unknown liquid from flow, fill, residual, and spill behavior.
    (b) The estimated viscosity is then used to select a pouring motion and stopping condition that transfers a target volume equal to half of the liquid initially in the source glass.
    }
    \label{fig:pouring_query_conditioned}
\end{figure}

\begin{figure}
    \centering

    \includegraphics[width=\linewidth, keepaspectratio]{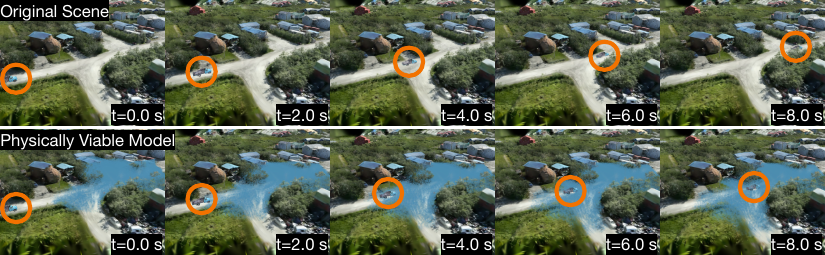}
    
    \caption{Query-conditioned construction of a physically viable driving model from an observation-derived Gaussian splat. Top: the original reconstructed scene representation. Bottom: the scene augmented with rigid support geometry and MPM-based fluid simulation for evaluating flooded-road traversal under rigid-fluid interaction. The orange circle highlights the position of the vehicle in each simulation.}
    \label{fig: physically viable truck demo}
\end{figure}

\textbf{Driving a truck on flooded roads:}
We next consider a driving query in a reconstructed outdoor scene with flooded roads. The intervention query asks whether the truck can traverse the flooded region without becoming immobilized, unstable, or submerged beyond critical depth. The output is a traversal judgment, feasible route, or unsafe region conditioned on vehicle assumptions.

The scene is reconstructed from aerial observations using a Gaussian splat representation~\cite{kerbl20233dgs}, which serves as a local geometric map of road layout, terrain shape, obstacles, and appearance from sparse views. Figure~\ref{fig: physically viable truck demo} compares this observation-based representation with the physically viable model constructed for the traversal query. The top row shows the original Gaussian splat scene with rigid support geometry, while the bottom row augments it with material point method (MPM)-based fluid simulation. The reconstruction alone does not contain the physical structure needed to answer the query, but it provides the geometric basis for constructing the query-specific model.

The orchestrator augments this geometry with intervention-relevant physics. Static scene Gaussians are treated as rigid terrain and support boundaries; reconstructed elevation determines where water pools; and an MPM fluid simulation evolves the flooded region under gravity and rigid interaction. The truck is modeled as a rigid body and evaluated using traversal-relevant quantities, including water depth, drag, wheel submersion, traction loss, and stability margins. Under these flooded-road dynamics, the truck fails the query’s safety condition, indicating that the flooded region is not safely traversable under the assumed vehicle and terrain parameters.

Several quantities remain unresolved from aerial observations alone, including terrain compliance, subsurface support, tire--soil interaction, road friction, water depth, current velocity, and vehicle mass distribution. A physically viable response should therefore expose these assumptions, evaluate sensitivity to unresolved parameters, or return conditional predictions when the outcome depends on variables not identifiable from the available evidence.

\section{Conclusion \& limitations}

In this paper, we argued that physically viable world models should instead construct query-conditioned representations that preserve the physical distinctions relevant to the intervention being considered. In this view, physical structure is not an emergent property of large-scale observation prediction, but an explicit requirement of the model construction process. While this perspective does not solve all aspects of world modeling, it identifies a class of failures that cannot be resolved through scaling alone and motivates architectures that explicitly represent actions, state, dynamics, constraints, and uncertainty.

Nevertheless, several open problems remain. Physical viability is relative to the query, so no single abstraction is sufficient for every intervention. Richer abstractions may improve robustness or transfer, but they also increase the burden of estimation and validation. Creating a model that can identify the right abstraction, at the right level of fidelity to answer a given query, remains an open problem. Physically viable constructions also do not overcome identifiability limits. Latent quantities such as mass, friction, viscosity, or hidden contact geometry may remain unresolved from passive observation, requiring estimation, active interaction, or conditional responses. Finally, the central open problem is orchestration. A physically viable world model must select the abstraction, identify unresolved quantities, choose dynamics and constraints, and determine the form of the response. Task and motion planning offers a precedent for hierarchical decomposition \citep{garrett2020pddlstream, kaelbling2013integrated}, and recent simulator-grounded world models compose perception, simulation, and policy in restricted settings \citep{barcellona2024dream}. Our demonstrations show how intervention queries can drive these choices in controlled settings, but autonomous orchestration remains an open research problem.

\bibliographystyle{plain}
\bibliography{bibliography}

@article{ha2018recurrent,
  title={Recurrent world models facilitate policy evolution},
  author={Ha, David and Schmidhuber, J{\"u}rgen},
  journal={Advances in neural information processing systems},
  volume={31},
  year={2018}
}

@inproceedings{hafner2019learning,
  title={Learning latent dynamics for planning from pixels},
  author={Hafner, Danijar and Lillicrap, Timothy and Fischer, Ian and Villegas, Ruben and Ha, David and Lee, Honglak and Davidson, James},
  booktitle={International conference on machine learning},
  pages={2555--2565},
  year={2019},
  organization={PMLR}
}

@article{kang2024far,
  title={How far is video generation from world model: A physical law perspective},
  author={Kang, Bingyi and Yue, Yang and Lu, Rui and Lin, Zhijie and Zhao, Yang and Wang, Kaixin and Huang, Gao and Feng, Jiashi},
  journal={arXiv preprint arXiv:2411.02385},
  year={2024}
}

@article{chow2025physbench,
  title={Physbench: Benchmarking and enhancing vision-language models for physical world understanding},
  author={Chow, Wei and Mao, Jiageng and Li, Boyi and Seita, Daniel and Guizilini, Vitor and Wang, Yue},
  journal={arXiv preprint arXiv:2501.16411},
  year={2025}
}

@article{meng2024towards,
  title={Towards world simulator: Crafting physical commonsense-based benchmark for video generation},
  author={Meng, Fanqing and Liao, Jiaqi and Tan, Xinyu and Shao, Wenqi and Lu, Quanfeng and Zhang, Kaipeng and Cheng, Yu and Li, Dianqi and Qiao, Yu and Luo, Ping},
  journal={arXiv preprint arXiv:2410.05363},
  year={2024}
}

@inproceedings{zhang2024physdreamer,
  title={Physdreamer: Physics-based interaction with 3d objects via video generation},
  author={Zhang, Tianyuan and Yu, Hong-Xing and Wu, Rundi and Feng, Brandon Y and Zheng, Changxi and Snavely, Noah and Wu, Jiajun and Freeman, William T},
  booktitle={European Conference on Computer Vision},
  pages={388--406},
  year={2024},
  organization={Springer}
}

@article{guo2025t2vphysbench,
  title={T2vphysbench: A first-principles benchmark for physical consistency in text-to-video generation},
  author={Guo, Xuyang and Huo, Jiayan and Shi, Zhenmei and Song, Zhao and Zhang, Jiahao and Zhao, Jiale},
  journal={arXiv preprint arXiv:2505.00337},
  year={2025}
}

@article{gu2025phyworldbench,
  title={" PhyWorldBench": A Comprehensive Evaluation of Physical Realism in Text-to-Video Models},
  author={Gu, Jing and Liu, Xian and Zeng, Yu and Nagarajan, Ashwin and Zhu, Fangrui and Hong, Daniel and Fan, Yue and Yan, Qianqi and Zhou, Kaiwen and Liu, Ming-Yu and others},
  journal={arXiv preprint arXiv:2507.13428},
  year={2025}
}

@article{zhang2025morpheus,
  title={Morpheus: Benchmarking physical reasoning of video generative models with real physical experiments},
  author={Zhang, Chenyu and Cherniavskii, Daniil and Tragoudaras, Antonios and Vozikis, Antonios and Nijdam, Thijmen and Prinzhorn, Derck WE and Bodracska, Mark and Sebe, Nicu and Zadaianchuk, Andrii and Gavves, Efstratios},
  journal={arXiv preprint arXiv:2504.02918},
  year={2025}
}

@article{hafner2025mastering,
  title={Mastering diverse control tasks through world models},
  author={Hafner, Danijar and Pasukonis, Jurgis and Ba, Jimmy and Lillicrap, Timothy},
  journal={Nature},
  volume={640},
  number={8059},
  pages={647--653},
  year={2025},
  publisher={Nature Publishing Group UK London}
}

@article{zhang2026physion,
  title={Physion-Eval: Evaluating Physical Realism in Generated Video via Human Reasoning},
  author={Zhang, Qin and Jing, Peiyu and Yu, Hong-Xing and Ding, Fangqiang and Nie, Fan and Wang, Weimin and Du, Yilun and Zou, James and Wu, Jiajun and Shuai, Bing},
  journal={arXiv preprint arXiv:2603.19607},
  year={2026}
}

@article{raissi2019physics,
  title={Physics-informed neural networks: A deep learning framework for solving forward and inverse problems involving nonlinear partial differential equations},
  author={Raissi, Maziar and Perdikaris, Paris and Karniadakis, George E},
  journal={Journal of Computational physics},
  volume={378},
  pages={686--707},
  year={2019},
  publisher={Elsevier}
}

@article{cranmer2020discovering,
  title={Discovering symbolic models from deep learning with inductive biases},
  author={Cranmer, Miles and Sanchez Gonzalez, Alvaro and Battaglia, Peter and Xu, Rui and Cranmer, Kyle and Spergel, David and Ho, Shirley},
  journal={Advances in neural information processing systems},
  volume={33},
  pages={17429--17442},
  year={2020}
}

@article{hacohen2026ltx,
  title={LTX-2: Efficient Joint Audio-Visual Foundation Model},
  author={HaCohen, Yoav and Brazowski, Benny and Chiprut, Nisan and Bitterman, Yaki and Kvochko, Andrew and Berkowitz, Avishai and Shalem, Daniel and Lifschitz, Daphna and Moshe, Dudu and Porat, Eitan and others},
  journal={arXiv preprint arXiv:2601.03233},
  year={2026}
}

@misc{ltx2_hf,
  title={LTX-2 Model Card},
  author={{Lightricks}},
  year={2026},
  howpublished={\url{https://huggingface.co/Lightricks/LTX-2}},
}

@misc{artificialanalysis_t2v,
  title={Artificial Analysis Text-to-Video Leaderboard (Open-Weights Models)},
  author={{Artificial Analysis}},
  year={2026},
  howpublished={\url{https://artificialanalysis.ai/video/leaderboard/text-to-video?open-weights=true}},
  note={Accessed: 2026-04-27}
}

@misc{artificialanalysis_i2v,
  title={Artificial Analysis Text-to-Video Leaderboard (Open-Weights Models)},
  author={{Artificial Analysis}},
  year={2026},
  howpublished={\url{https://artificialanalysis.ai/video/leaderboard/image-to-video?open-weights=true}},
  note={Accessed: 2026-04-27}
}

@article{bardes2024revisiting,
  title={Revisiting feature prediction for learning visual representations from video},
  author={Bardes, Adrien and Garrido, Quentin and Ponce, Jean and Chen, Xinlei and Rabbat, Michael and LeCun, Yann and Assran, Mahmoud and Ballas, Nicolas},
  journal={arXiv preprint arXiv:2404.08471},
  year={2024}
}

@article{assran2025v,
  title={V-jepa 2: Self-supervised video models enable understanding, prediction and planning},
  author={Assran, Mido and Bardes, Adrien and Fan, David and Garrido, Quentin and Howes, Russell and Muckley, Matthew and Rizvi, Ammar and Roberts, Claire and Sinha, Koustuv and Zholus, Artem and others},
  journal={arXiv preprint arXiv:2506.09985},
  year={2025}
}

@article{zhang2026hierarchical,
  title={Hierarchical Planning with Latent World Models},
  author={Zhang, Wancong and Terver, Basile and Zholus, Artem and Chitnis, Soham and Sutaria, Harsh and Assran, Mido and Balestriero, Randall and Bar, Amir and Bardes, Adrien and LeCun, Yann and others},
  journal={arXiv preprint arXiv:2604.03208},
  year={2026}
}

@inproceedings{motamed2026generative,
  title={Do generative video models understand physical principles?},
  author={Motamed, Saman and Culp, Laura and Swersky, Kevin and Jaini, Priyank and Geirhos, Robert},
  booktitle={Proceedings of the IEEE/CVF Winter Conference on Applications of Computer Vision},
  pages={948--958},
  year={2026}
}

@article{o2025vdaworld,
  title={VDAWorld: World Modelling via VLM-Directed Abstraction and Simulation},
  author={O'Mahony, Felix and Cipolla, Roberto and Tewari, Ayush},
  journal={arXiv preprint arXiv:2512.11061},
  year={2025}
}

@article{schrittwieser2020muzero,
  title={Mastering {Atari}, {Go}, Chess and Shogi by Planning with a Learned Model},
  author={Schrittwieser, Julian and Antonoglou, Ioannis and Hubert, Thomas and Simonyan, Karen and Sifre, Laurent and Schmitt, Simon and Guez, Arthur and Lockhart, Edward and Hassabis, Demis and Graepel, Thore and Lillicrap, Timothy and Silver, David},
  journal={Nature},
  volume={588},
  number={7839},
  pages={604--609},
  year={2020},
  note={arXiv:1911.08265}
}

@inproceedings{hafner2020dreamerv2,
  title={Mastering {Atari} with Discrete World Models},
  author={Hafner, Danijar and Lillicrap, Timothy and Norouzi, Mohammad and Ba, Jimmy},
  booktitle={International Conference on Learning Representations (ICLR)},
  year={2021},
  note={arXiv:2010.02193}
}

@inproceedings{chen2022transdreamer,
  title={{TransDreamer}: Reinforcement Learning with Transformer World Models},
  author={Chen, Chang and Wu, Yi-Fu and Yoon, Jaesik and Ahn, Sungjin},
  booktitle={Deep Reinforcement Learning Workshop, NeurIPS 2021},
  year={2022},
  note={arXiv:2202.09481}
}

@misc{Genesis,
          author = {Genesis Authors},
          title = {Genesis: A Generative and Universal Physics Engine for Robotics and Beyond},
          month = {December},
          year = {2024},
          url = {https://github.com/Genesis-Embodied-AI/Genesis}
        }

@inproceedings{micheli2022transformers,
  title={Transformers are Sample-Efficient World Models},
  author={Micheli, Vincent and Alonso, Eloi and Fleuret, Fran{\c{c}}ois},
  booktitle={International Conference on Learning Representations (ICLR)},
  year={2023},
  note={arXiv:2209.00588}
}

@inproceedings{garrett2020pddlstream,
  title={Pddlstream: Integrating symbolic planners and blackbox samplers via optimistic adaptive planning},
  author={Garrett, Caelan Reed and Lozano-P{\'e}rez, Tom{\'a}s and Kaelbling, Leslie Pack},
  booktitle={Proceedings of the international conference on automated planning and scheduling},
  volume={30},
  pages={440--448},
  year={2020}
}

@article{kaelbling2013integrated,
  title={Integrated task and motion planning in belief space},
  author={Kaelbling, Leslie Pack and Lozano-P{\'e}rez, Tom{\'a}s},
  journal={The International Journal of Robotics Research},
  volume={32},
  number={9-10},
  pages={1194--1227},
  year={2013},
  publisher={Sage Publications Sage UK: London, England}
}

@article{barcellona2024dream,
  title={Dream to manipulate: Compositional world models empowering robot imitation learning with imagination},
  author={Barcellona, Leonardo and Zadaianchuk, Andrii and Allegro, Davide and Papa, Samuele and Ghidoni, Stefano and Gavves, Efstratios},
  journal={arXiv preprint arXiv:2412.14957},
  year={2024}
}

@inproceedings{hansen2023tdmpc2,
  title={{TD-MPC2}: Scalable, Robust World Models for Continuous Control},
  author={Hansen, Nicklas and Su, Hao and Wang, Xiaolong},
  booktitle={International Conference on Learning Representations (ICLR)},
  year={2024},
  note={arXiv:2310.16828}
}

@inproceedings{deng2023s4wm,
  title={Facing Off World Model Backbones: {RNNs}, Transformers, and {S4}},
  author={Deng, Fei and Park, Junyeong and Ahn, Sungjin},
  booktitle={Advances in Neural Information Processing Systems (NeurIPS)},
  year={2023},
  note={arXiv:2307.02064}
}

@misc{The_Newton_Contributors_Newton_GPU-accelerated_physics_2025,
author = {{The Newton Contributors}},
license = {Apache-2.0},
month = apr,
title = {{Newton: GPU-accelerated physics simulation for robotics and simulation research}},
url = {https://github.com/newton-physics/newton},
year = {2025}
}

@article{monaghan2005smoothed,
  title={Smoothed particle hydrodynamics},
  author={Monaghan, Joe J},
  journal={Reports on progress in physics},
  volume={68},
  number={8},
  pages={1703--1759},
  year={2005}
}

@inproceedings{bender2015divergence,
  title={Divergence-free smoothed particle hydrodynamics},
  author={Bender, Jan and Koschier, Dan},
  booktitle={Proceedings of the 14th ACM SIGGRAPH/Eurographics symposium on computer animation},
  pages={147--155},
  year={2015}
}

@inproceedings{gupta2022maskvit,
  title={{MaskViT}: Masked Visual Pre-Training for Video Prediction},
  author={Gupta, Agrim and Tian, Stephen and Zhang, Yunzhi and Wu, Jiajun and Mart{\'i}n-Mart{\'i}n, Roberto and Li, Fei-Fei},
  booktitle={International Conference on Learning Representations (ICLR)},
  year={2023},
  note={arXiv:2206.11894}
}

@inproceedings{janner2022planning,
  title={Planning with Diffusion for Flexible Behavior Synthesis},
  author={Janner, Michael and Du, Yilun and Tenenbaum, Joshua B. and Levine, Sergey},
  booktitle={Proceedings of the 39th International Conference on Machine Learning (ICML)},
  series={Proceedings of Machine Learning Research},
  volume={162},
  pages={9902--9915},
  publisher={PMLR},
  year={2022},
  note={arXiv:2205.09991}
}

@article{ding2024diffusion,
  title={Diffusion World Model: Future Modeling Beyond Step-by-Step Rollout for Offline Reinforcement Learning},
  author={Ding, Zihan and Zhang, Amy and Tian, Yuandong and Zheng, Qinqing},
  journal={arXiv preprint arXiv:2402.03570},
  year={2024}
}

@article{rigter2024avid,
  title={{AVID}: Adapting Video Diffusion Models to World Models},
  author={Rigter, Marc and Gupta, Tarun and Hilmkil, Agrin and Ma, Chao},
  journal={arXiv preprint arXiv:2410.12822},
  year={2024}
}

@article{huang2025vid2world,
  title={{Vid2World}: Crafting Video Diffusion Models to Interactive World Models},
  author={Huang, Siqiao and Wu, Jialong and Zhou, Qixing and Miao, Shangchen and Long, Mingsheng},
  journal={arXiv preprint arXiv:2505.14357},
  year={2025}
}

@inproceedings{chang2016compositional,
  title={A Compositional Object-Based Approach to Learning Physical Dynamics},
  author={Chang, Michael B. and Ullman, Tomer and Torralba, Antonio and Tenenbaum, Joshua B.},
  booktitle={International Conference on Learning Representations (ICLR)},
  year={2017},
  note={arXiv:1612.00341}
}

@inproceedings{li2018learning,
  title={Learning Particle Dynamics for Manipulating Rigid Bodies, Deformable Objects, and Fluids},
  author={Li, Yunzhu and Wu, Jiajun and Tedrake, Russ and Tenenbaum, Joshua B. and Torralba, Antonio},
  booktitle={International Conference on Learning Representations (ICLR)},
  year={2019},
  note={arXiv:1810.01566}
}

@inproceedings{sanchezgonzalez2020learning,
  title={Learning to Simulate Complex Physics with Graph Networks},
  author={Sanchez-Gonzalez, Alvaro and Godwin, Jonathan and Pfaff, Tobias and Ying, Rex and Leskovec, Jure and Battaglia, Peter W.},
  booktitle={International Conference on Machine Learning (ICML)},
  year={2020},
  note={arXiv:2002.09405}
}

@inproceedings{pfaff2020learning,
  title={Learning Mesh-Based Simulation with Graph Networks},
  author={Pfaff, Tobias and Fortunato, Meire and Sanchez-Gonzalez, Alvaro and Battaglia, Peter W.},
  booktitle={International Conference on Learning Representations (ICLR)},
  year={2021},
  note={arXiv:2010.03409}
}

@article{hafner2019dream,
  title={Dream to control: Learning behaviors by latent imagination},
  author={Hafner, Danijar and Lillicrap, Timothy and Ba, Jimmy and Norouzi, Mohammad},
  journal={arXiv preprint arXiv:1912.01603},
  year={2019}
}

@inproceedings{difftaichi,
  title={{DiffTaichi}: Differentiable Programming for Physical Simulation},
  author={Hu, Yuanming and Anderson, Luke and Li, Tzu-Mao and Sun, Qi and Carr, Nathan and Ragan-Kelley, Jonathan and Durand, Fr{\'e}do},
  booktitle={International Conference on Learning Representations (ICLR)},
  year={2020},
  note={arXiv:1910.00935}
}

@inproceedings{jatavallabhula2021gradsim,
  title={{gradSim}: Differentiable simulation for system identification and visuomotor control},
  author={Jatavallabhula, Krishna Murthy and Macklin, Miles and Golemo, Florian and Voleti, Vikram and Petrini, Linda and Weiss, Martin and Considine, Breandan and Parent-Levesque, Jerome and Xie, Kevin and Erleben, Kenny and Paull, Liam and Shkurti, Florian and Nowrouzezahrai, Derek and Fidler, Sanja},
  booktitle={International Conference on Learning Representations (ICLR)},
  year={2021},
  note={arXiv:2104.02646}
}

@article{brax,
  title={{Brax} -- A Differentiable Physics Engine for Large Scale Rigid Body Simulation},
  author={Freeman, C. Daniel and Frey, Erik and Raichuk, Anton and Girgin, Sertan and Mordatch, Igor and Bachem, Olivier},
  journal={arXiv preprint arXiv:2106.13281},
  year={2021}
}

@inproceedings{physgaussian,
  title={{PhysGaussian}: Physics-Integrated 3D Gaussians for Generative Dynamics},
  author={Xie, Tianyi and Zong, Zeshun and Qiu, Yuxing and Li, Xuan and Feng, Yutao and Yang, Yin and Jiang, Chenfanfu},
  booktitle={IEEE/CVF Conference on Computer Vision and Pattern Recognition (CVPR)},
  year={2024},
  note={arXiv:2311.12198}
}

@article{abouchakra2024,
  title={Physically Embodied {Gaussian} Splatting: A Realtime Correctable World Model for Robotics},
  author={Abou-Chakra, Jad and Rana, Krishan and Dayoub, Feras and S{\"u}nderhauf, Niko},
  journal={arXiv preprint arXiv:2406.10788},
  year={2024}
}

@inproceedings{huang2024dreamphysics,
  title={{DreamPhysics}: Learning Physics-Based 3D Dynamics with Video Diffusion Priors},
  author={Huang, Tianyu and Zhang, Haoze and Zeng, Yihan and Zhang, Zhilu and Li, Hui and Zuo, Wangmeng and Lau, Rynson W. H.},
  booktitle={AAAI Conference on Artificial Intelligence},
  year={2025},
  note={arXiv:2406.01476}
}

@inproceedings{liu2024physflow,
  title={{PhysFlow}: Unleashing the Potential of Multi-modal Foundation Models and Video Diffusion for 4D Dynamic Physical Scene Simulation},
  author={Liu, Zhuoman and Ye, Weicai and Luximon, Yan and Wan, Pengfei and Zhang, Di},
  booktitle={IEEE/CVF Conference on Computer Vision and Pattern Recognition (CVPR)},
  year={2025},
  note={arXiv:2411.14423}
}

@inproceedings{chen2025physgen3d,
  title={{PhysGen3D}: Crafting a Miniature Interactive World from a Single Image},
  author={Chen, Boyuan and Jiang, Hanxiao and Liu, Shaowei and Gupta, Saurabh and Li, Yunzhu and Zhao, Hao and Wang, Shenlong},
  booktitle={IEEE/CVF Conference on Computer Vision and Pattern Recognition (CVPR)},
  year={2025},
  note={arXiv:2503.20746}
}

@article{zhou2025chronodreamer,
  title={{ChronoDreamer}: Action-Conditioned World Model as an Online Simulator for Robotic Planning},
  author={Zhou, Zhenhao and Negrut, Dan},
  journal={arXiv preprint arXiv:2512.18619},
  year={2025}
}

@article{kerbl20233dgs,
  title={{3D Gaussian Splatting} for Real-Time Radiance Field Rendering},
  author={Kerbl, Bernhard and Kopanas, Georgios and Leimk{\"u}hler, Thomas and Drettakis, George},
  journal={ACM Transactions on Graphics},
  volume={42},
  number={4},
  year={2023},
  note={arXiv:2308.04079}
}

@article{liu2023ifactor,
  title={Learning World Models with Identifiable Factorization},
  author={Liu, Yu-Ren and Huang, Biwei and Zhu, Zhengmao and Tian, Honglong and Gong, Mingming and Yu, Yang and Zhang, Kun},
  journal={arXiv preprint arXiv:2306.06561},
  year={2023}
}

@inproceedings{locatello2020,
  title={Object-Centric Learning with Slot Attention},
  author={Locatello, Francesco and Weissenborn, Dirk and Unterthiner, Thomas and Mahendran, Aravindh and Heigold, Georg and Uszkoreit, Jakob and Dosovitskiy, Alexey and Kipf, Thomas},
  booktitle={Advances in Neural Information Processing Systems (NeurIPS)},
  year={2020},
  note={arXiv:2006.15055}
}

@inproceedings{kipf2020,
  title={Contrastive Learning of Structured World Models},
  author={Kipf, Thomas and van der Pol, Elise and Welling, Max},
  booktitle={International Conference on Learning Representations (ICLR)},
  year={2020},
  note={arXiv:1911.12247}
}

@inproceedings{greydanus2019hnn,
  title={Hamiltonian Neural Networks},
  author={Greydanus, Sam and Dzamba, Misko and Yosinski, Jason},
  booktitle={Advances in Neural Information Processing Systems (NeurIPS)},
  year={2019},
  note={arXiv:1906.01563}
}

@inproceedings{cranmer2020lnn,
  title={Lagrangian Neural Networks},
  author={Cranmer, Miles and Greydanus, Sam and Hoyer, Stephan and Battaglia, Peter and Spergel, David and Ho, Shirley},
  booktitle={ICLR Workshop on Deep Differential Equations},
  year={2020},
  note={arXiv:2003.04630}
}

@article{fourcastnet,
  title={{FourCastNet}: A Global Data-driven High-resolution Weather Model using Adaptive Fourier Neural Operators},
  author={Pathak, Jaideep and Subramanian, Shashank and Harrington, Peter and Raja, Sanjeev and Chattopadhyay, Ashesh and Mardani, Morteza and Kurth, Thorsten and Hall, David and Li, Zongyi and Azizzadenesheli, Kamyar and Hassanzadeh, Pedram and Kashinath, Karthik and Anandkumar, Animashree},
  journal={arXiv preprint arXiv:2202.11214},
  year={2022}
}

@article{graphcast,
  title={{GraphCast}: Learning skillful medium-range global weather forecasting},
  author={Lam, Remi and Sanchez-Gonzalez, Alvaro and Willson, Matthew and Wirnsberger, Peter and Fortunato, Meire and Alet, Ferran and Ravuri, Suman and Ewalds, Timo and Eaton-Rosen, Zach and Hu, Weihua and Merose, Alexander and Hoyer, Stephan and Holland, George and Vinyals, Oriol and Stott, Jacklynn and Pritzel, Alexander and Mohamed, Shakir and Battaglia, Peter},
  journal={arXiv preprint arXiv:2212.12794},
  year={2022}
}

@article{panguweather,
  title={{Pangu-Weather}: A 3D High-Resolution Model for Fast and Accurate Global Weather Forecast},
  author={Bi, Kaifeng and Xie, Lingxi and Zhang, Hengheng and Chen, Xin and Gu, Xiaotao and Tian, Qi},
  journal={arXiv preprint arXiv:2211.02556},
  year={2022}
}

@article{hafner2023mastering,
  title={Mastering diverse domains through world models},
  author={Hafner, Danijar and Pasukonis, Jurgis and Ba, Jimmy and Lillicrap, Timothy},
  journal={arXiv preprint arXiv:2301.04104},
  year={2023}
}

@article{ha2018world,
  title={World models},
  author={Ha, David and Schmidhuber, J{\"u}rgen},
  journal={arXiv preprint arXiv:1803.10122},
  volume={2},
  number={3},
  pages={440},
  year={2018}
}

@article{chen2018neural,
  title={Neural ordinary differential equations},
  author={Chen, Ricky TQ and Rubanova, Yulia and Bettencourt, Jesse and Duvenaud, David K},
  journal={Advances in neural information processing systems},
  volume={31},
  year={2018}
}

@article{battaglia2016interaction,
  title={Interaction networks for learning about objects, relations and physics},
  author={Battaglia, Peter and Pascanu, Razvan and Lai, Matthew and Jimenez Rezende, Danilo and others},
  journal={Advances in neural information processing systems},
  volume={29},
  year={2016}
}

@article{beucler2021enforcing,
  title={Enforcing analytic constraints in neural networks emulating physical systems},
  author={Beucler, Tom and Pritchard, Michael and Rasp, Stephan and Ott, Jordan and Baldi, Pierre and Gentine, Pierre},
  journal={Physical review letters},
  volume={126},
  number={9},
  pages={098302},
  year={2021},
  publisher={APS}
}

\newpage
\appendix

\section{Platforms for Numerical Experimentation}
In this section, we describe the computational tools and platforms used in our experimental pipeline. These include foundation models (e.g., vision-language models, video diffusion models) as well as simulation environments for controlled evaluation.
We prioritize open-source models with competitive performance to balance reproducibility and capability. Where applicable, we report model variants, prompts and implementation details to facilitate replication of our results.

\subsection{Physics Engines for Controlled Experiments}
\label{sec:simulator_grounded_rollouts}
We use simulator-generated rollouts as controlled references for intervention queries. In each rollout, the initial scene, material and contact parameters, solver settings, and intervention are specified explicitly, and the state is advanced by numerical time integration rather than by a learned transition model. Rendered videos are then derived from the simulated state trajectory. 
These rollouts are not necessarily identical to real-world physics, yet they serve as controlled physical references under known modeling assumptions.

We use NVIDIA Newton~\cite{The_Newton_Contributors_Newton_GPU-accelerated_physics_2025}, a Warp-based physics engine, for rigid-body, deformable-body, and robot-contact experiments. Newton provides GPU-accelerated simulation with XPBD-based contact for rigid bodies and particle-based methods for deformable objects, along with support for articulated robot interaction.
We instantiate a small set of controlled scenes, most following a ramp interaction: a ball descends an incline, transitions to a flat surface, and interacts with a target object (e.g., rigid stack, container, or deformable body). Scene geometry, camera, and rendering are held fixed while physical parameters—such as density, friction, and restitution—are varied through Newton’s material and contact models.
Rigid interactions are resolved with the XPBD contact solver, while deformable objects use particle-based simulation such as material point method (MPM) with self-contact. We additionally include a robot-contact setting in which a Franka Panda end-effector follows a prescribed trajectory and interacts with the environment under Newton’s articulated-body and contact simulation.

We also use Genesis~\cite{Genesis}, a GPU-accelerated physics simulation platform designed for robotics and embodied AI in fluid-coupled experiments. In particular, Genesis supports particle-based fluid simulation via smoothed-particle hydrodynamics (SPH)~\cite{monaghan2005smoothed}, including divergence-free formulations (DFSPH)~\cite{bender2015divergence}, and couples these with rigid-body dynamics through shared collision and boundary representations.
In our setup, liquids are represented explicitly as SPH particles and evolved jointly with rigid bodies under the same simulation loop. Rigid–fluid interaction is handled through mesh-derived signed distance fields, enabling consistent coupling between container motion and fluid response. We use this capability to construct a small set of controlled fluid–interaction scenarios, including container transport and pouring, where object motion may be prescribed but fluid dynamics are fully simulated.
Similarly, across these experiments, we hold scene geometry, camera, and control trajectories fixed, and vary only fluid properties such as viscosity. This isolates fluid parameters as the latent factors governing outcomes such as retention, transfer, and spillage, while all state evolution is determined by numerical simulation rather than learned dynamics.

Across both engines, the simulator state defines the reference trajectory, and rendered videos are derived from this state. Physical parameters are specified or calibrated independently rather than inferred end-to-end from visual input, separating simulation from perception and parameter estimation.

\subsection{Vision Language Model}

We evaluate \texttt{GPT-5.5} as a vision-language baseline for direct physical outcome prediction from static scene observations. We test whether a VLM can infer the likely physical evolution of a scene directly from an image and a natural-language query.
We repeat each prompt five times and exclude blank responses from the qualitative analysis. The model uses ``medium'' reasoning effort by default. Filenames are hidden from the model so that labels such as \texttt{cup\_water.png} do not leak physical information.

\subsection{Audio-Visual Foundation Model}
We employ \texttt{LTX-2}~\cite{hacohen2026ltx} as our primary video generation model. 
LTX-2 is an open-source diffusion transformer (DiT)-based audio-visual foundation model designed to generate temporally consistent video with synchronized audio within a unified architecture. In this work, we focus primarily on video fidelity and temporal accuracy.
The model supports multiple conditioning modalities, including text-to-video and image-to-video generation. In our experiments, we primarily use text-conditioned video generation with keyframe anchoring. The resulting outputs are used as a diffusion-based baseline for comparison against physics-driven simulation results.

For our experiments, we follow the two-stage inference pipeline provided in the official \texttt{LTX-2} implementation~\cite{ltx2_hf}. The base model consists of approximately $19$ billion parameters, with additional upscaling and refinement models used in the second stage. To ensure reproducibility, all trials are seeded. We additionally anchor the initial frame (frame $0$) with full strength to enforce a consistent initialization aligned with the reference video.

As of April 2026, \texttt{LTX-2} ranks highly among open-source video generation models. According to Artificial Analysis, which evaluates models across multiple quality and temporal consistency metrics, \texttt{LTX-2} is ranked first on both the text-to-video and image-to-video leaderboards~\cite{artificialanalysis_t2v,artificialanalysis_i2v}. This performance supports its use as a strong baseline in our comparative evaluation. 

\subsection{Vision-based World Models}
Vision-based world models remain the most common and actively researched subcategory for this subject. In this work, we also experiment with the \texttt{V-JEPA 2} platform~\cite{assran2025v} from Meta. \texttt{V-JEPA 2} is a publicly available self-supervised video representation model based on the joint-embedding predictive architecture. Unlike diffusion-based video generators, V-JEPA~\cite{bardes2024revisiting} does not generate pixels directly; instead, it learns predictive representations by modeling masked or future regions in latent space. V-JEPA 2 extends this framework with large-scale video pretraining and limited robot interaction data, enabling and achieving competitive results in downstream tasks such as video understanding and motion planning in robotics. 

In our experiments, \texttt{V-JEPA 2} is used as a latent predictive world-modeling baseline; in particular, we utilize the \texttt{V-JEPA 2-AC} pipeline~\cite{assran2025v,zhang2026hierarchical}, an action-conditioned dynamics model that predicts future latent states given the current state and candidate actions, enabling forward simulation without explicit physics modeling. At runtime, control is performed via model predictive control (MPC), where sequences of actions are sampled, rolled out through the learned dynamics, and evaluated using a task-specific cost defined in latent space, usually the distance to a goal representation. The process is purely vision-based and not explicitly physically constrained; we compare the action sequences to true physics simulation results across different control scenarios.

\section{Additional Experimental Results}
In this section, we provide details and additional numerical experiments and results. 

\subsection{VLM prediction under controlled physical variation}
\label{subsec:vlm_details}

\begin{figure}[!t]
    \centering
    \includegraphics[width=0.24\linewidth, keepaspectratio]{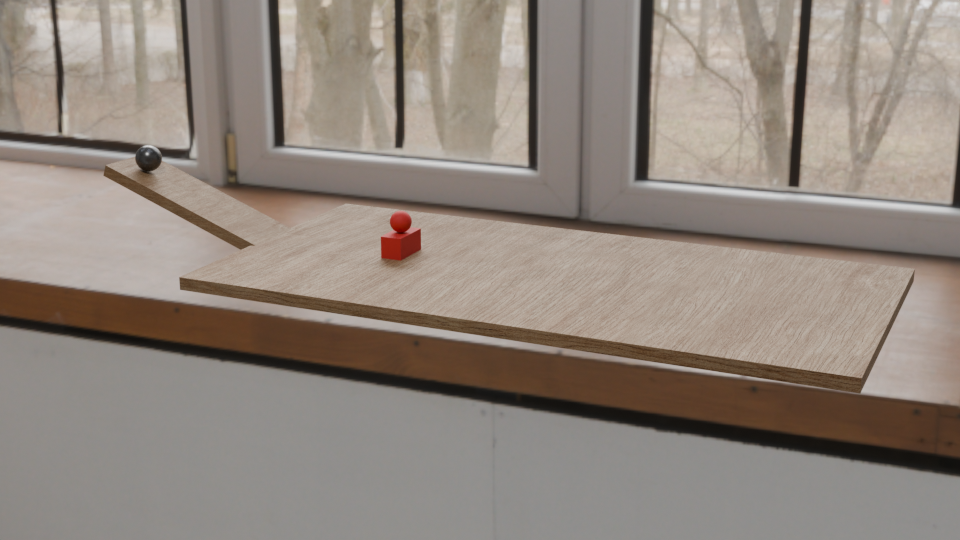}
    \includegraphics[width=0.24\linewidth, keepaspectratio]{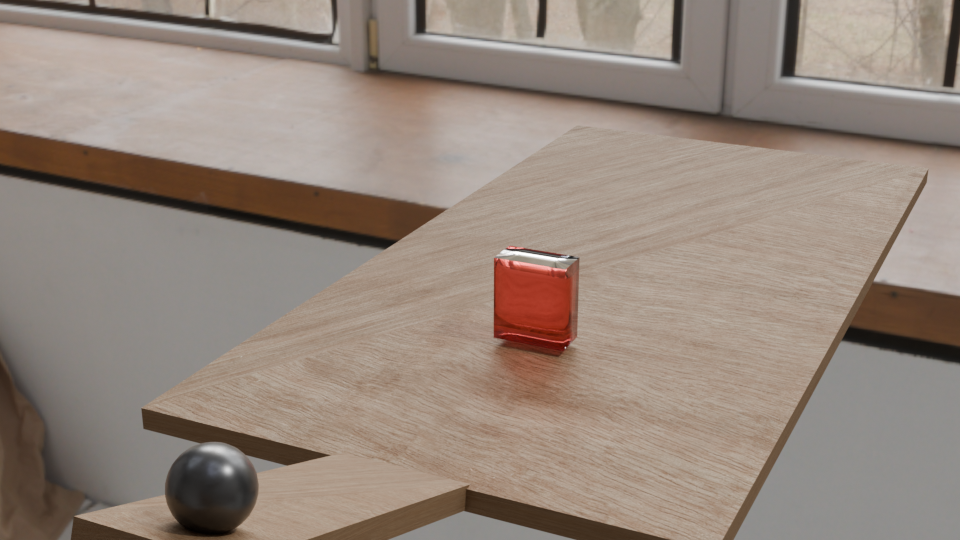}
    \includegraphics[width=0.24\linewidth, keepaspectratio]{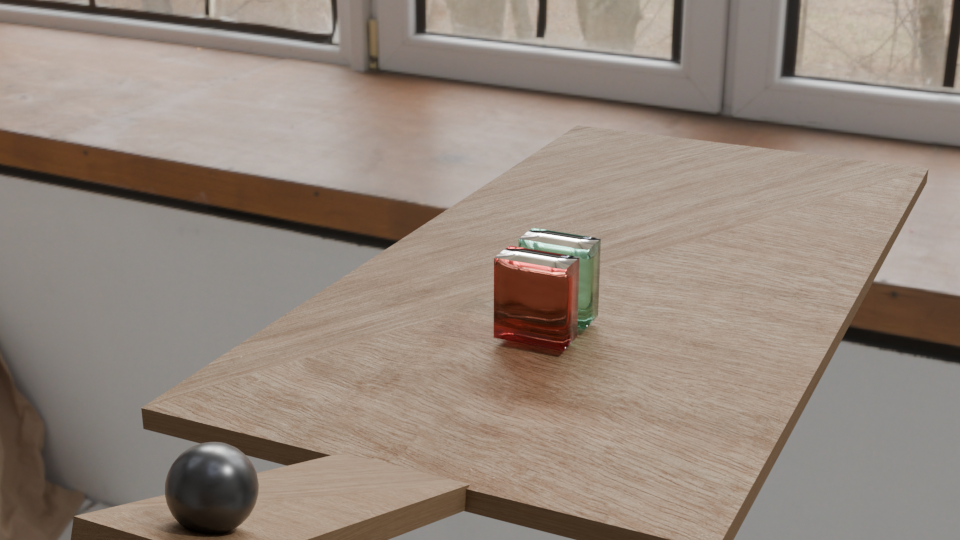}
    \includegraphics[width=0.24\linewidth, keepaspectratio]{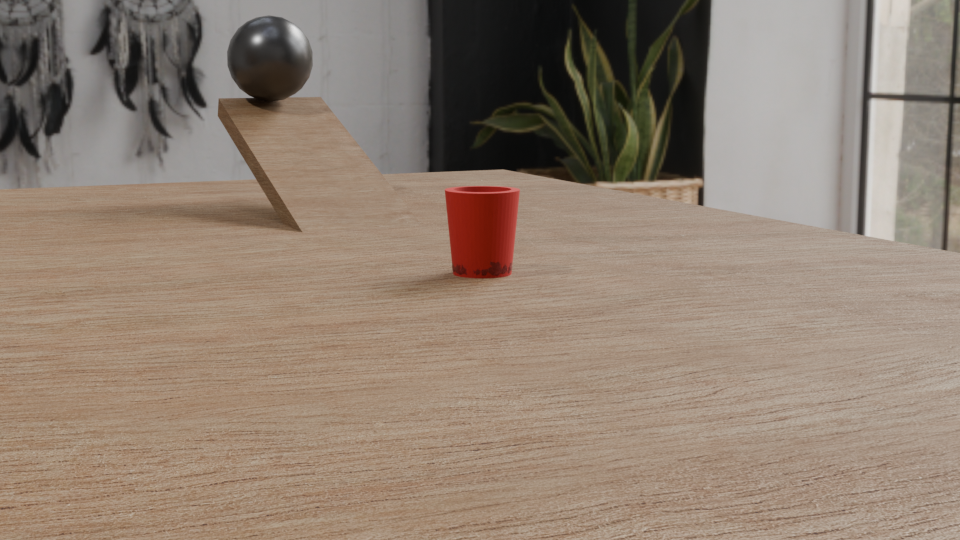}
    \caption{Static-image benchmark for VLM physical prediction. Each scene shows a ball rolling down a ramp into different interaction regimes: rigid-body collision, single deformable-wall impact, double deformable-wall impact, and rigid–fluid interaction with a water-filled cup. The benchmark tests whether VLM predictions track the underlying physical response under intervention rather than only visual similarity.}
    \label{fig: vlm tests}
\end{figure}

We evaluate \texttt{GPT-5.5} on static-image physical prediction using four rendered ramp-interaction scenes: rigid blocks and ball, a single deformable jelly wall, two deformable jelly walls, and a water-filled cup shown in Figure \ref{fig: vlm tests}. Each scene places a ball near the top of a wooden ramp connected to a flat platform, with the target in the ball’s downstream path. 

Each image is queried under multiple levels of context. The \texttt{no\_context} prompt asks the model to describe the motion and final resting locations of the objects using only the image and a generic instruction. The \texttt{low\_context} prompt names the main objects and the fixed ramp-platform geometry but does not specify material properties or the full intervention. The \texttt{high\_context} prompt specifies that the ball is made of aluminum, that it is released from rest, that it rolls down the ramp under gravity, and that it collides with the relevant target object. For the cup scene, the high-context prompt also states that the cup is filled with water. This design probes whether additional textual context changes the model's physical prediction, especially when the relevant physical variables are latent, visually ambiguous, or only partially specified.
We additionally evaluate \texttt{counterfactual} prompts that modify one physical property of the scene while keeping the visible configuration fixed, including material changes, altered release height, increased friction, and increased fluid viscosity.

For the cup scene, the no-context query asks only: \texttt{Describe the motion and final resting location of the objects in the scene. Explain your reasoning.} The low-context query additionally specifies the ramp, platform, ball, and cup. The high-context query specifies that an aluminum ball is released from rest, rolls down the ramp under gravity, and collides with a water-filled cup positioned in its path. The counterfactual query modifies the setup further by asking how the outcome changes if the ball starts from half the ramp height and the liquid is honey instead of water. The other scenes are worded similarly. 

The reference outcomes are: (i) the rigid blocks and red ball scatter; (ii) the single jelly wall topples forward because the ball strikes below its center of mass; (iii) the two jelly walls deform but remain upright and slide because table friction is low; and (iv) the water-filled cup tips and spills.

The responses show several recurring patterns.

\begin{enumerate}[leftmargin=*]
    \item 
    \textbf{Low-context prompts sometimes produce incorrect physical interpretations of the scene.} In several trials, the model predicts no collision because it states that the target is not in the ball’s path, particularly in the cup and double-jelly scenes. Other responses infer incorrect scene geometry or motion entirely, including predicting that objects slide off the table due to the apparent camera angle or assuming that the ball is never released. These responses show that, under weak contextual specification, the model can construct a qualitatively incorrect physical scenario before detailed reasoning about collision, deformation, or fluid interaction occurs.
    \item 
    \textbf{High-context prompts isolate the remaining physical prediction errors.} When release and collision are explicitly specified, many low-context ambiguities disappear and the responses usually invoke qualitatively relevant effects including rolling motion, momentum transfer, deformation, frictional dissipation, and water sloshing. However, the model still does not consistently or precisely predict the realized outcome. In the single-jelly-wall scene, responses often predict deformation or sliding but omit the observed forward toppling caused by the low strike point. In the two-jelly-wall scene, some responses correctly describe sliding deformation without overturning, while others predict collapse or toppling. In the cup scene, the model often states that tipping and spilling may occur rather than predicting the observed spill outcome. The remaining errors therefore persist even after the intervention itself is explicitly specified.
    \item 
    \textbf{Counterfactual prompts improve directional reasoning but still produce threshold uncertainty.} The model usually predicts the correct qualitative trend: lower release height reduces impact energy, honey damps sloshing, high friction suppresses sliding, and gelatin dissipates more energy than rigid blocks. However, the responses still frequently describe tipping, toppling, sliding, and spilling as possible outcomes rather than resolving which event occurs. In one response, the model also assumes that an aluminum ball has lower mass than the original ball even though the original material is unspecified.
\end{enumerate}

These results separate qualitative explanation from intervention prediction. The VLM can identify relevant physical effects, especially when the prompt supplies the intervention. It does not consistently predict the realized outcome when that outcome depends on thresholded contact, deformation, or rigid–fluid dynamics. A physically viable model must represent the relevant variables, propagate them through the selected dynamics, and return the realized outcome or a justified uncertainty set.

\subsection{Visual Foundation Models are Not Physically Viable Simulators}
\label{subsec:diffusion_results}
While recent advances in diffusion-based video generation have significantly improved visual fidelity and temporal coherence, these models remain fundamentally limited and unfit as physical simulators. In particular, conditioning on key frames and using text-based prompt guidance do not sufficiently constrain the underlying dynamics to ensure physical consistency over time.

\begin{figure}[t]
    \centering
    \includegraphics[width=\linewidth]{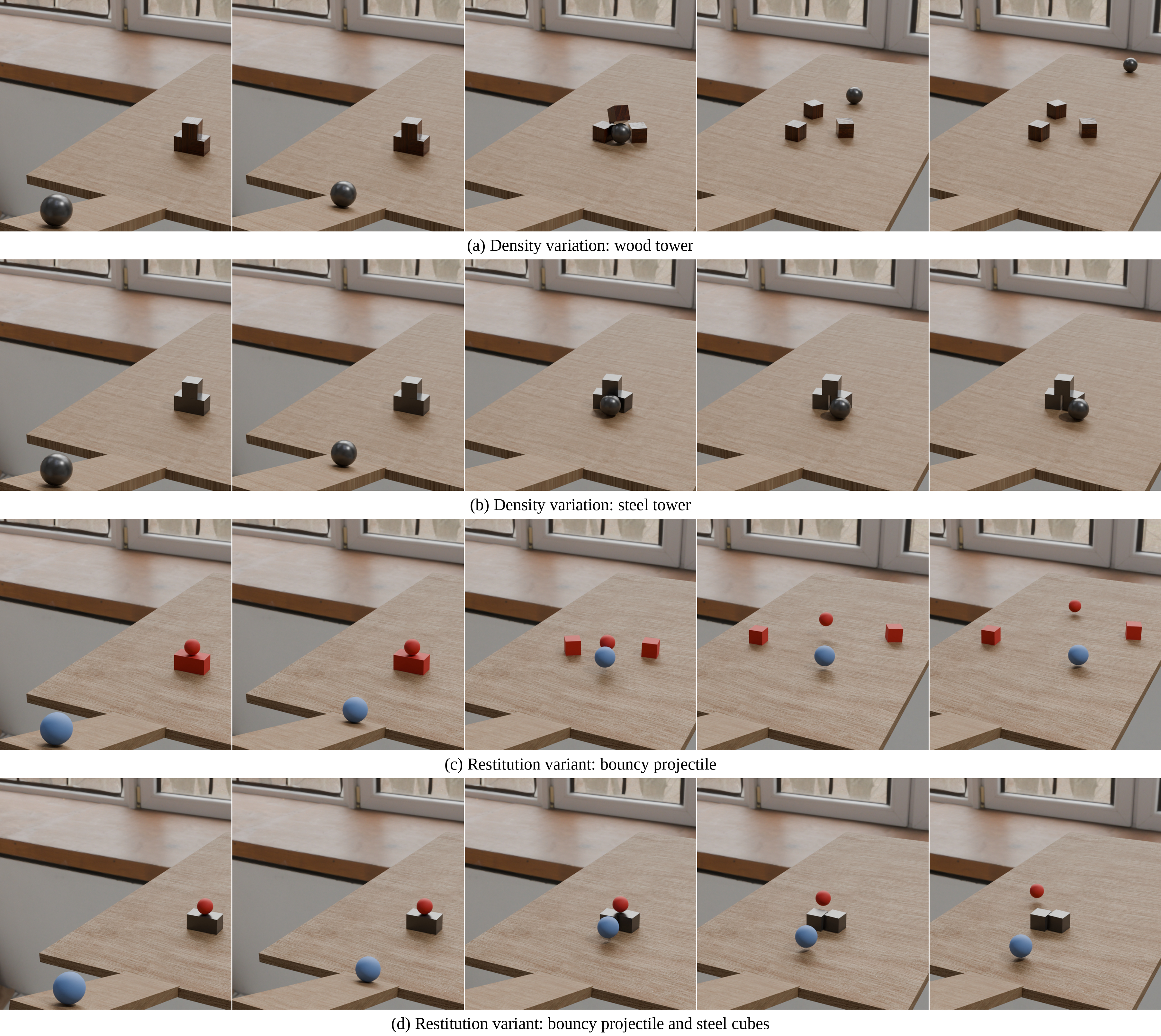}
    \caption{
    Controlled ramp-to-tower rigid-body interactions under latent material variation.
    Each row shows a time sequence from one condition in the simulation suite:
    (a) the density-variation setup with a wood tower,
    (b) the density-variation setup with a steel tower,
    (c) the restitution setup with a high-restitution projectile, and
    (d) the restitution setup with both a high-restitution projectile and steel cubes.
    }
    \label{fig:ramp_tower_variants}
\end{figure}

\begin{figure}[t]
    \centering
    \begin{subfigure}{\textwidth}
        \centering
        \includegraphics[width=0.16\textwidth]{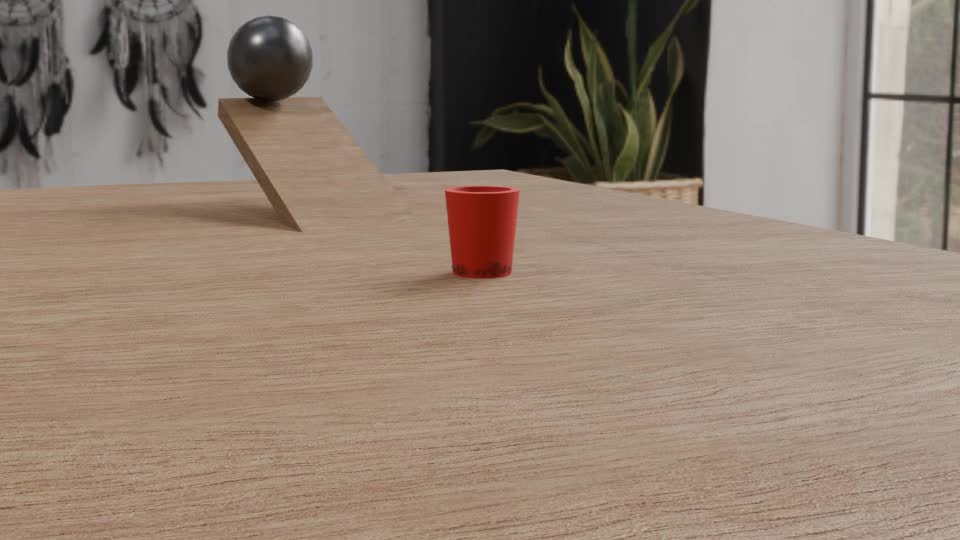}
        \includegraphics[width=0.16\textwidth]{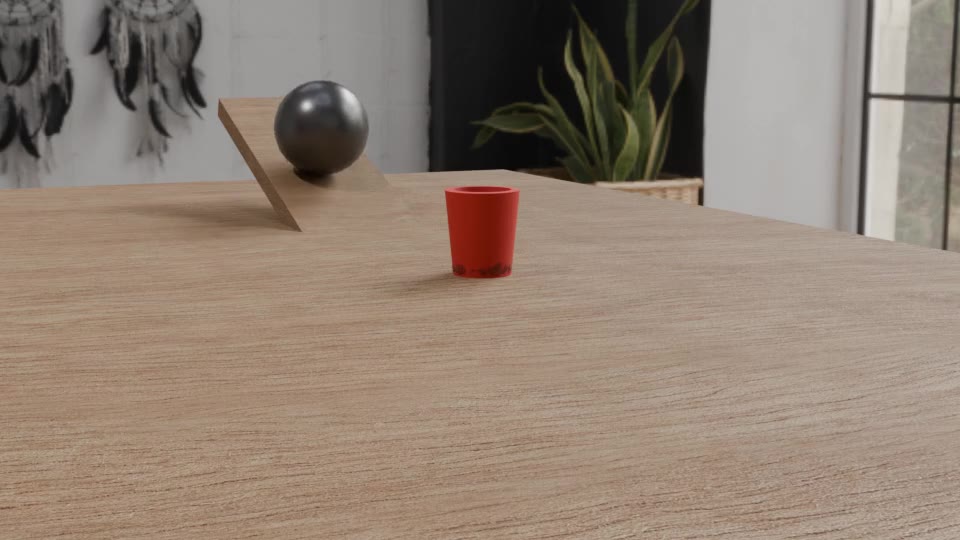}
        \includegraphics[width=0.16\textwidth]{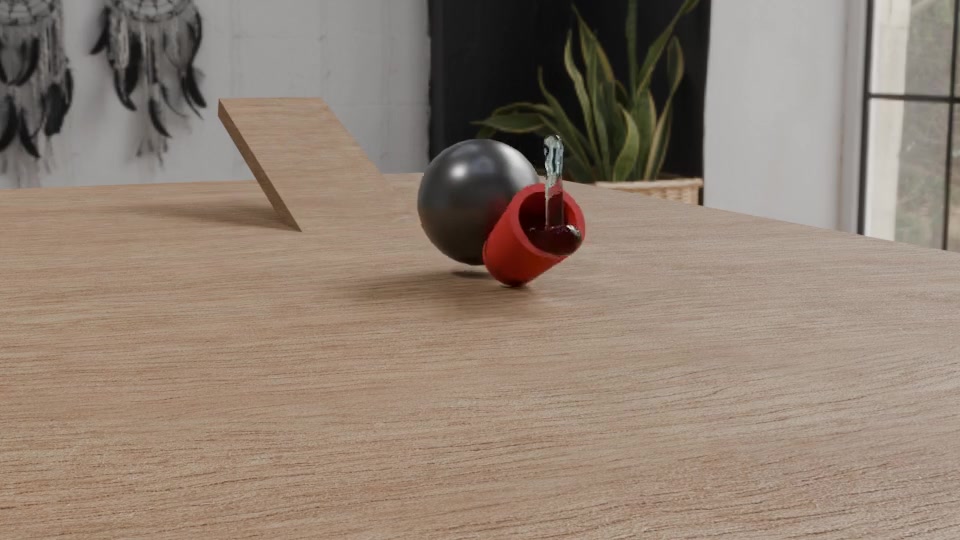}
        \includegraphics[width=0.16\textwidth]{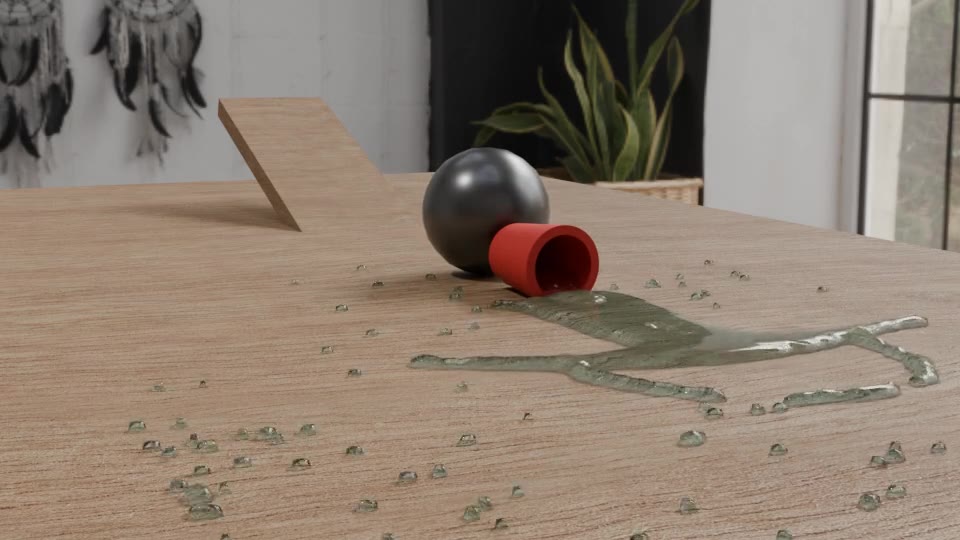}
        \includegraphics[width=0.16\textwidth]{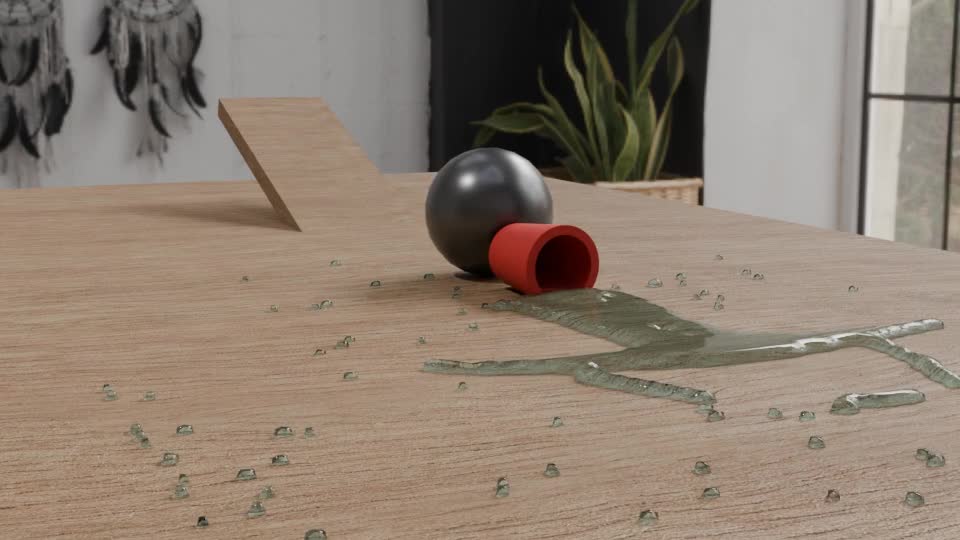}
        \includegraphics[width=0.16\textwidth]{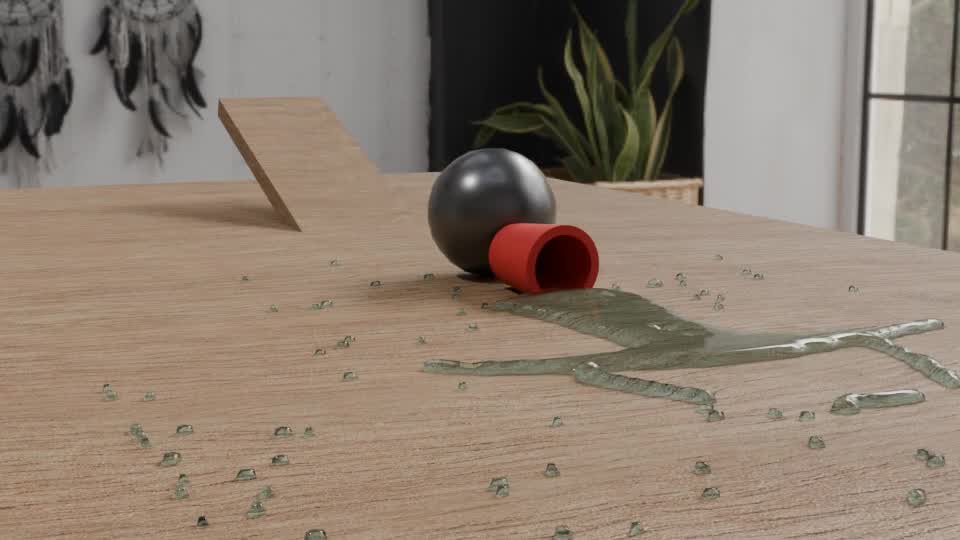}
        \caption{Physical simulation plus rendering.}
        \label{fig:cup_demo}
    \end{subfigure}
    \hfill
    \begin{subfigure}{\textwidth}
        \centering
        \includegraphics[width=0.16\textwidth]{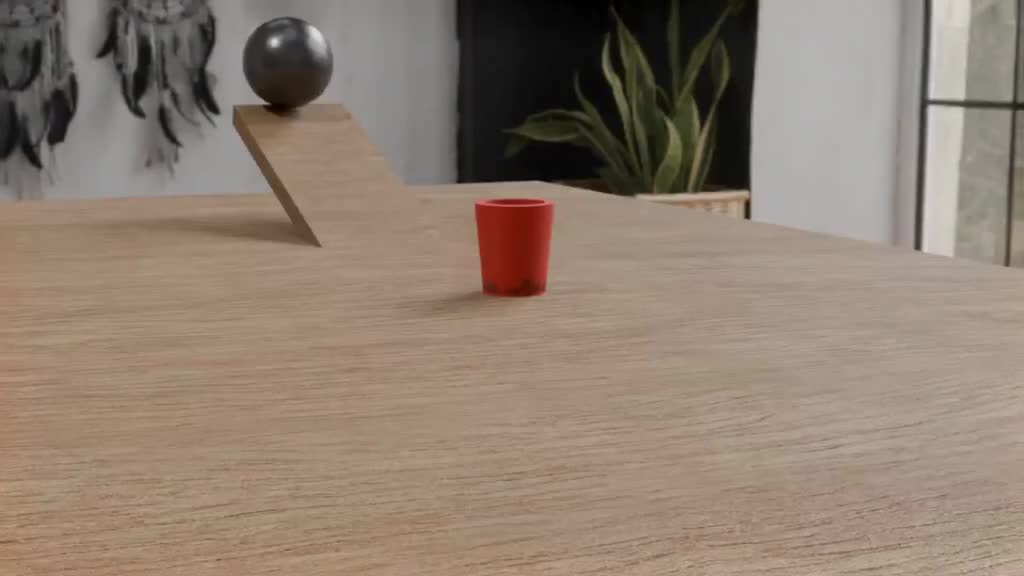}
        \includegraphics[width=0.16\textwidth]{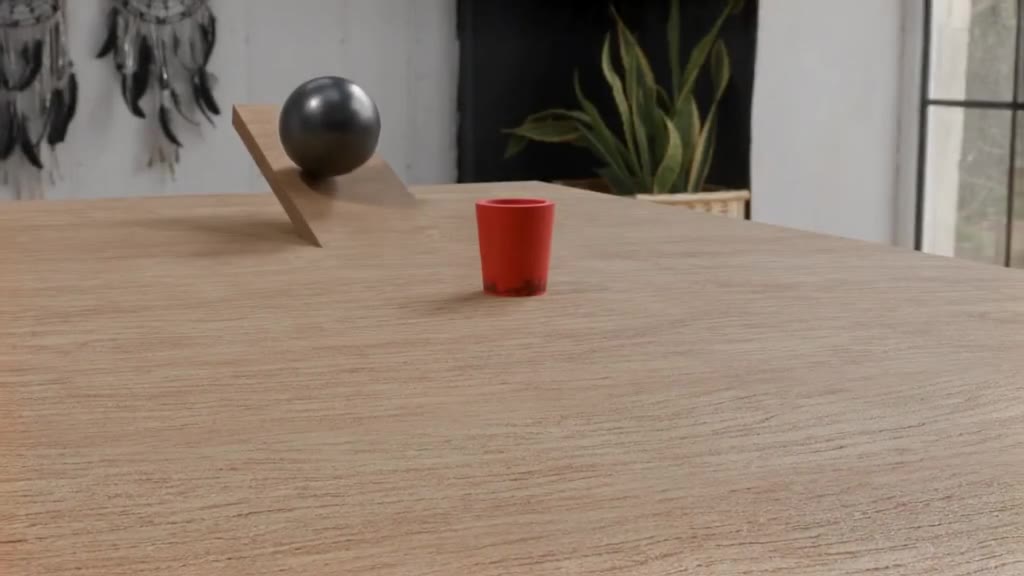}
        \includegraphics[width=0.16\textwidth]{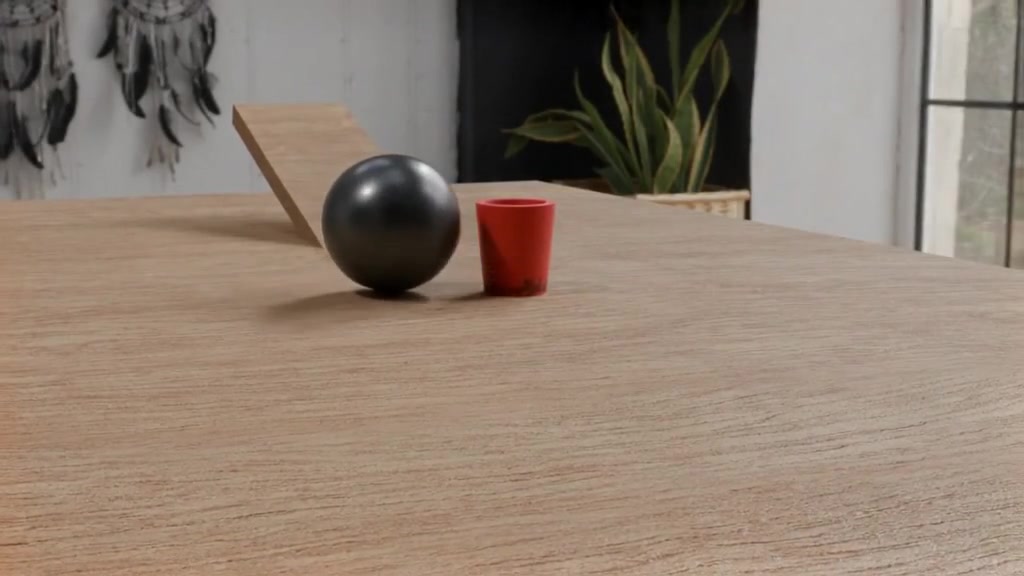}
        \includegraphics[width=0.16\textwidth]{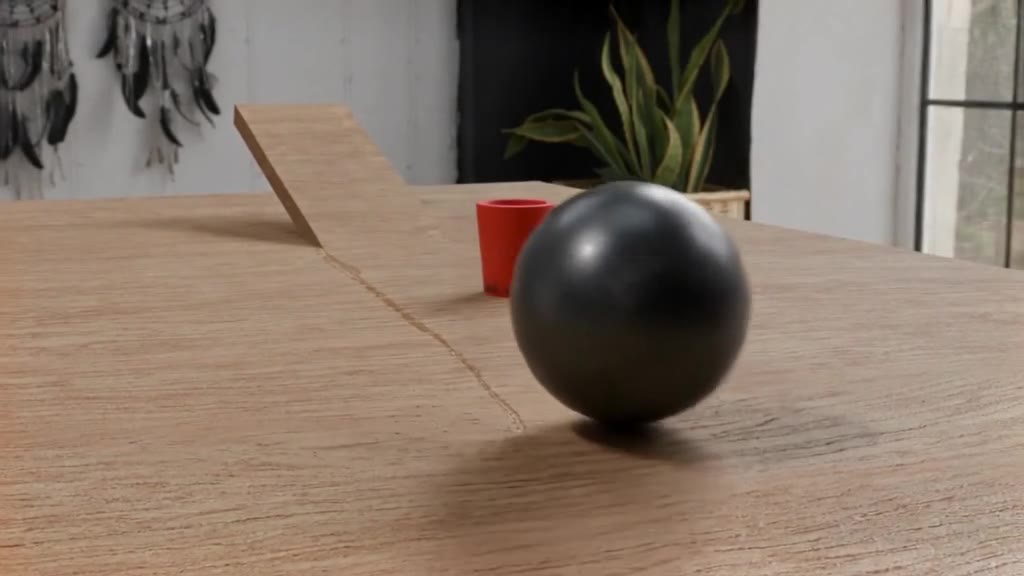}
        \includegraphics[width=0.16\textwidth]{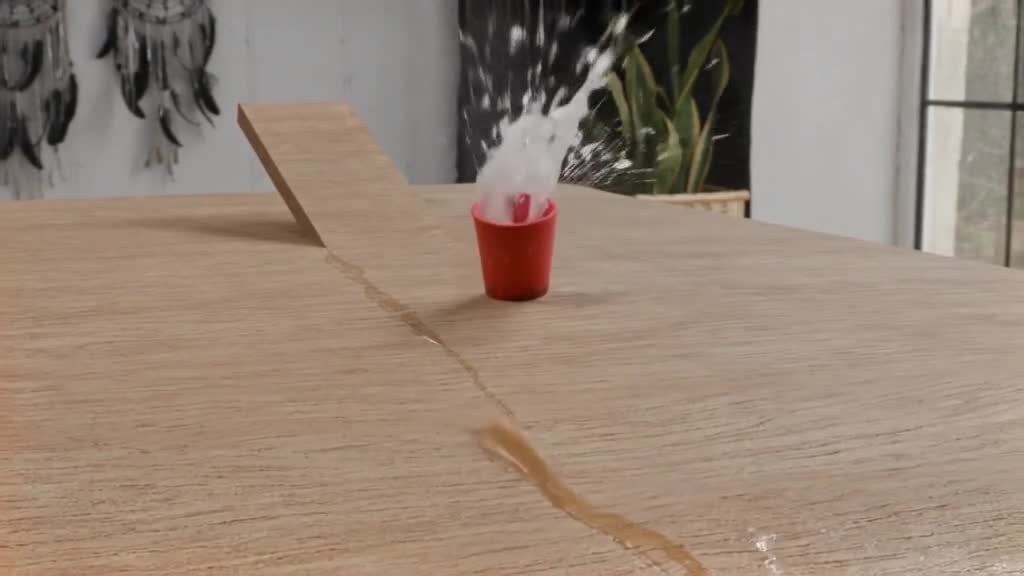}
        \includegraphics[width=0.16\textwidth]{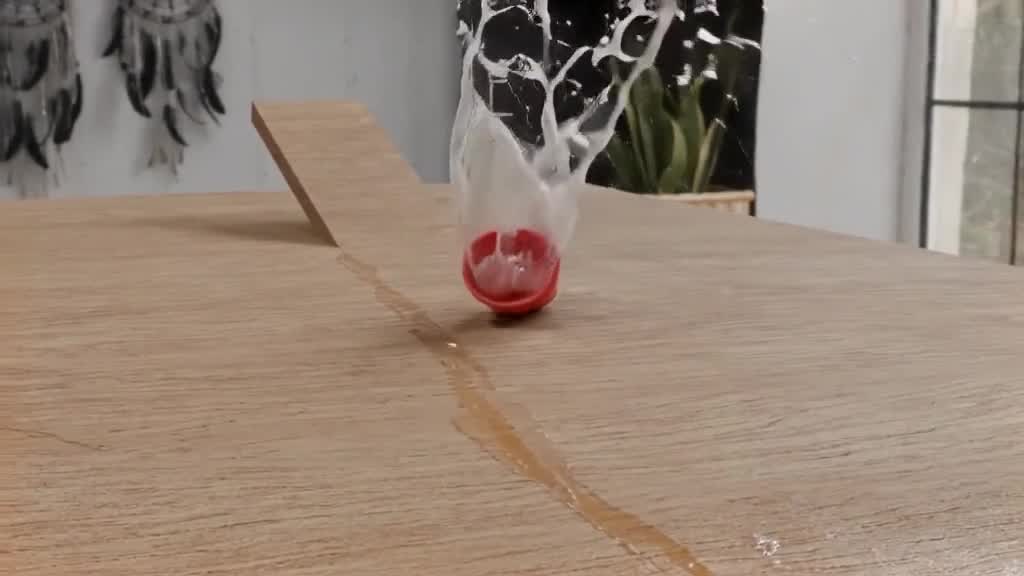}
        \caption{Diffusion-based video generation with fixed initial frame and text guidance. }
        \label{fig:cup_diff}
    \end{subfigure}

    \caption{Comparison between physical simulation and visual diffusion model in a ball and cup collision scenario, the simulation captures both rigid body interaction, as well as fluid behavior through particle dynamics. The diffusion model is prompted with \texttt{Continue from this frame, ball rolling down ramp, accelerating naturally while staying in contact with the surface, transfers momentum into the cup filled with water, knocks it over, causing water to spill onto the surface}.}
    \label{fig:diffusion_cup}
\end{figure}

\begin{figure}[t]
    \centering
    \begin{subfigure}{\textwidth}
        \centering
        \includegraphics[width=0.16\textwidth]{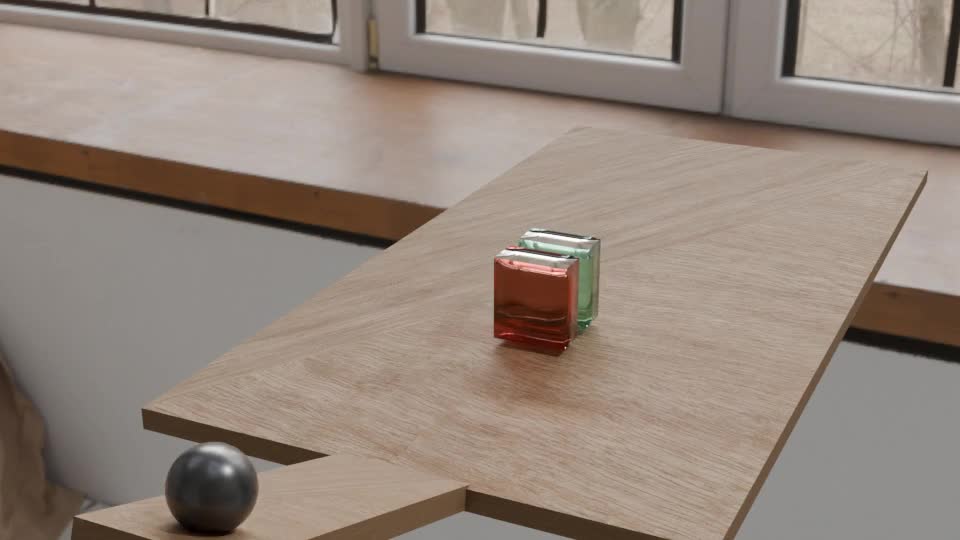}
        \includegraphics[width=0.16\textwidth]{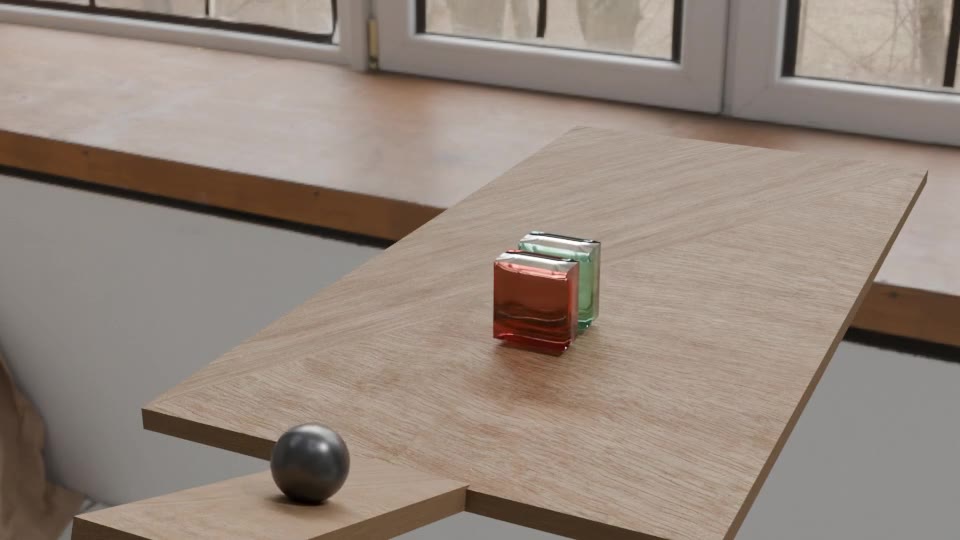}
        \includegraphics[width=0.16\textwidth]{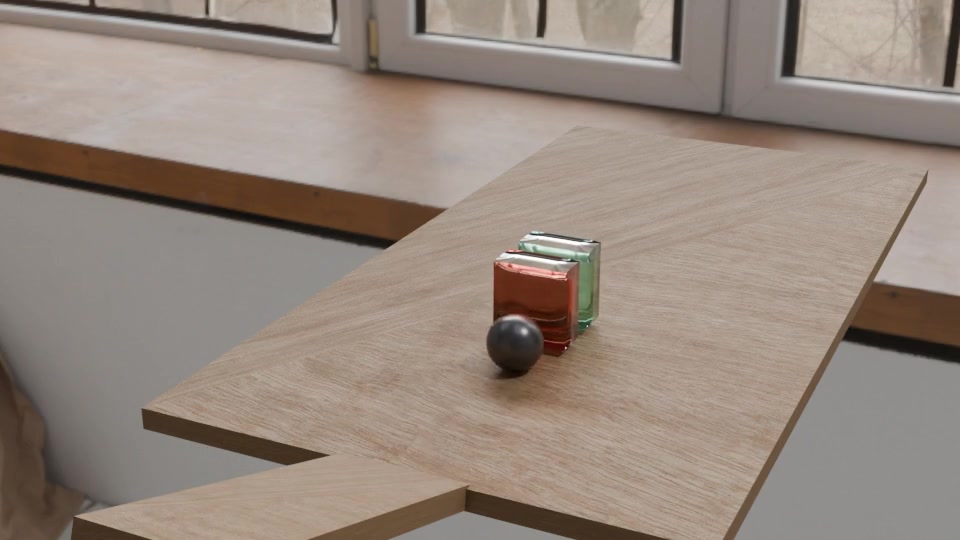}
        \includegraphics[width=0.16\textwidth]{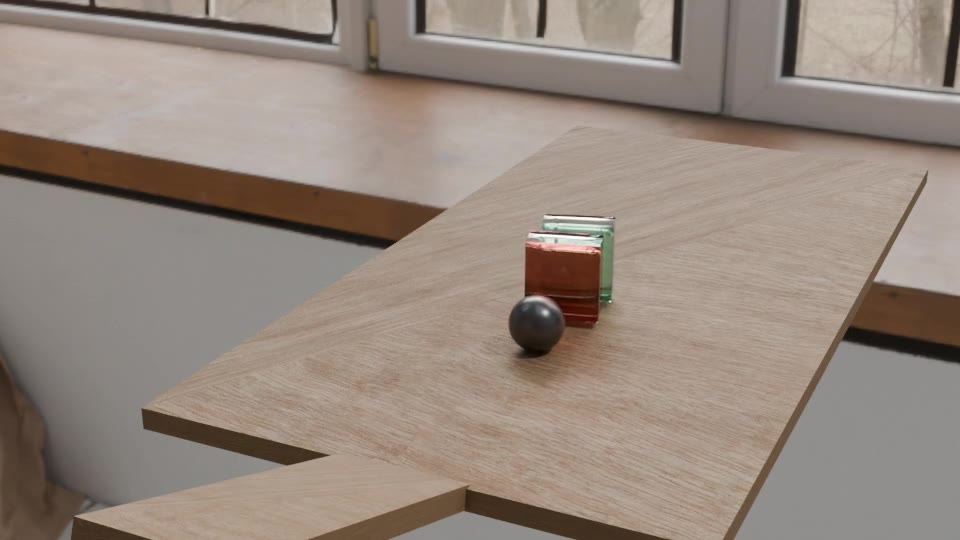}
        \includegraphics[width=0.16\textwidth]{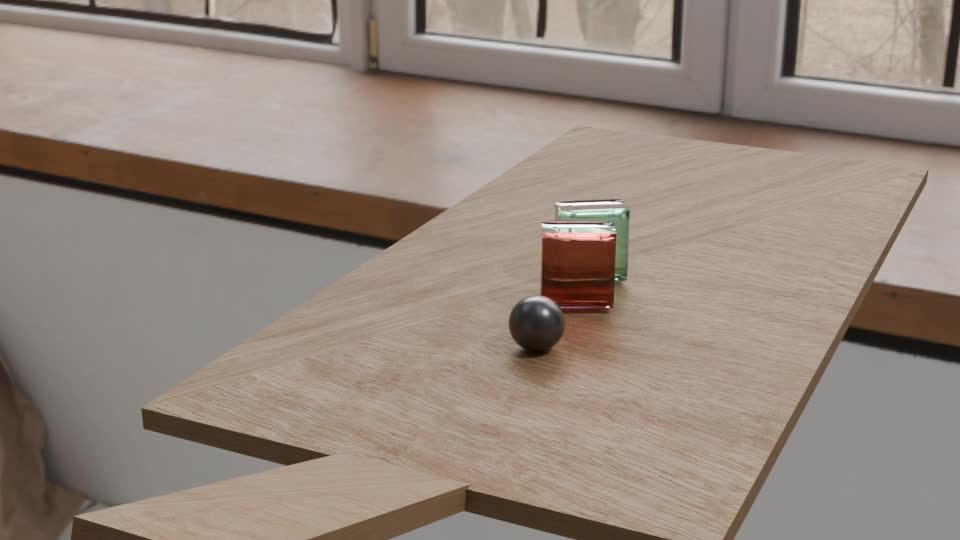}
        \includegraphics[width=0.16\textwidth]{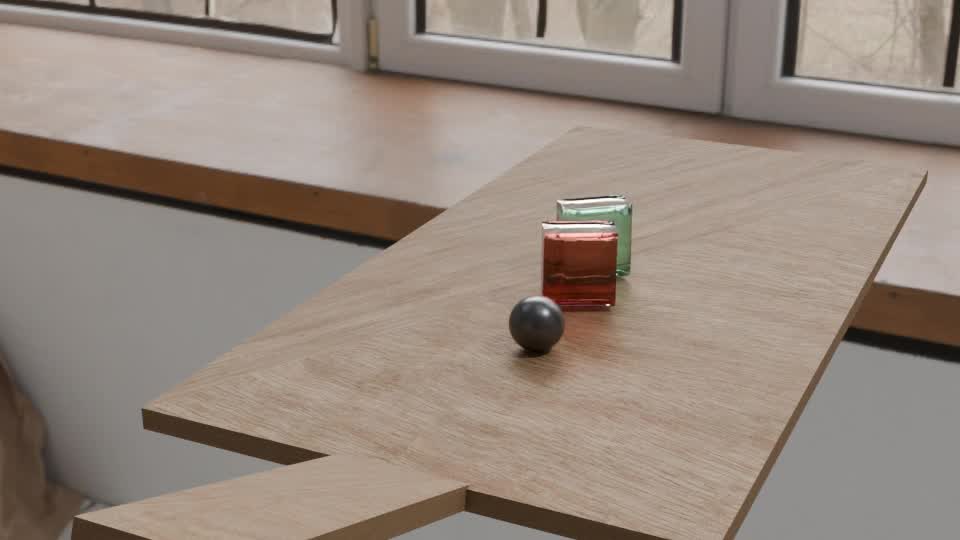}
        \caption{Physical simulation plus rendering.}
        \label{fig:jelly_2_demo}
    \end{subfigure}
    \hfill
    \begin{subfigure}{\textwidth}
        \centering
        \includegraphics[width=0.16\textwidth]{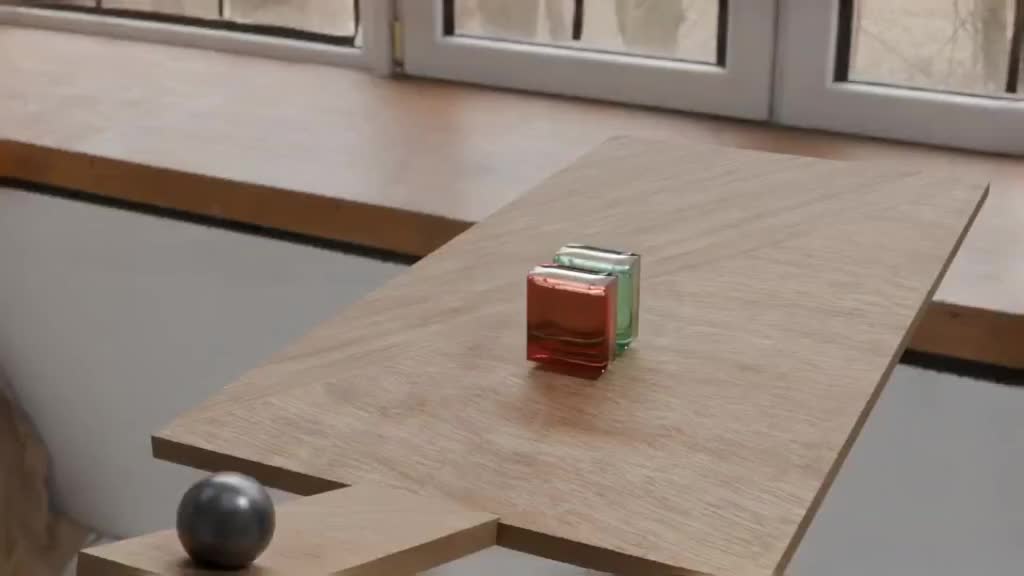}
        \includegraphics[width=0.16\textwidth]{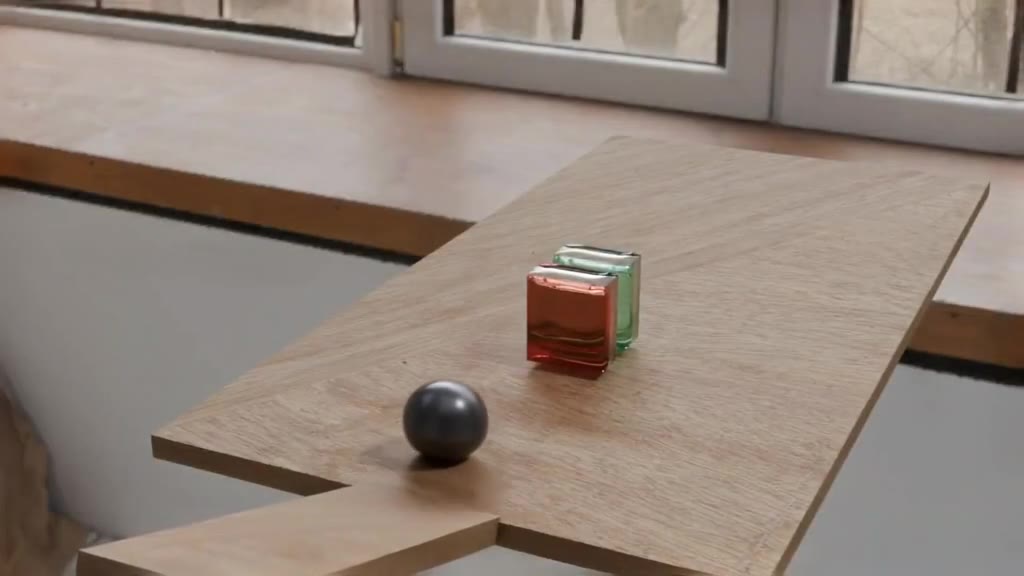}
        \includegraphics[width=0.16\textwidth]{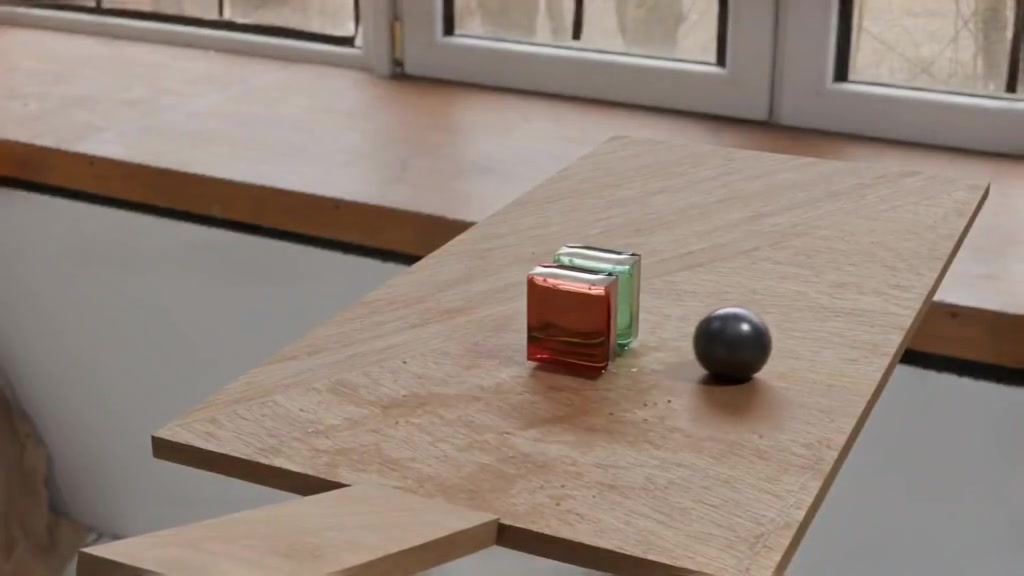}
        \includegraphics[width=0.16\textwidth]{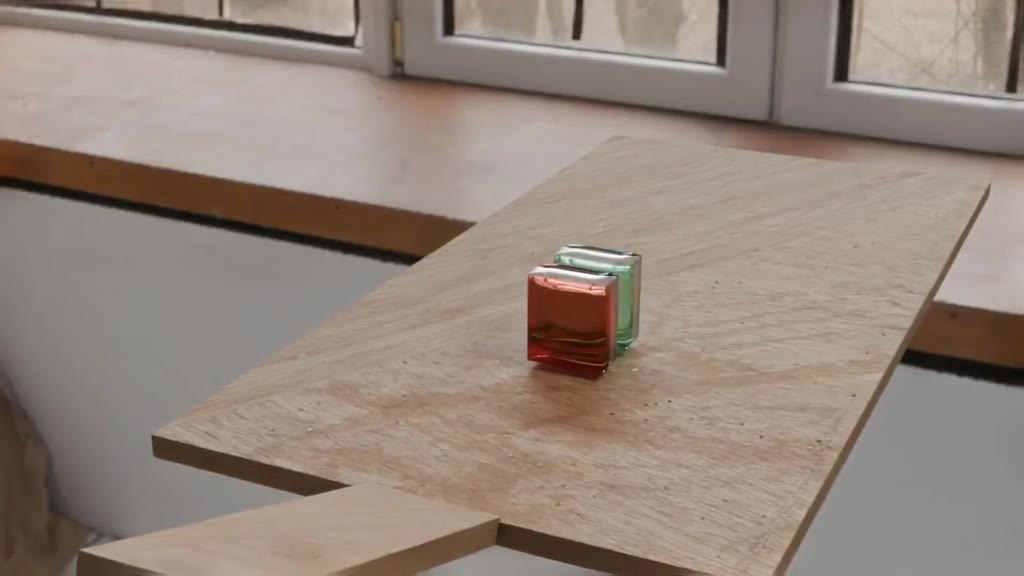}
        \includegraphics[width=0.16\textwidth]{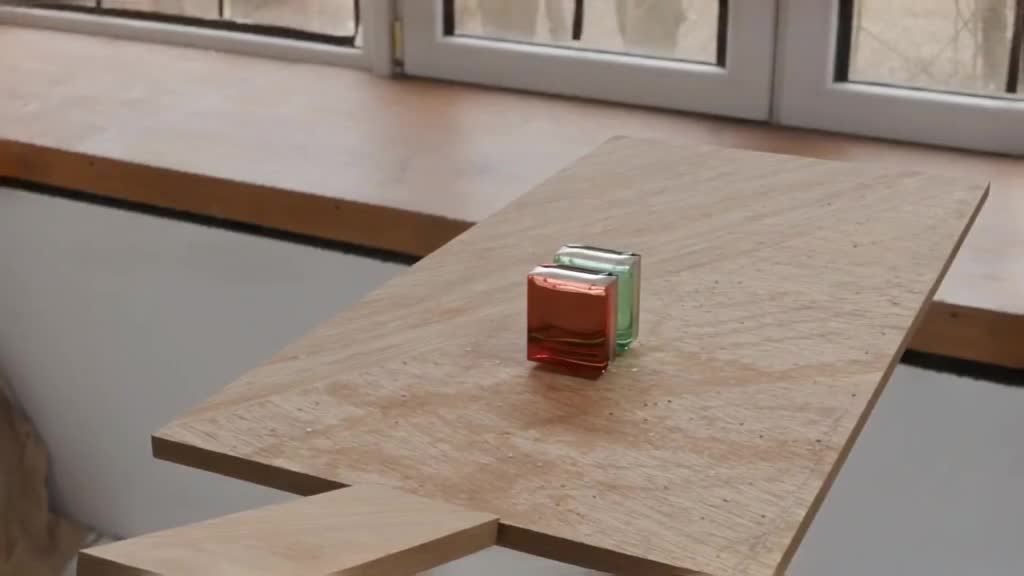}
        \includegraphics[width=0.16\textwidth]{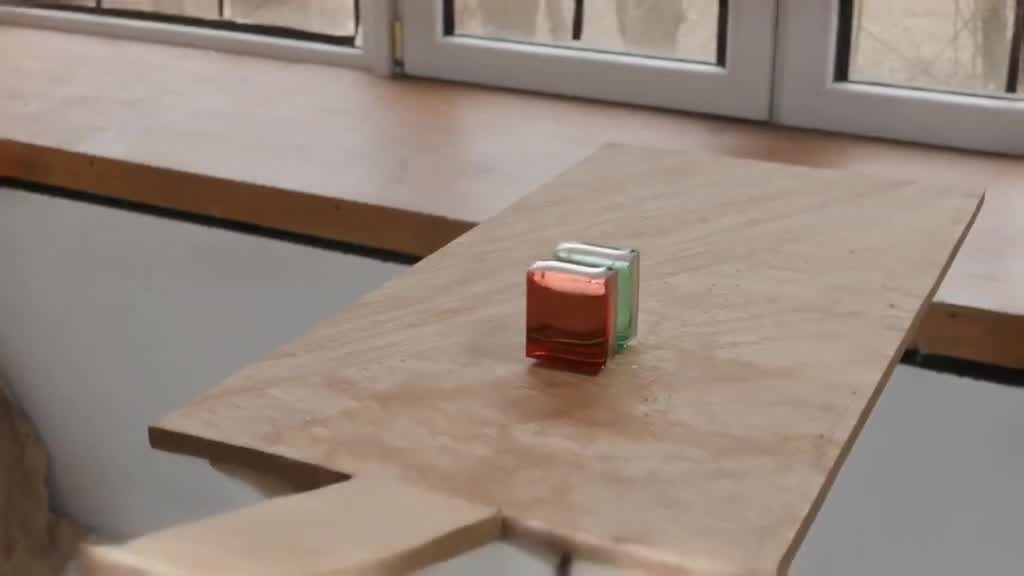}
        \caption{Diffusion-based video generation with fixed initial frame and text guidance. }
        \label{fig:jelly_2_diff}
    \end{subfigure}

    \caption{Comparison between physical simulation and visual diffusion model in a ball and two jelly blocks collision scenario, the physicaly simulation correctly handles interaction between plastic and elastic materials. The diffusion model is prompted with \texttt{Continue from this frame, ball rolling down ramp, accelerating naturally while staying in contact with the surface, and transfers momentum into the jelly blocks on impact}.}
    \label{fig:diffusion_jelly_2}
\end{figure}

We illustrate this limitation through two representative examples. In each case, we provide a fixed initial state and a carefully specified prompt describing the intended physical process. We use physics-based simulation with rendering as a comparative baseline for both examples. To ensure a fair comparison, we keep key hyperparameters such as the total number of frames, aspect ratio, and resolution to be fixed across all runs. Additionally, following the \texttt{LTX-2} guidelines, we include negative prompts (e.g., unrealistic physics'', extra balls'') to further constrain the diffusion model outputs.

We display the results in~\Cref{fig:diffusion_cup} and~\Cref{fig:diffusion_jelly_2}. Despite detailed instructions specifying the desired physical behavior, the generated videos fail to remain even close to the true dynamics. In~\Cref{fig:diffusion_cup}, contact does not yield correct motion for either the ball or the cup, and the fluid behavior is unstable and inconsistent. Similarly, in~\Cref{fig:diffusion_jelly_2}, we do not observe physically correct contact between objects. We also notice that despite being explicit in our instruction, probabilistic models like text conditioned diffusion models can often fail to follow he exact prompts. 

Across both examples, the failure modes are systematic rather than incidental. They stem from the absence of explicit physical constraints in the generative process. As a result, despite improvements in visual fidelity, even state-of-the-art visual foundation models cannot reliably replace even simple physics-based simulators in settings where accurate dynamics are required. As noted in~\cite{o2025vdaworld}, direct pixel prediction is insufficient and unreliable for the accuracy requirements of world models.

\subsection{Failure Modes of Vision-Based World Models in Control}
\label{subsec:vjepa_control_results}

We next evaluate whether a vision-based world model can be used as a control-oriented simulator in contact-rich manipulation. 
Unlike the diffusion models considered in~\Cref{subsec:diffusion_results}, \texttt{V-JEPA 2} does not directly synthesize future image frames. 
Instead, the action-conditioned model predicts future latent representations conditioned on the current visual state, robot pose, and candidate action sequences. 
We then use MPC to select the action trajectory whose predicted latent state is closest to the goal representation. 
The selected trajectory is subsequently rendered through the Newton physics simulator in order to inspect the physical outcome.

This setting is more favorable than direct pixel generation because the final visualization is still produced by a physics engine. 
However, it exposes a different and more control-specific failure mode. The planner optimizes for visual latent similarity, not for physical feasibility. 
The model is not explicitly constrained by contact mechanics, friction cones, object mass, material density, torque limits, or the force histories required to move an object. 
Consequently, a trajectory that appears promising under the learned latent objective may still fail when executed as a physical interaction.

\begin{figure}[t]
    \centering

    \begin{subfigure}{\textwidth}
        \centering
        \includegraphics[width=0.16\textwidth]{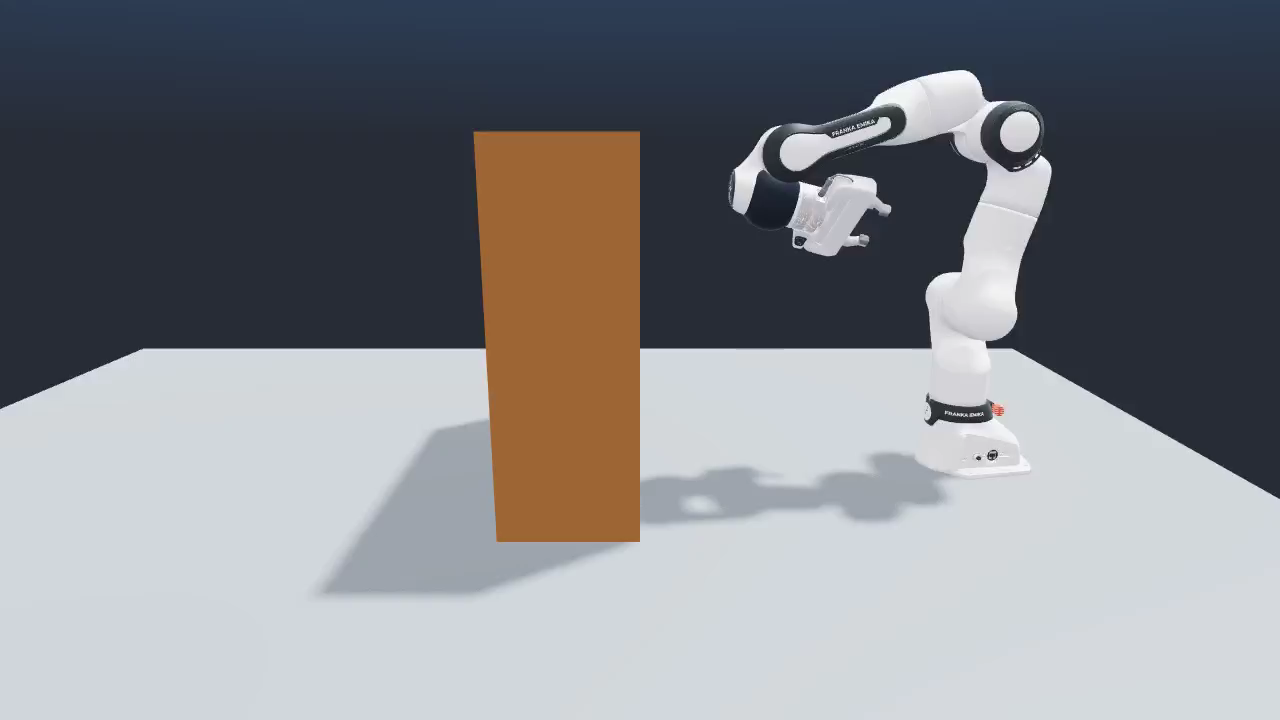}
        \includegraphics[width=0.16\textwidth]{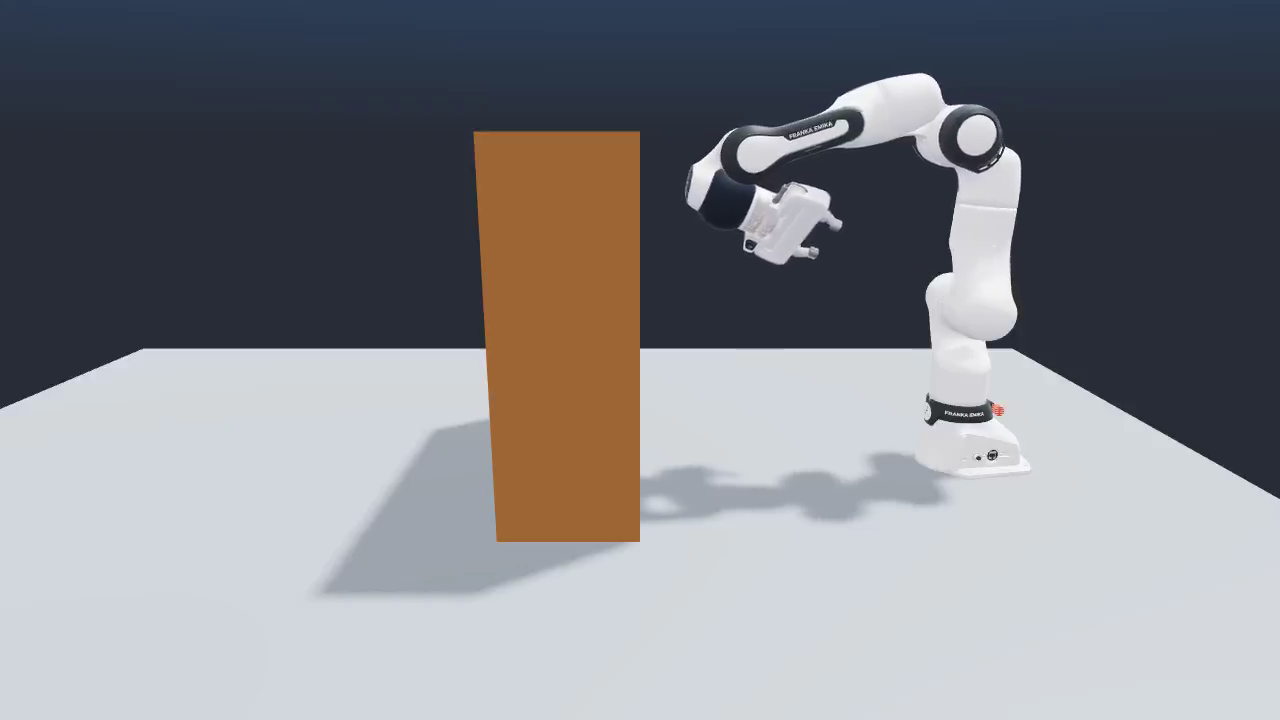}
        \includegraphics[width=0.16\textwidth]{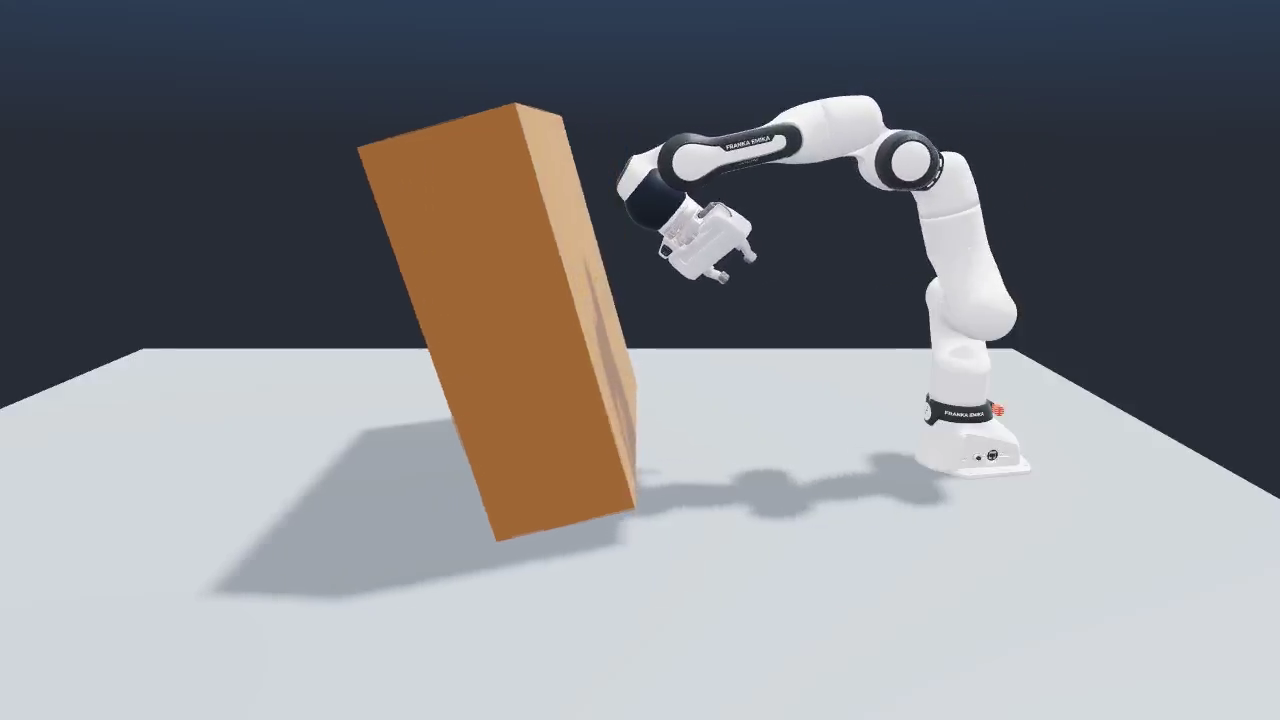}
        \includegraphics[width=0.16\textwidth]{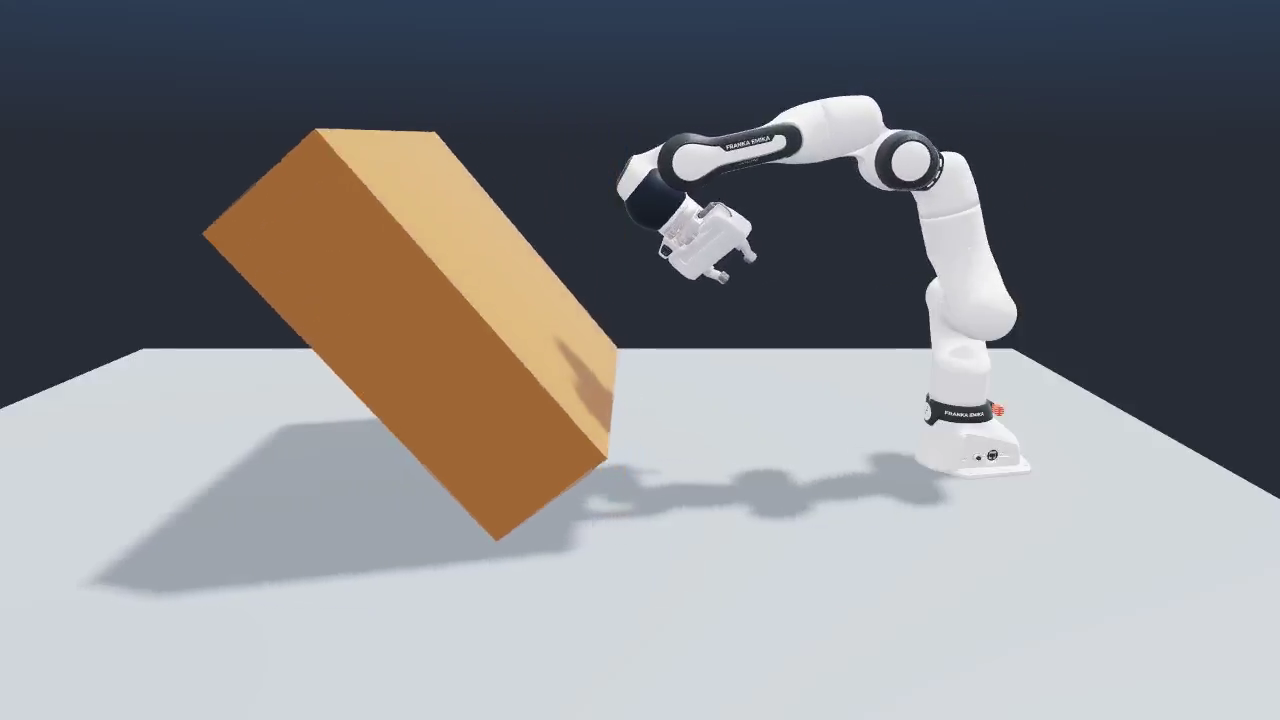}
        \includegraphics[width=0.16\textwidth]{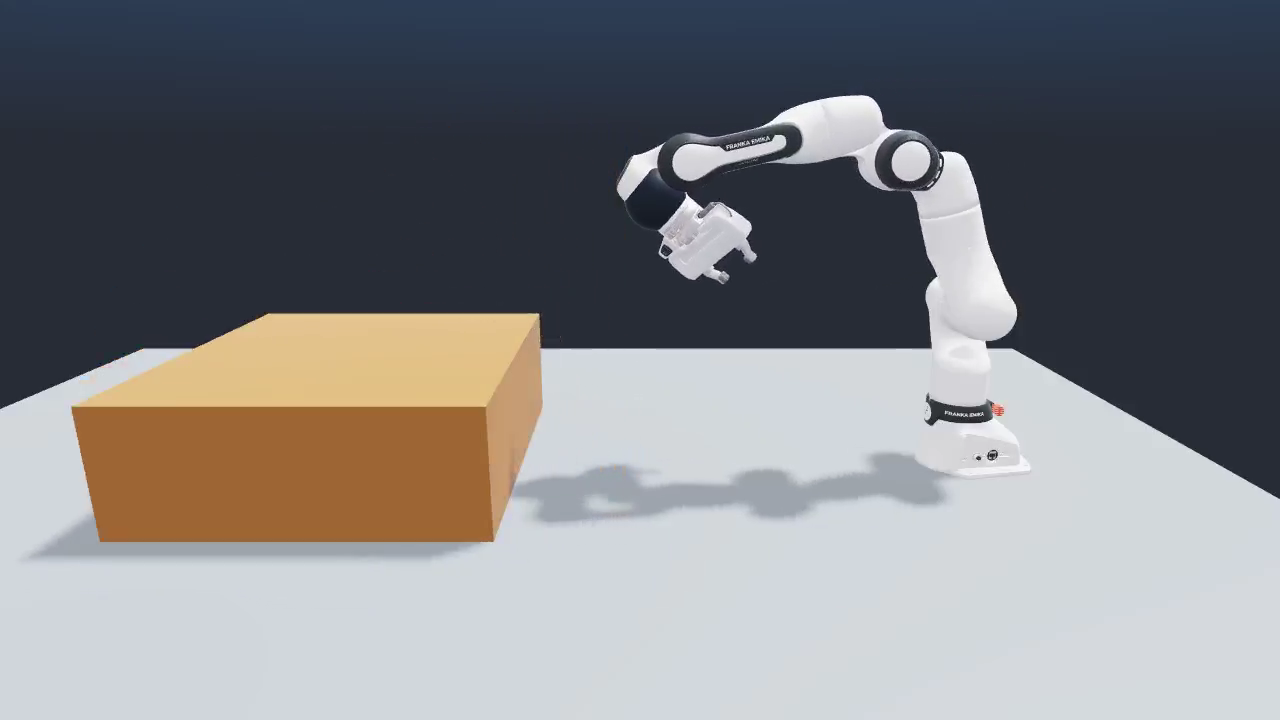}
        \includegraphics[width=0.16\textwidth]{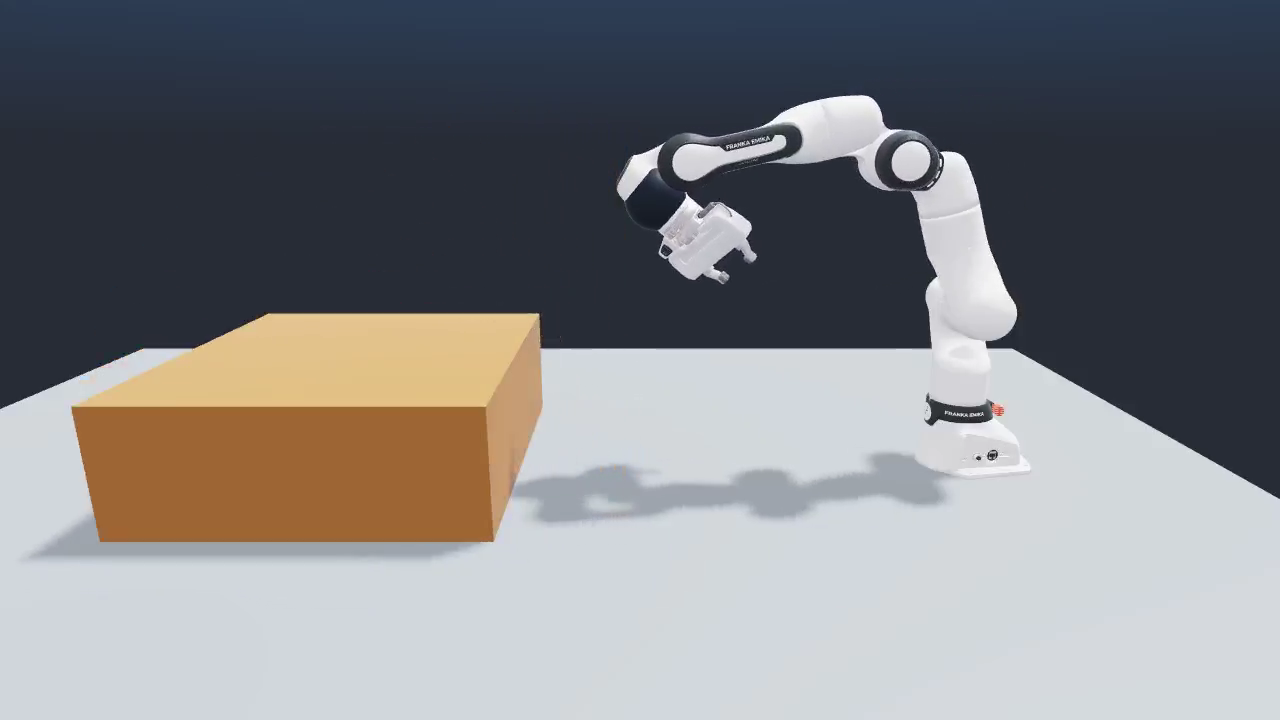}
        \caption{Ground-truth physical rollout of a wooden wall.}
        \label{fig:vjepa_high_push_gt}
    \end{subfigure}

    \vspace{0.5em}

    \begin{subfigure}{\textwidth}
        \centering
        \includegraphics[width=0.16\textwidth]{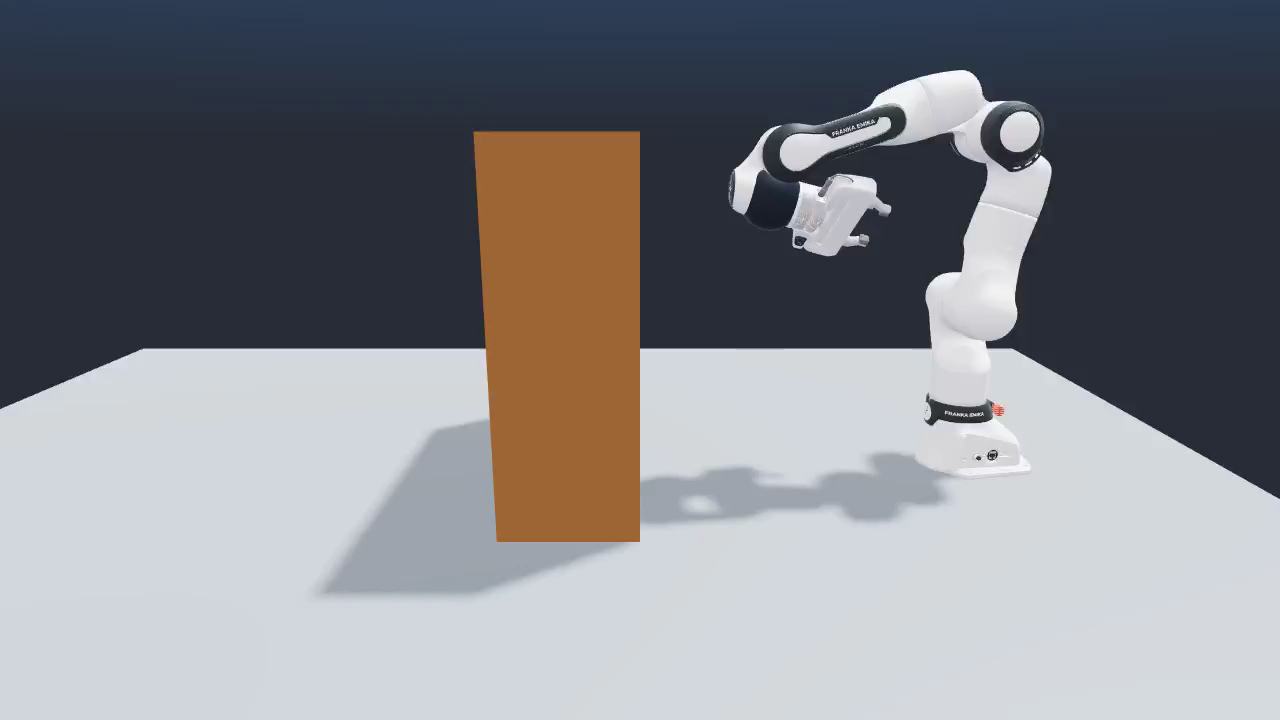}
        \includegraphics[width=0.16\textwidth]{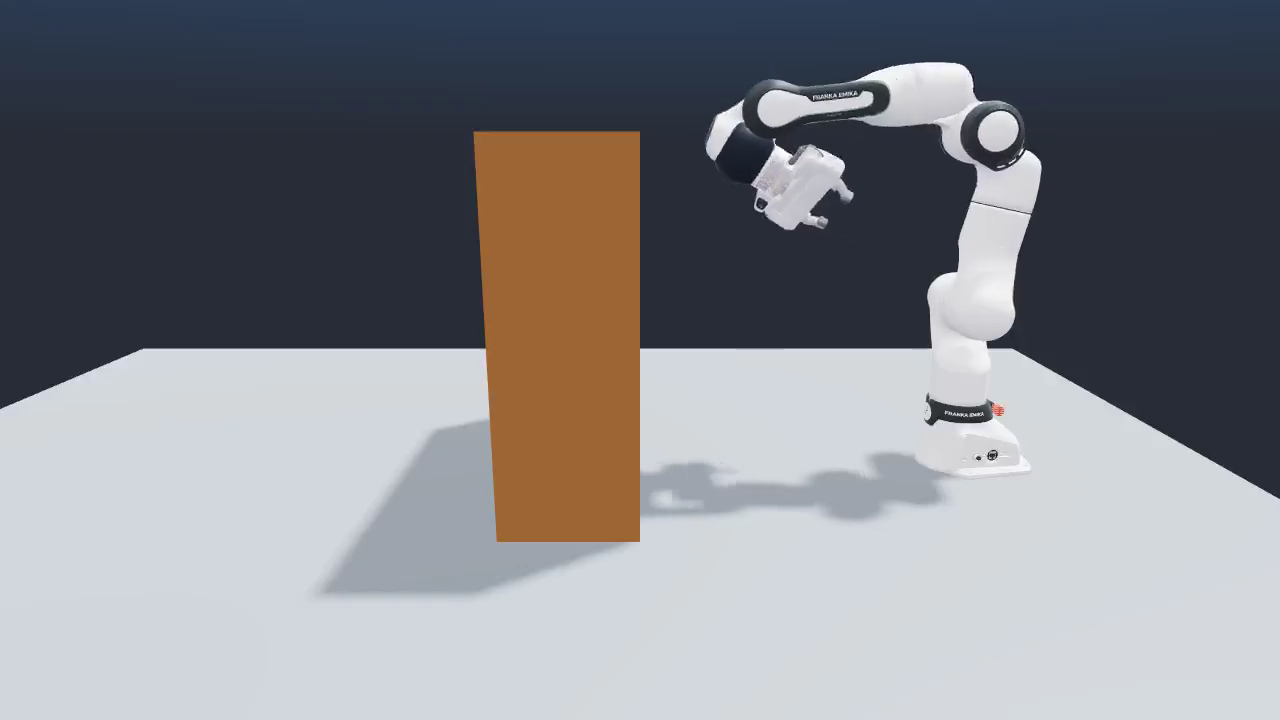}
        \includegraphics[width=0.16\textwidth]{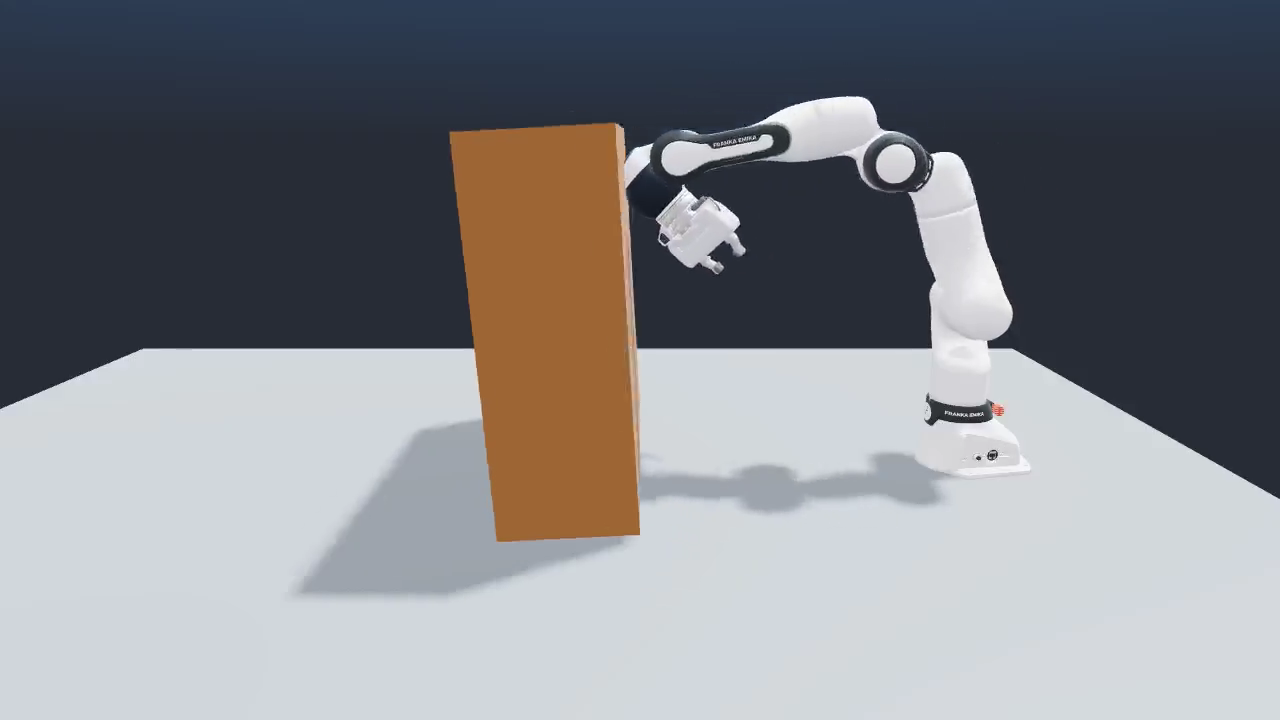}
        \includegraphics[width=0.16\textwidth]{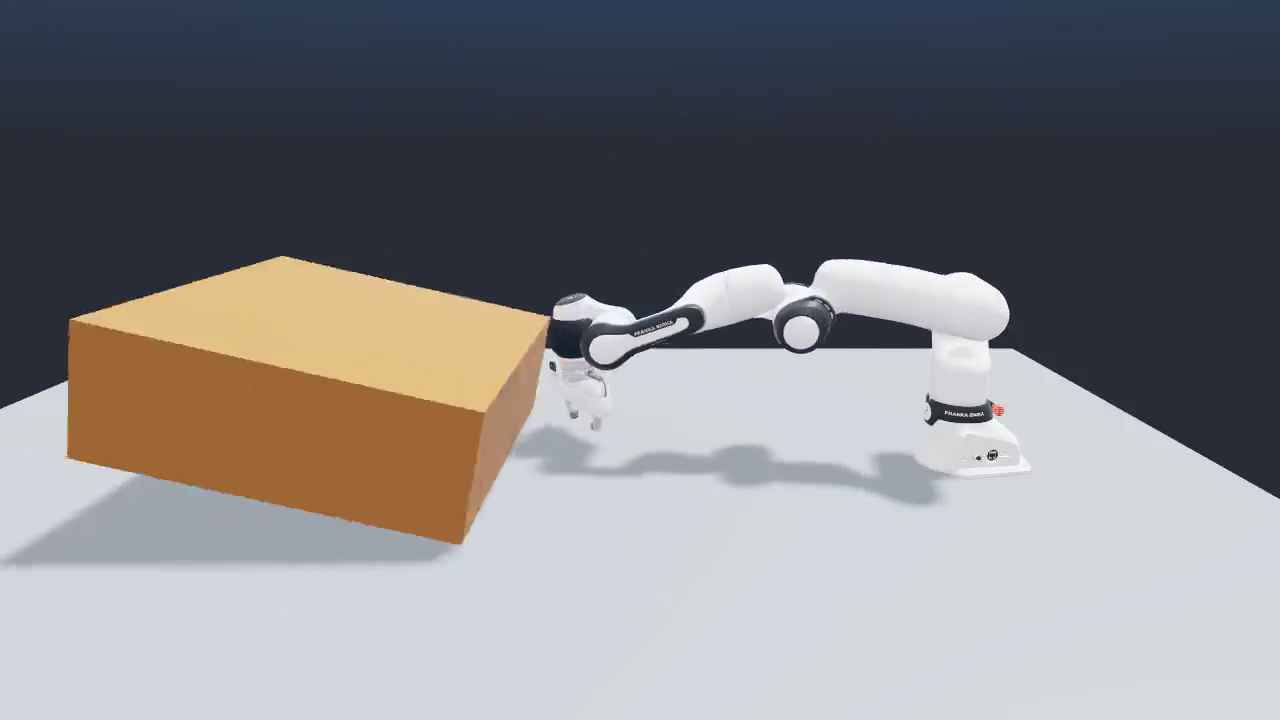}
        \includegraphics[width=0.16\textwidth]{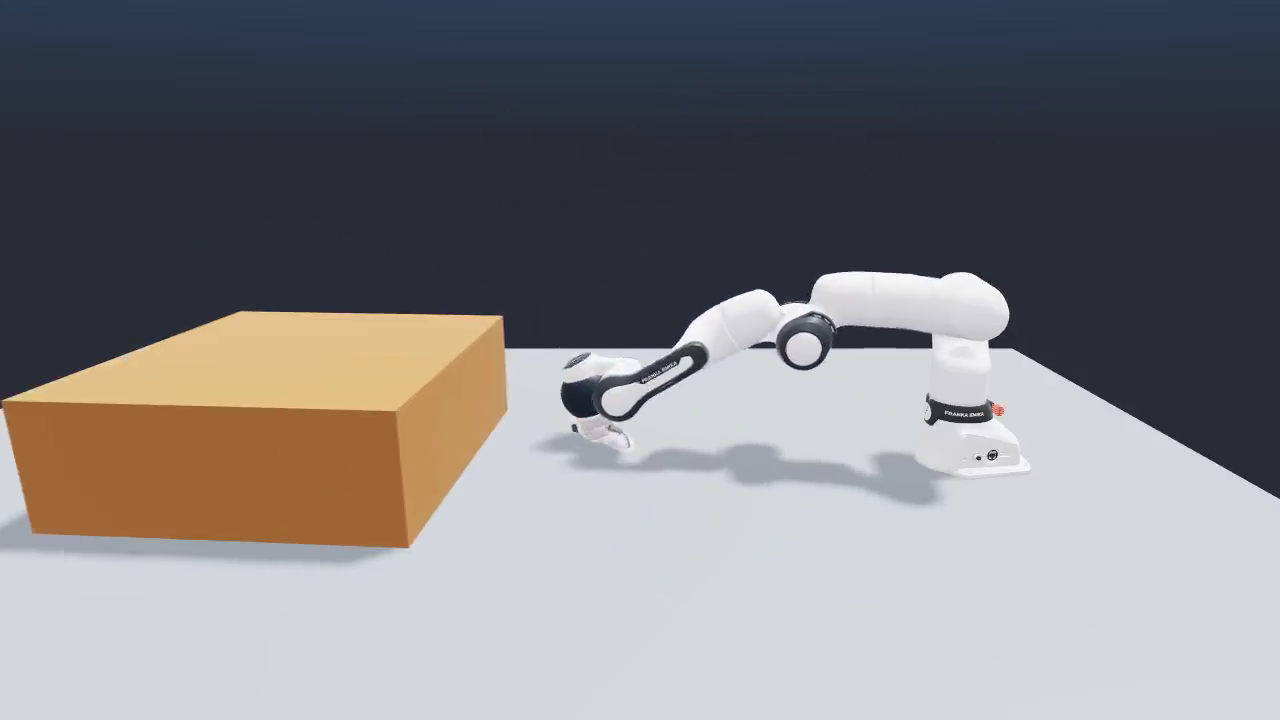}
        \includegraphics[width=0.16\textwidth]{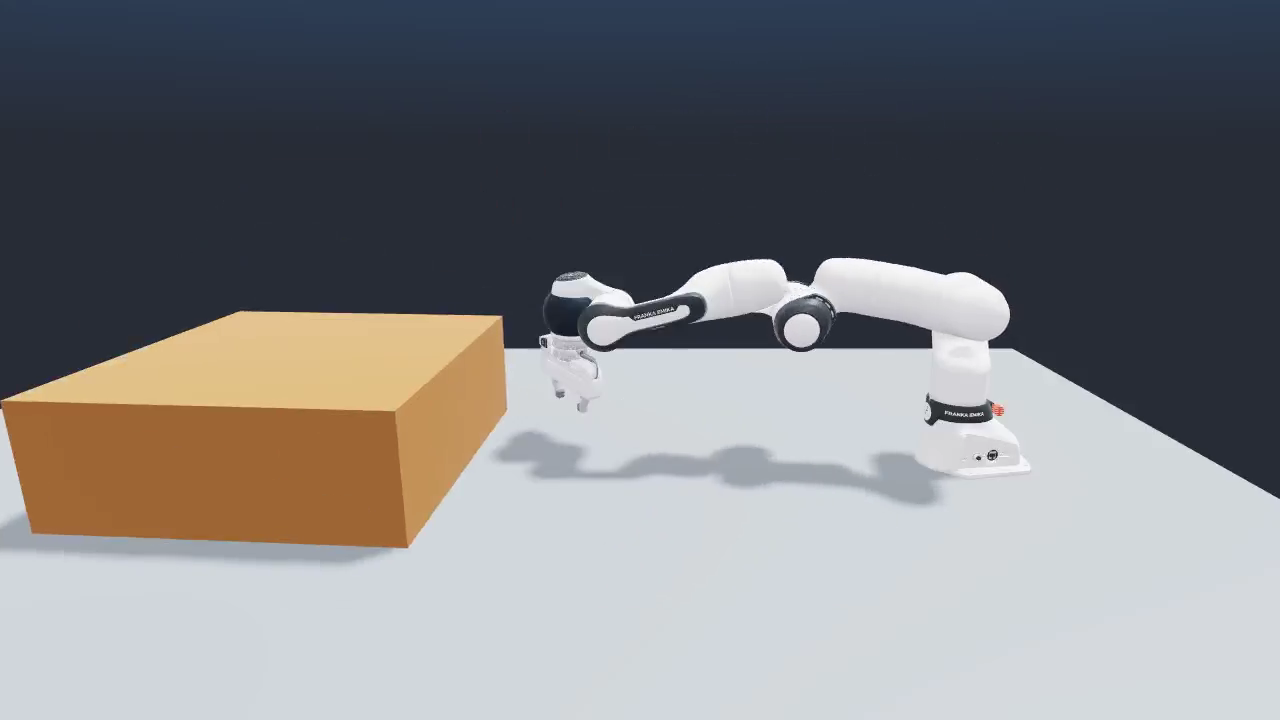}
        \caption{V-JEPA2 trajectory rendered in the simulation under high-push setting.}
        \label{fig:vjepa_high_push}
    \end{subfigure}
    
    \vspace{0.5em}

    \begin{subfigure}{\textwidth}
        \centering
        \includegraphics[width=0.16\textwidth]{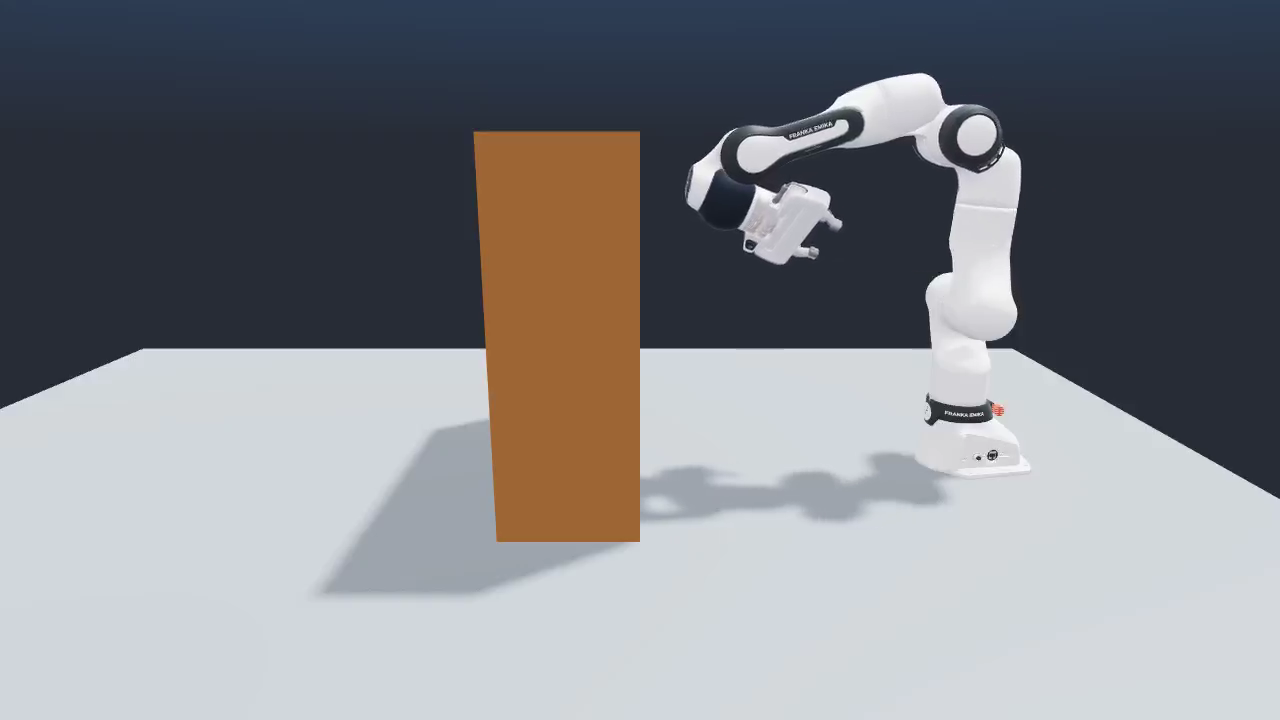}
        \includegraphics[width=0.16\textwidth]{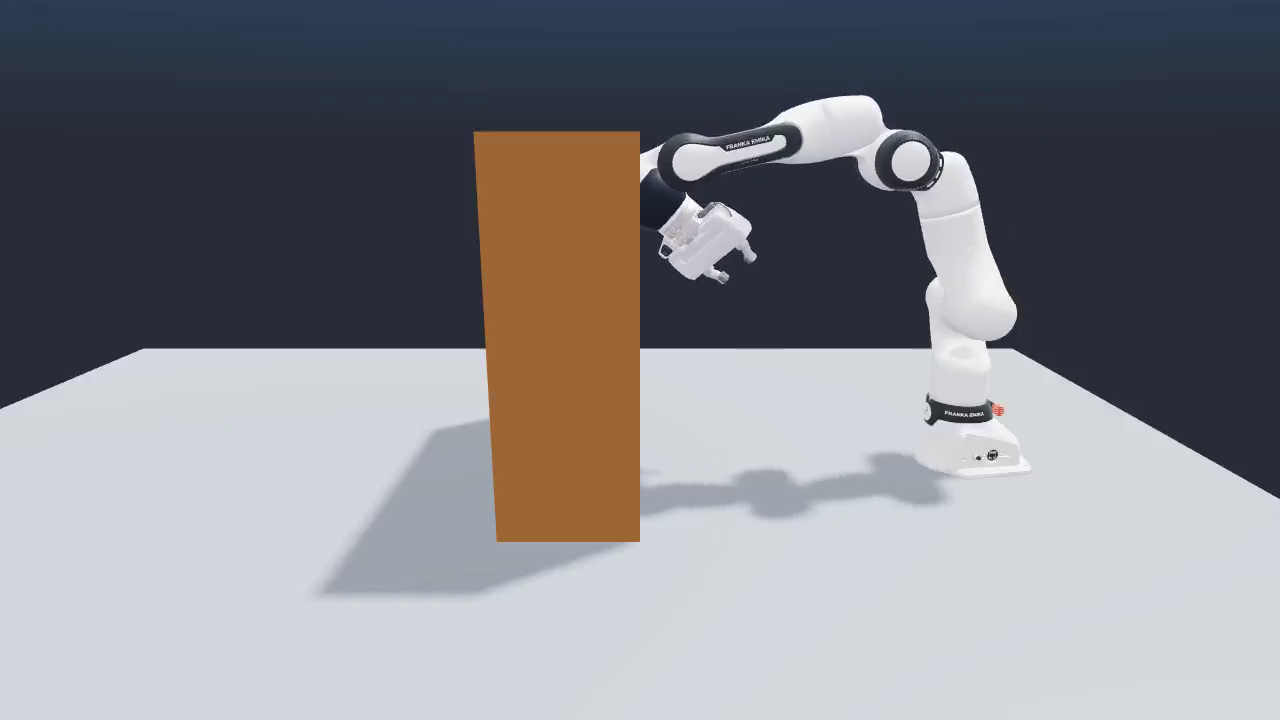}
        \includegraphics[width=0.16\textwidth]{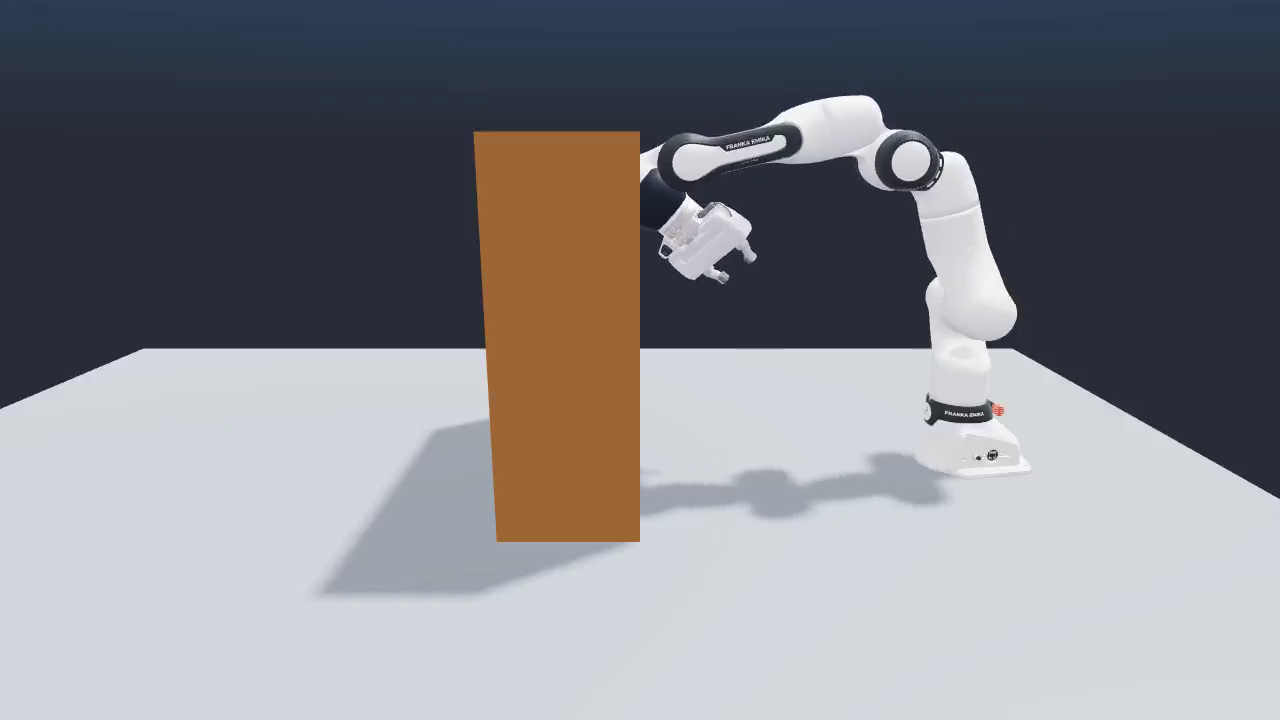}
        \includegraphics[width=0.16\textwidth]{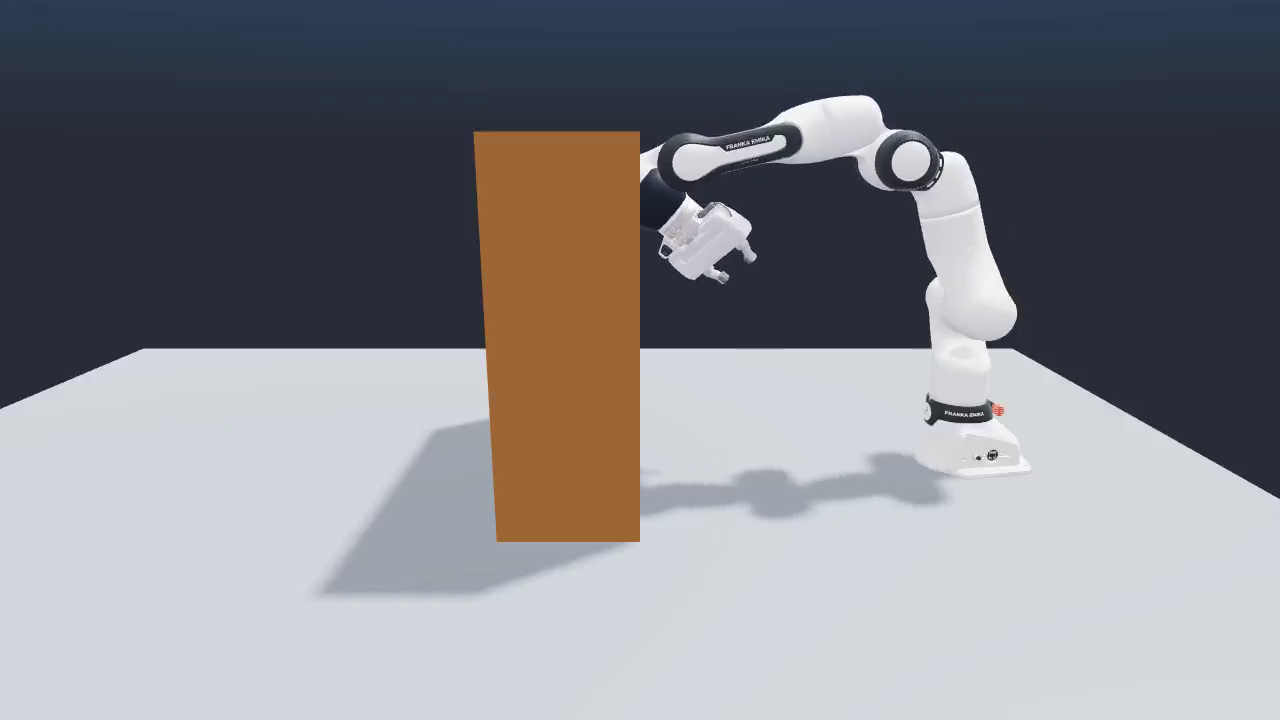}
        \includegraphics[width=0.16\textwidth]{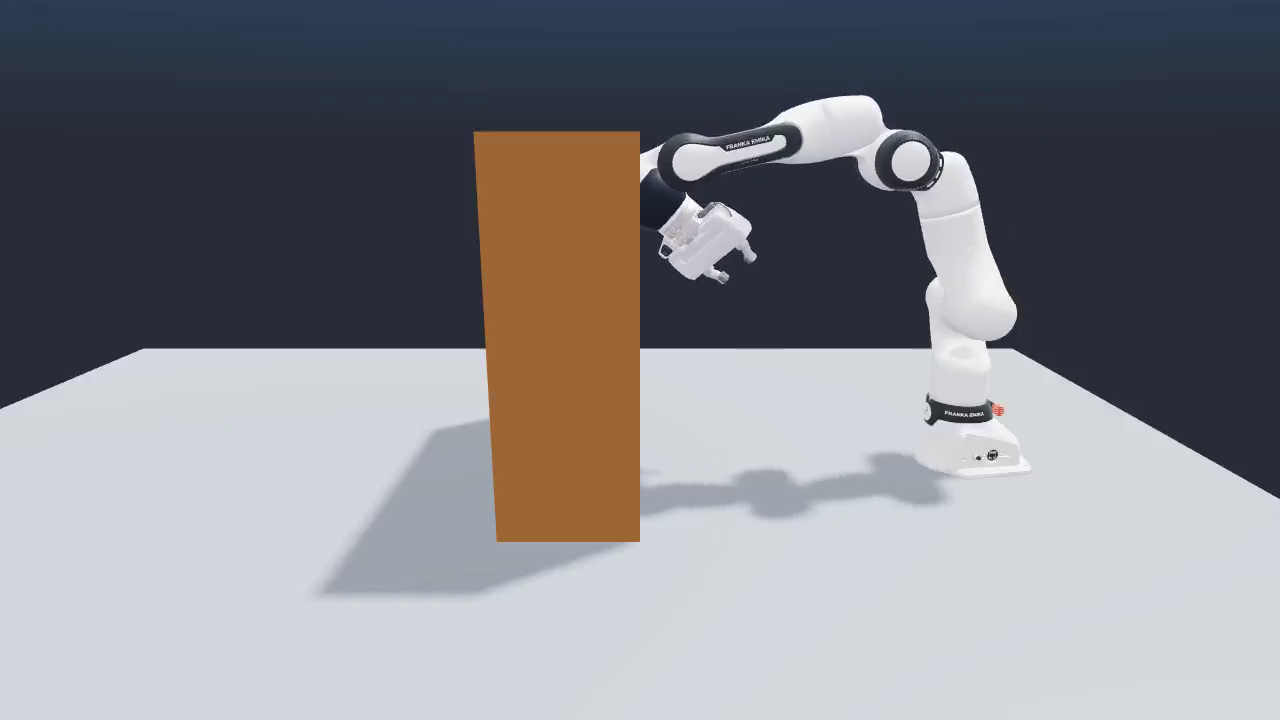}
        \includegraphics[width=0.16\textwidth]{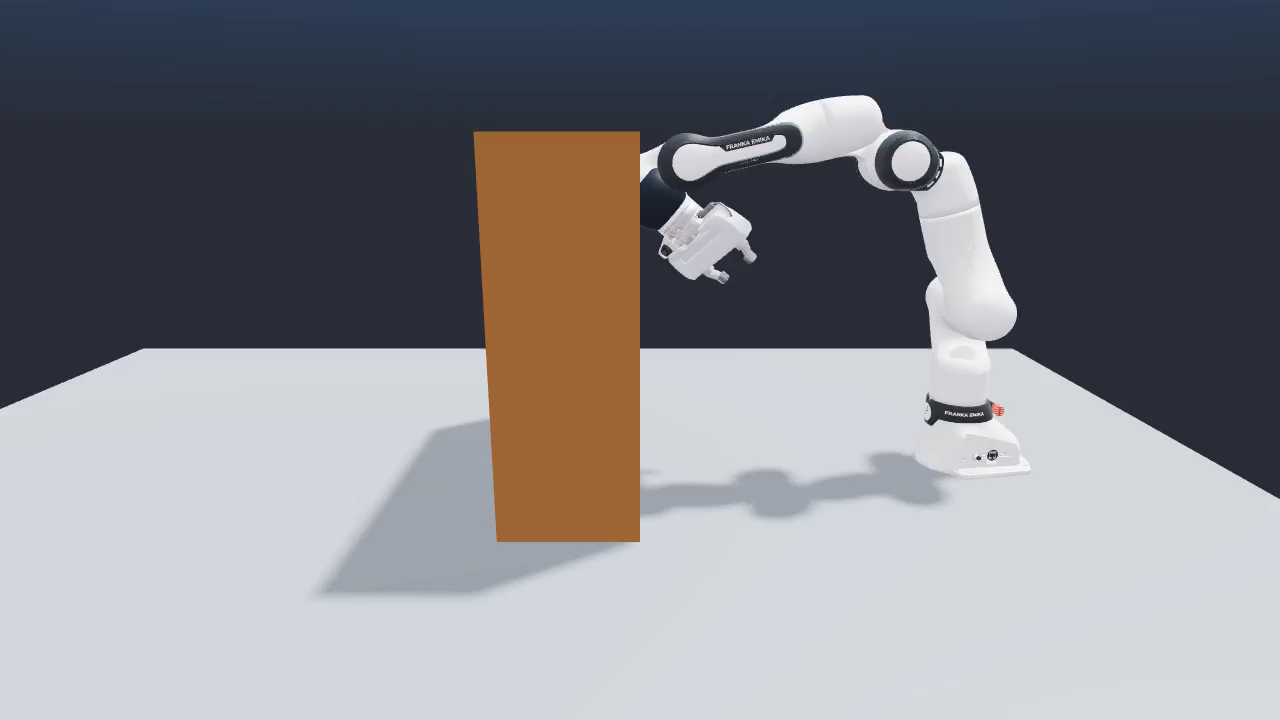}
        \caption{Ground-truth physical rollout of a concrete wall.}
        \label{fig:vjepa_concrete_gt}
    \end{subfigure}
    
    \vspace{0.5em}

    \begin{subfigure}{\textwidth}
        \centering
        \includegraphics[width=0.16\textwidth]{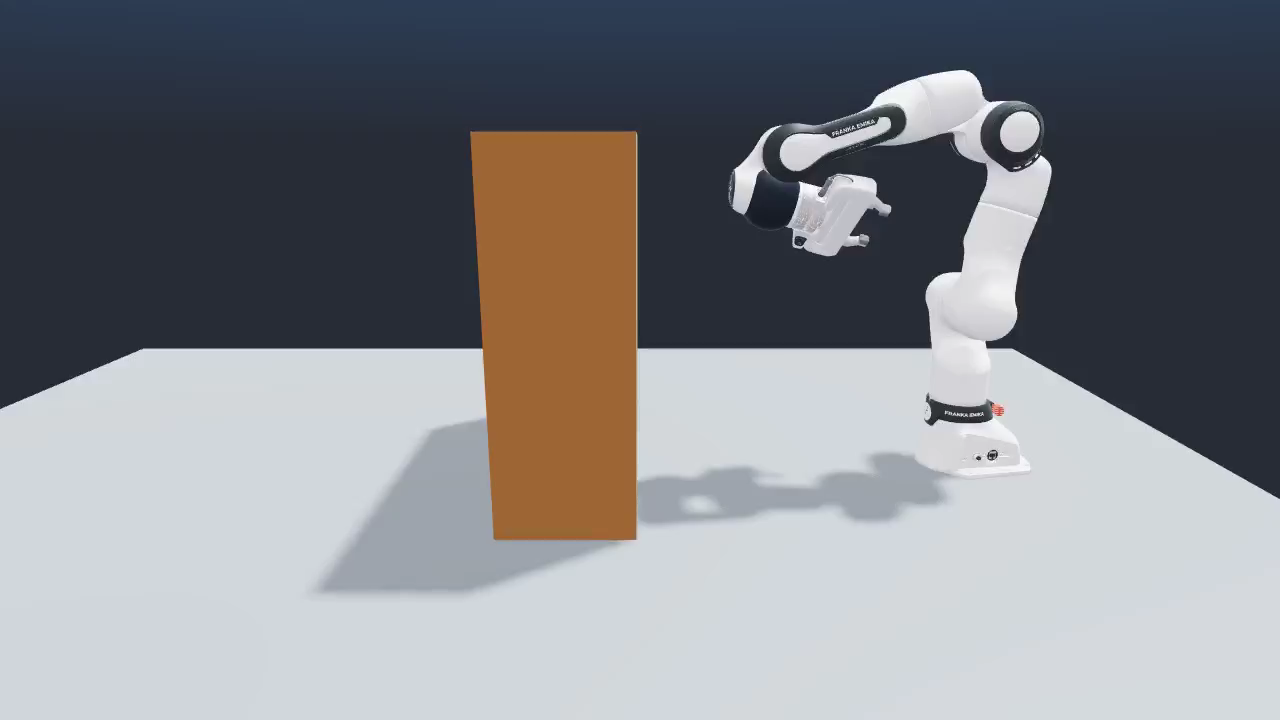}
        \includegraphics[width=0.16\textwidth]{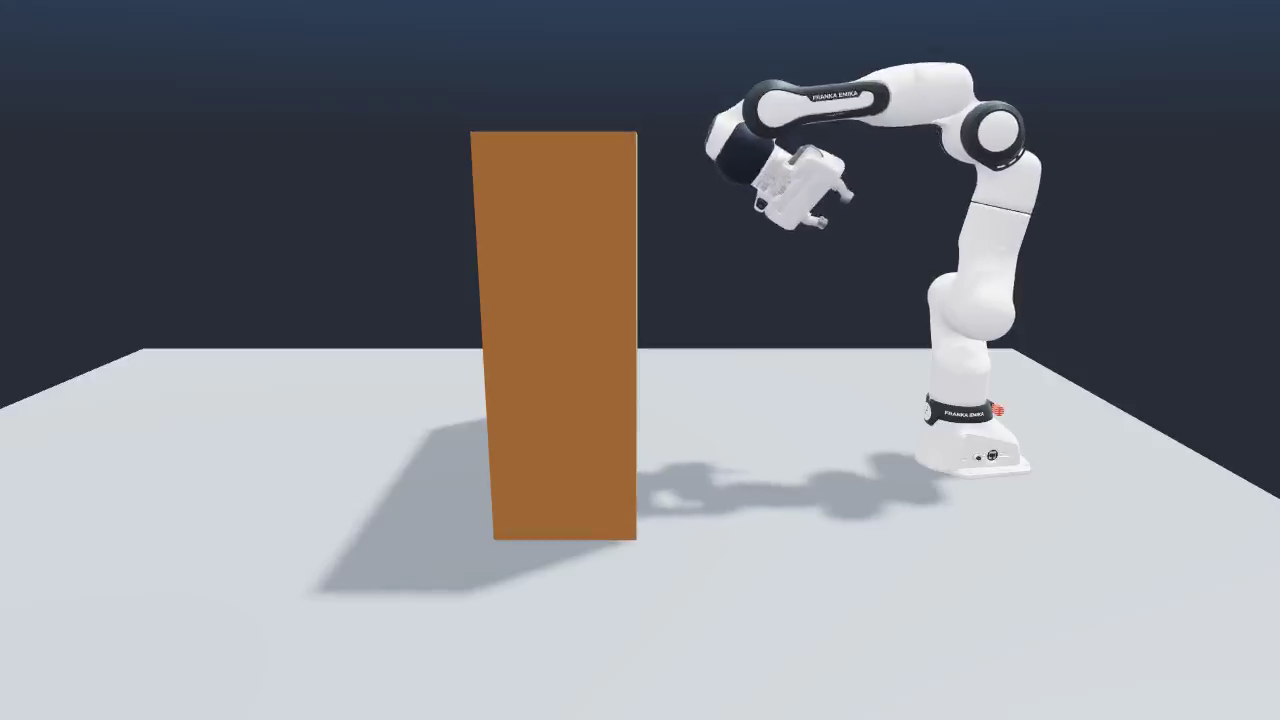}
        \includegraphics[width=0.16\textwidth]{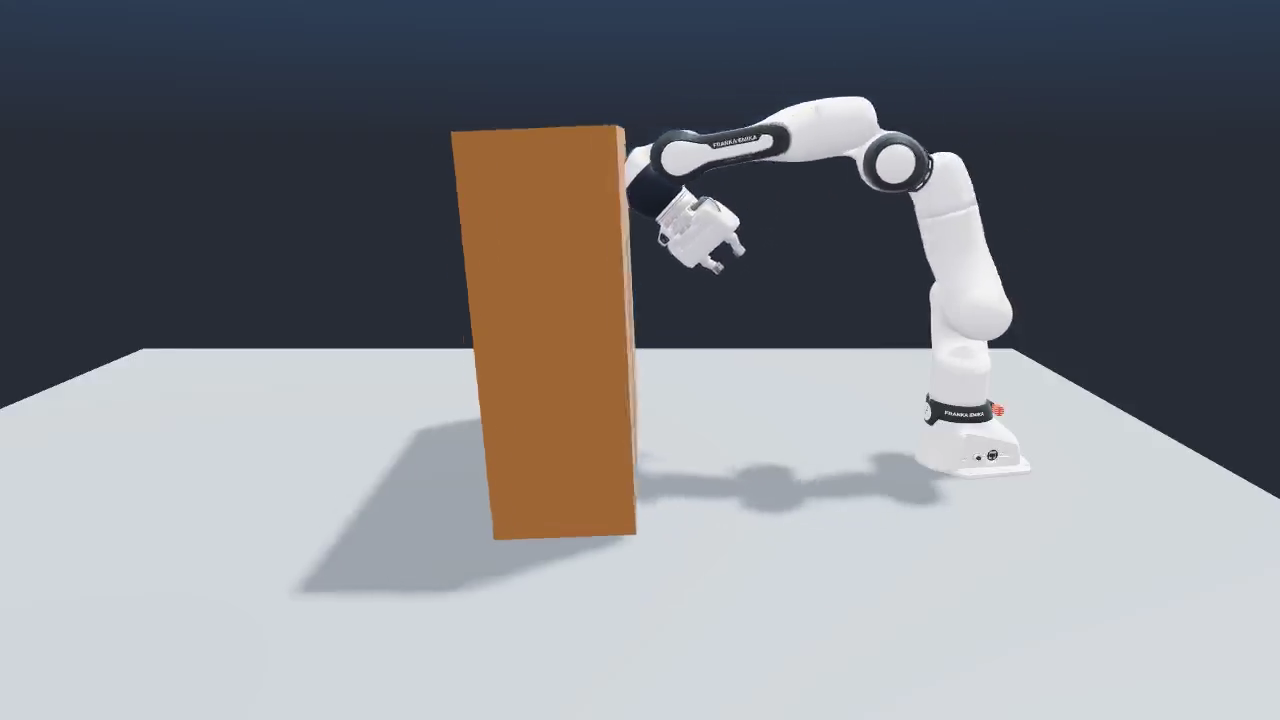}
        \includegraphics[width=0.16\textwidth]{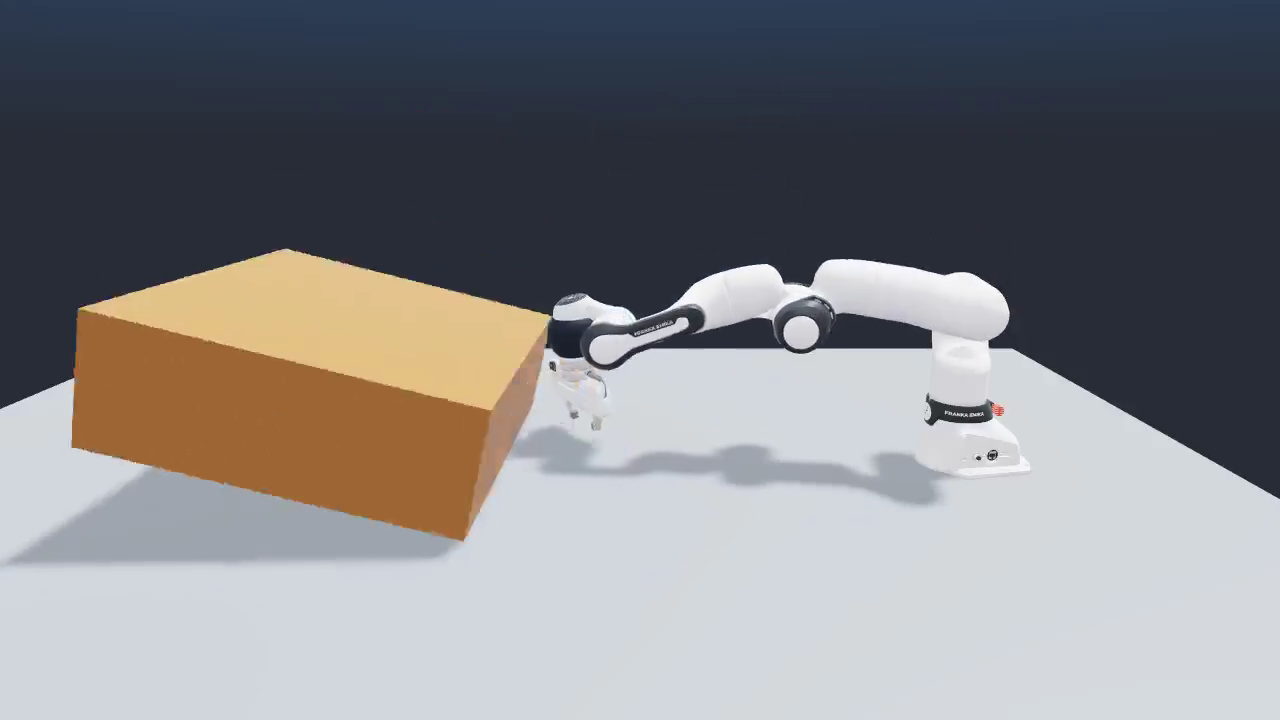}
        \includegraphics[width=0.16\textwidth]{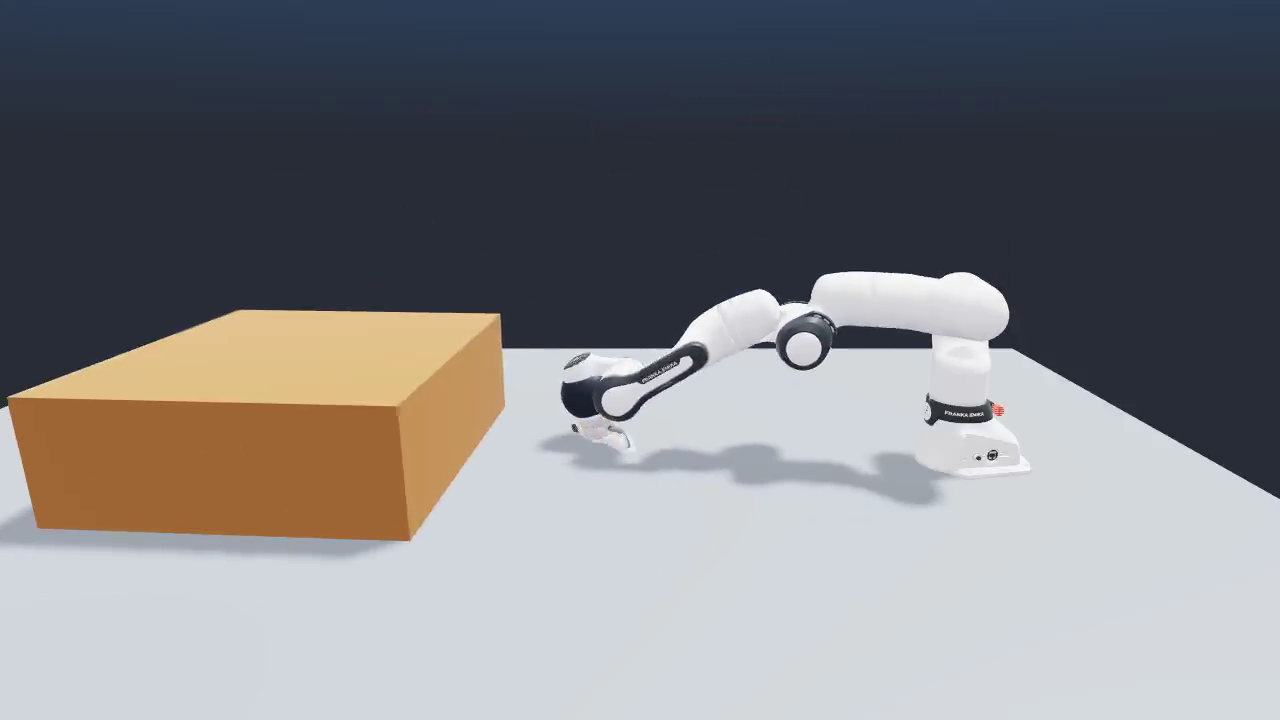}
        \includegraphics[width=0.16\textwidth]{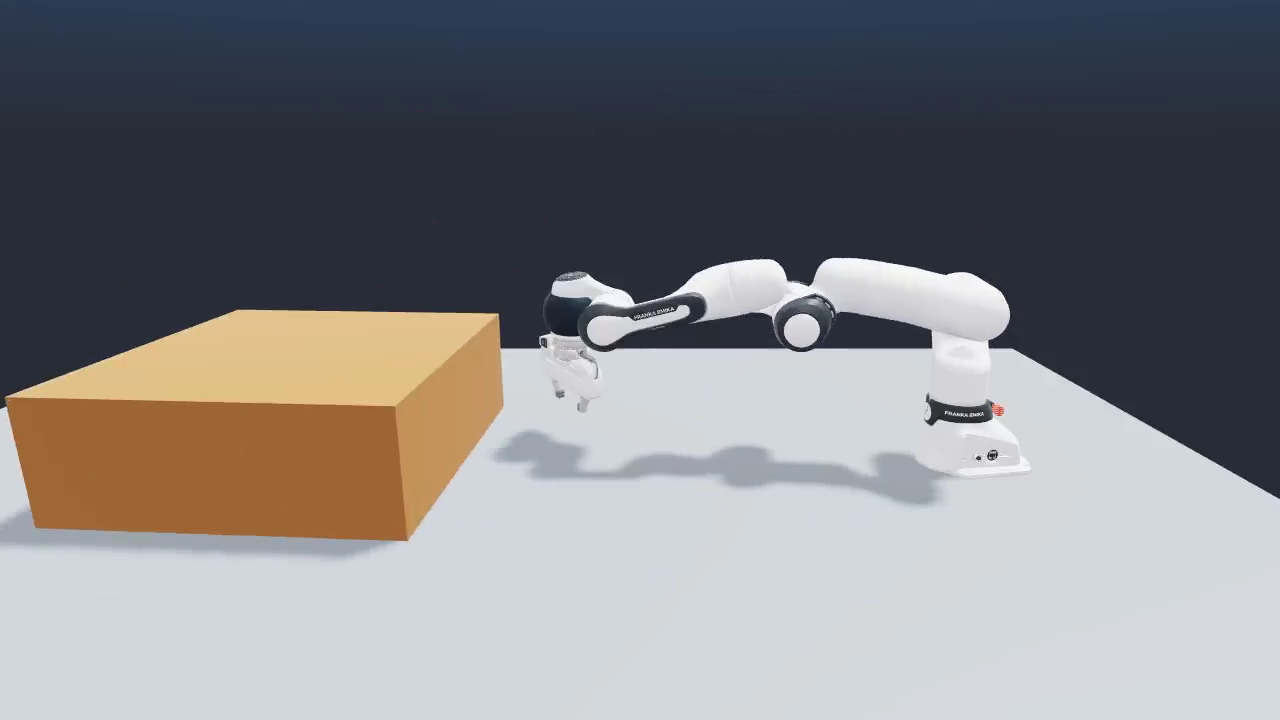}
        \caption{V-JEPA 2 trajectory rendered in the simulation with the wall treated as concrete.}
        \label{fig:vjepa_concrete}
    \end{subfigure}

    \caption{Comparison between the ground-truth physical rollout and V-JEPA 2-generated trajectories rendered in simulation. The high-push and concrete cases show that the trajectory generator can produce visually plausible motion, but does not reason about material-dependent contact dynamics or force feasibility.}
    \label{fig:vjepa_wall_comparison}
\end{figure}

\begin{figure}[t]
    \centering

    \begin{subfigure}{\textwidth}
        \centering
        \includegraphics[width=0.16\textwidth]{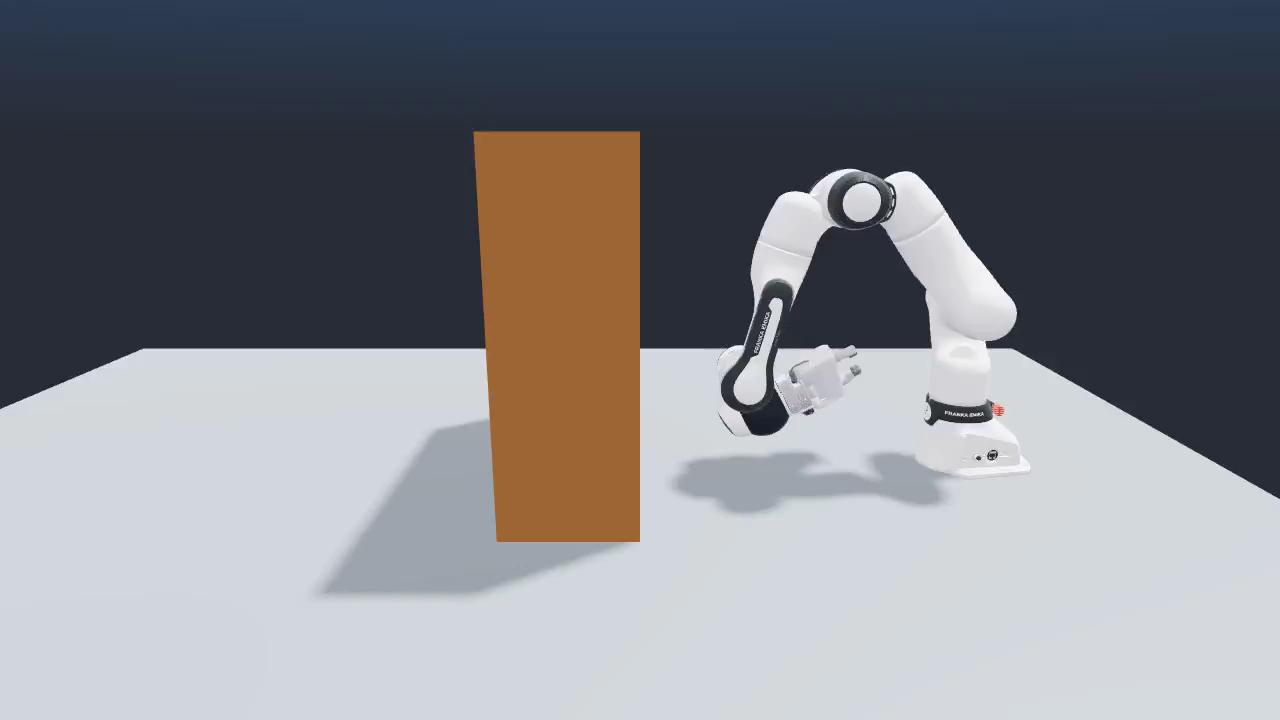}
        \includegraphics[width=0.16\textwidth]{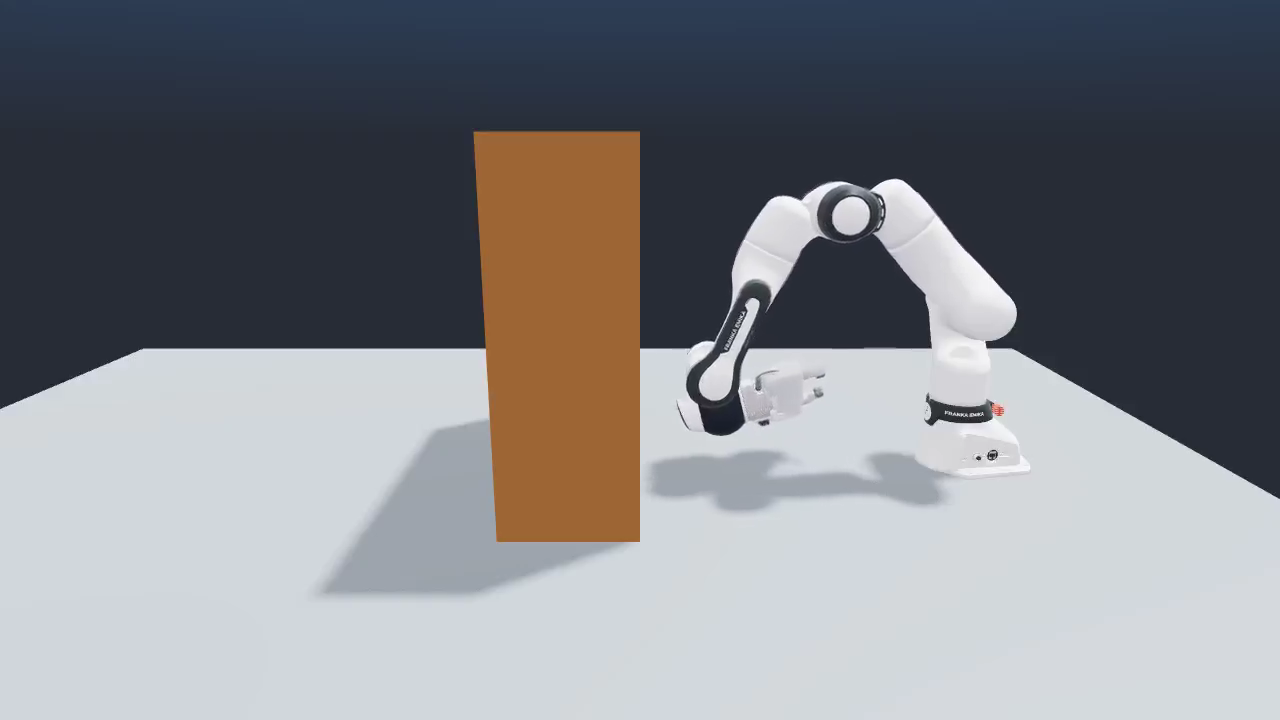}
        \includegraphics[width=0.16\textwidth]{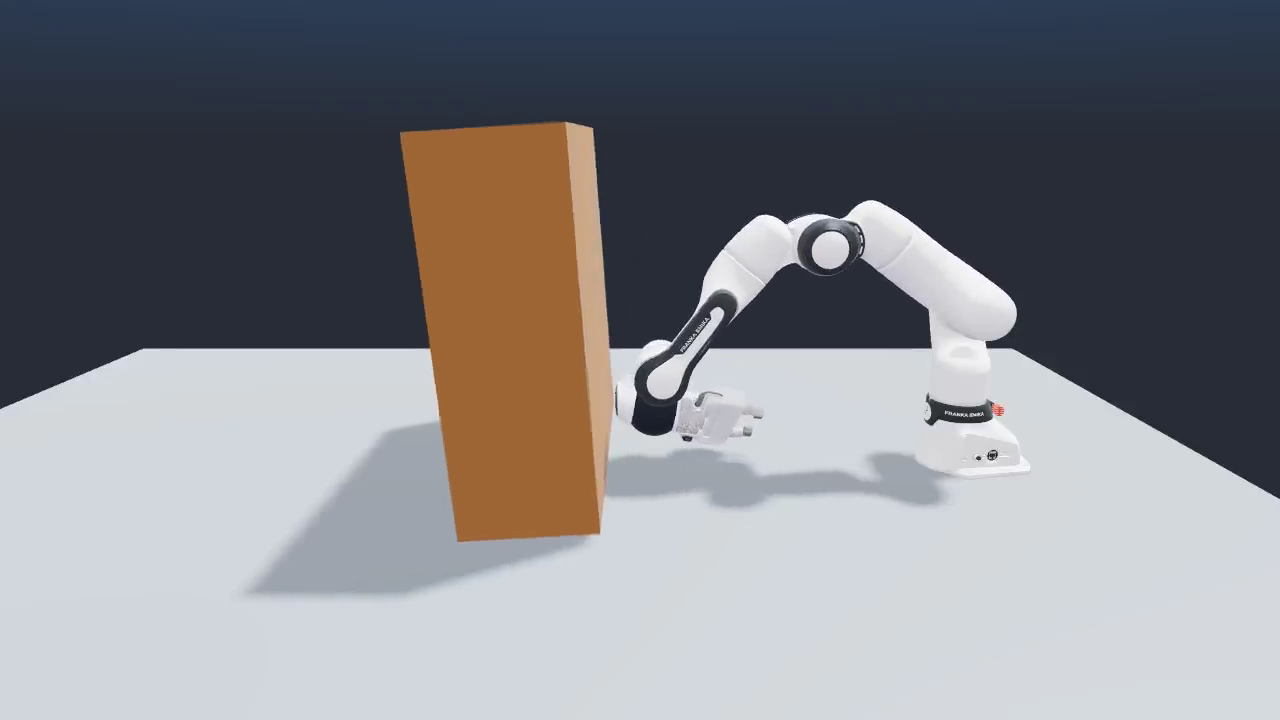}
        \includegraphics[width=0.16\textwidth]{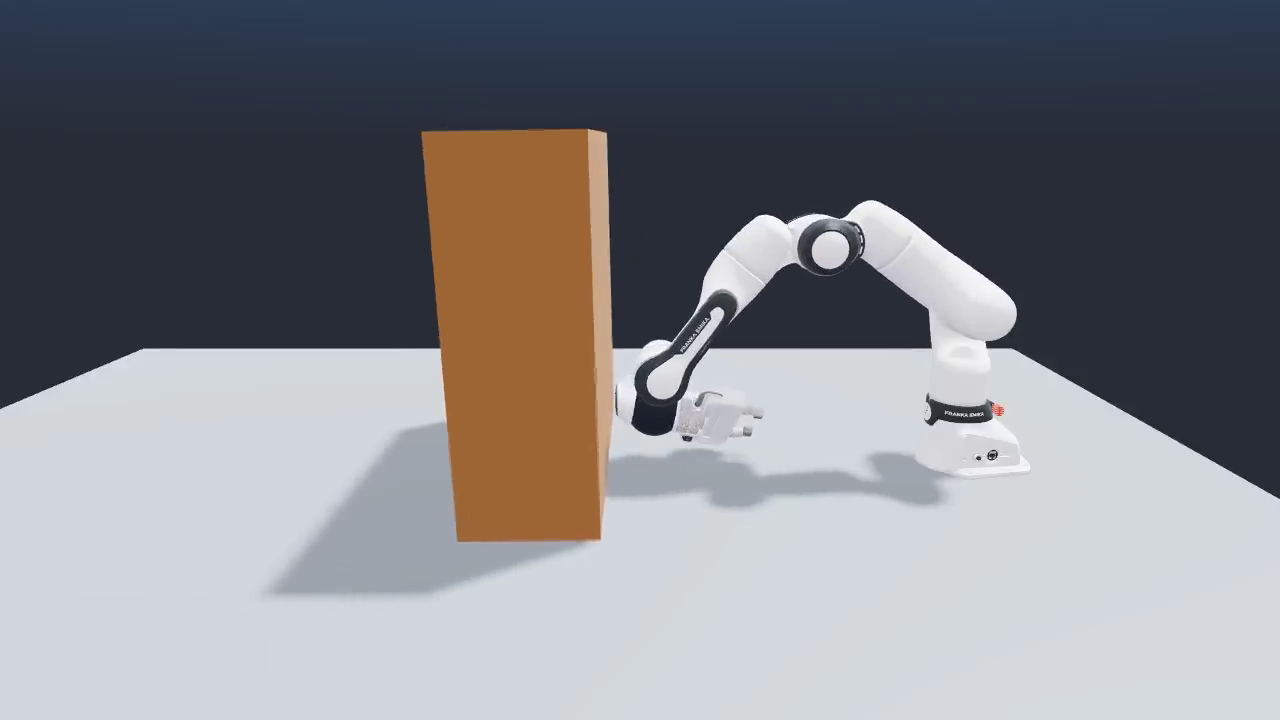}
        \includegraphics[width=0.16\textwidth]{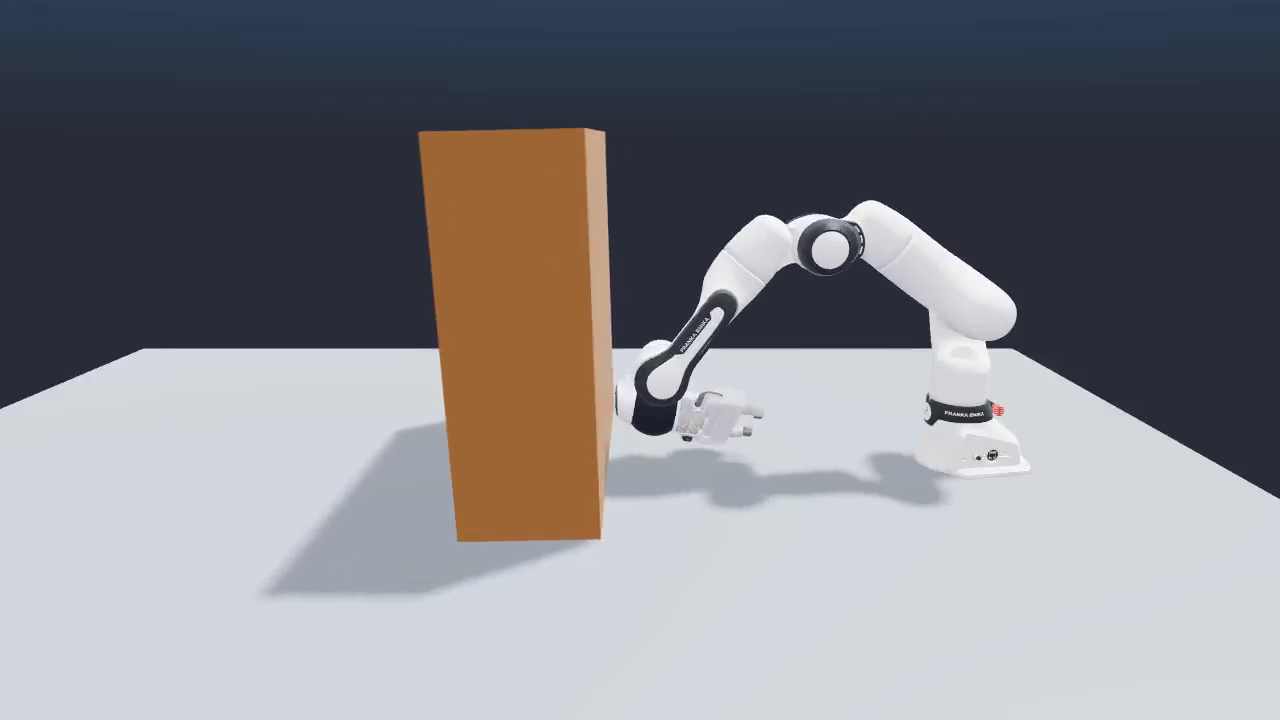}
        \includegraphics[width=0.16\textwidth]{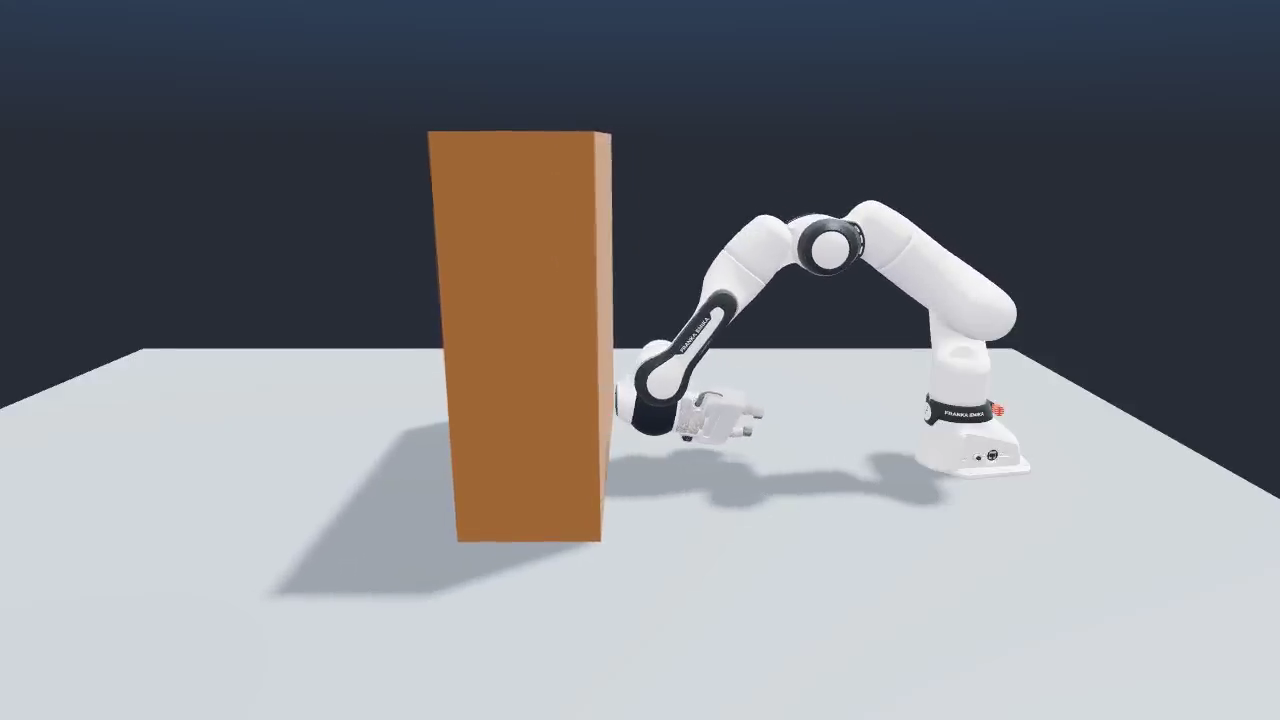}
        \caption{Ground-truth physical rollout of low pushing point.}
        \label{fig:vjepa_low_push_gt}
    \end{subfigure}

    \vspace{0.5em}

    \begin{subfigure}{\textwidth}
        \centering
        \includegraphics[width=0.16\textwidth]{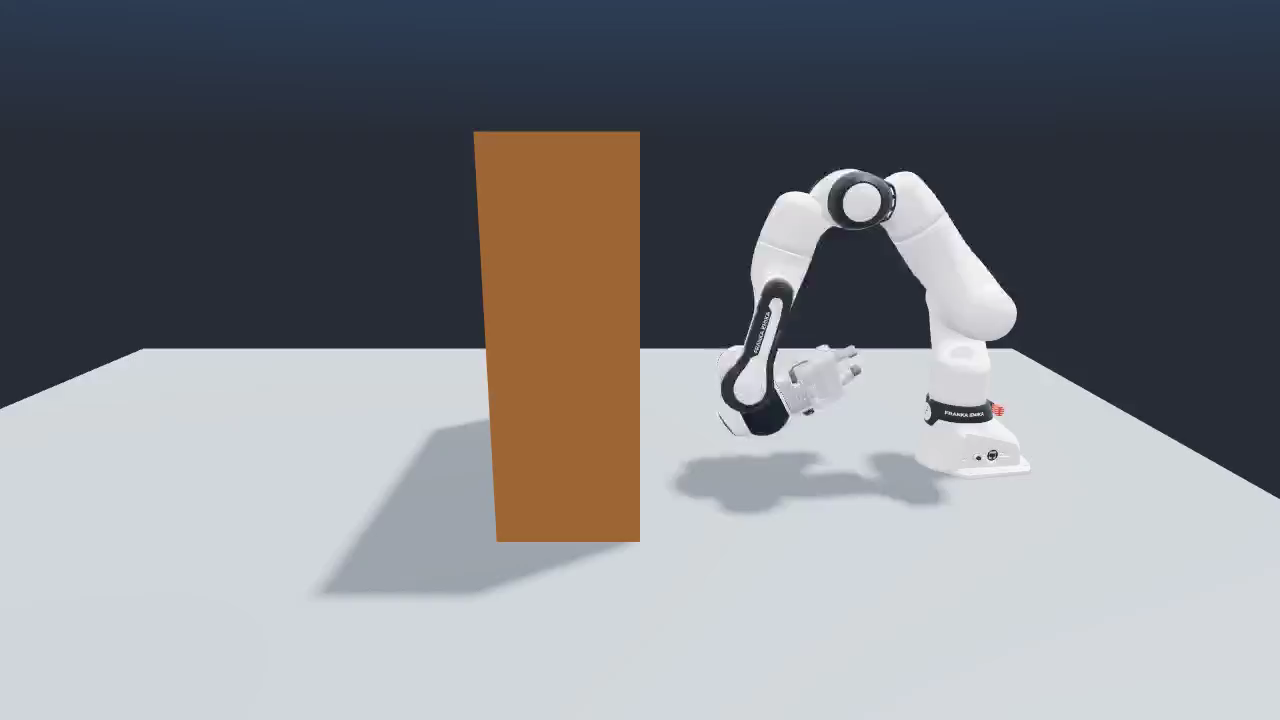}
        \includegraphics[width=0.16\textwidth]{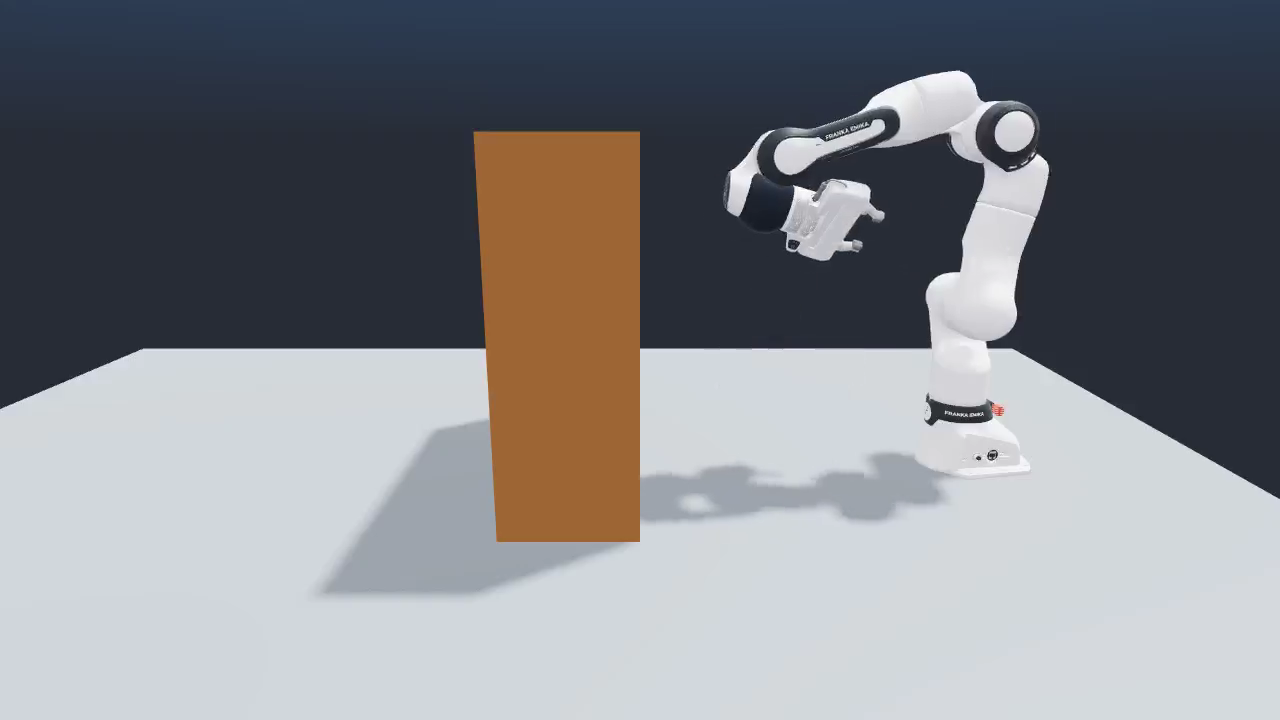}
        \includegraphics[width=0.16\textwidth]{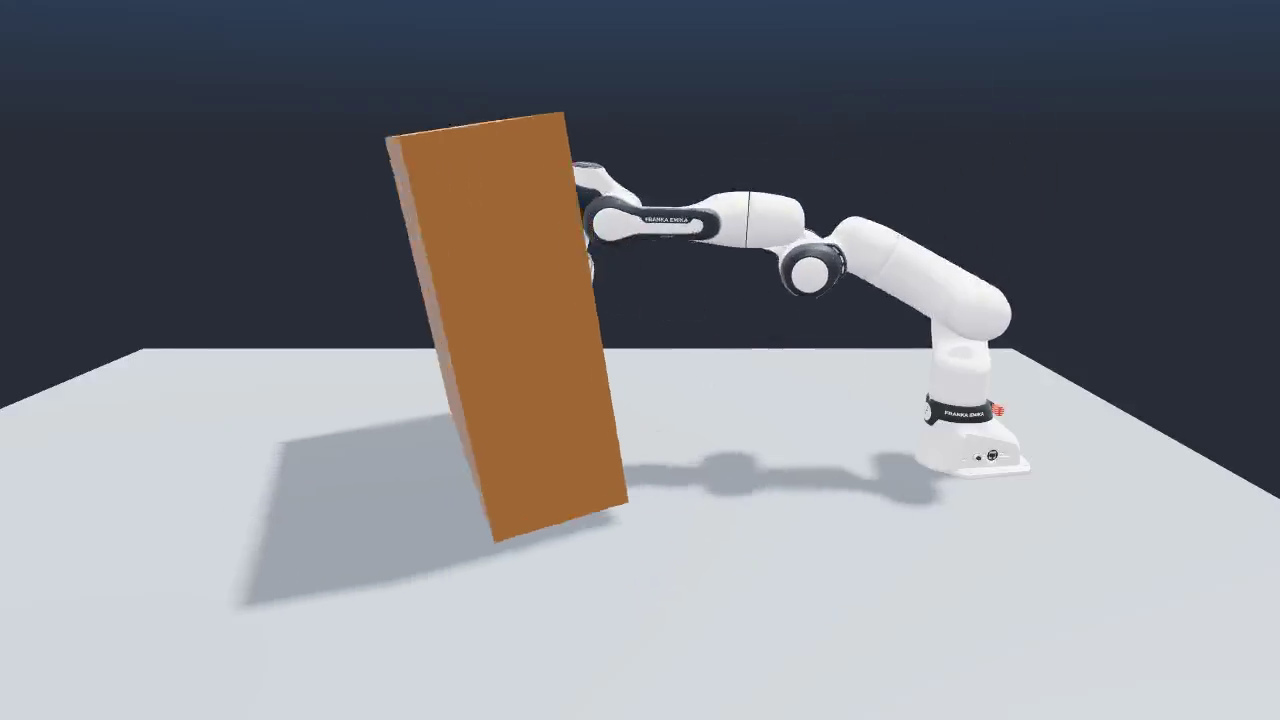}
        \includegraphics[width=0.16\textwidth]{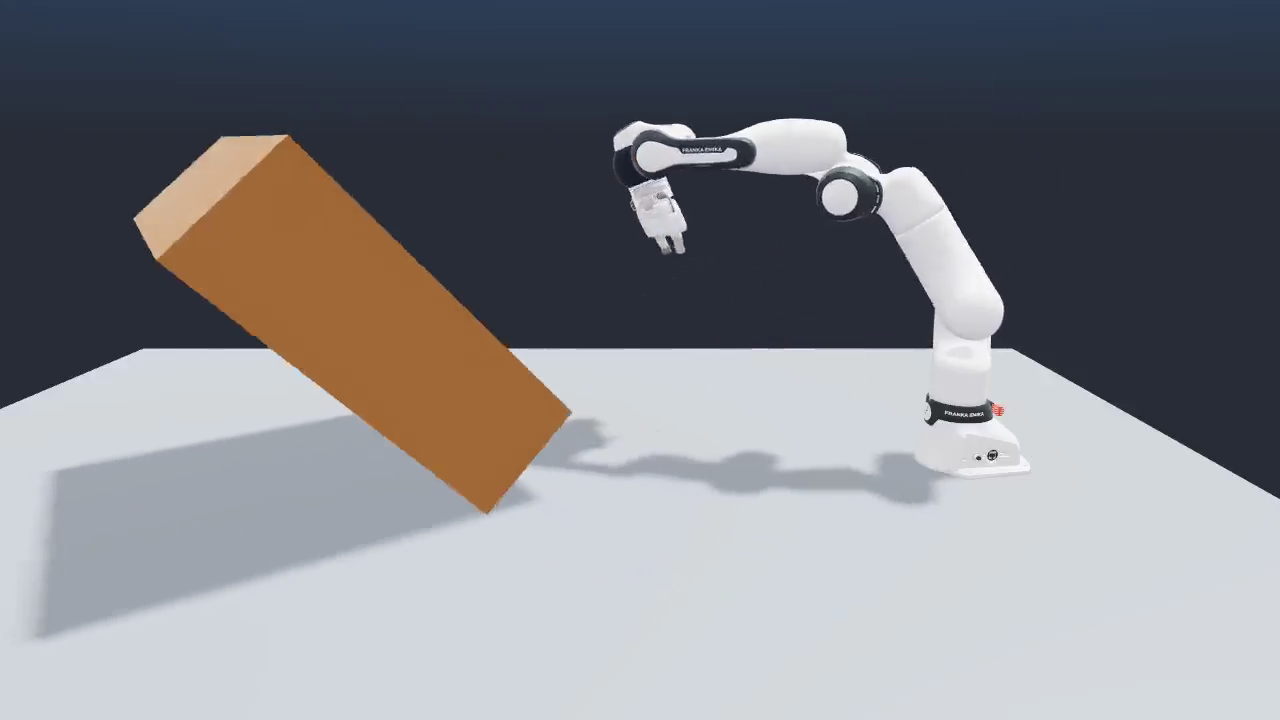}
        \includegraphics[width=0.16\textwidth]{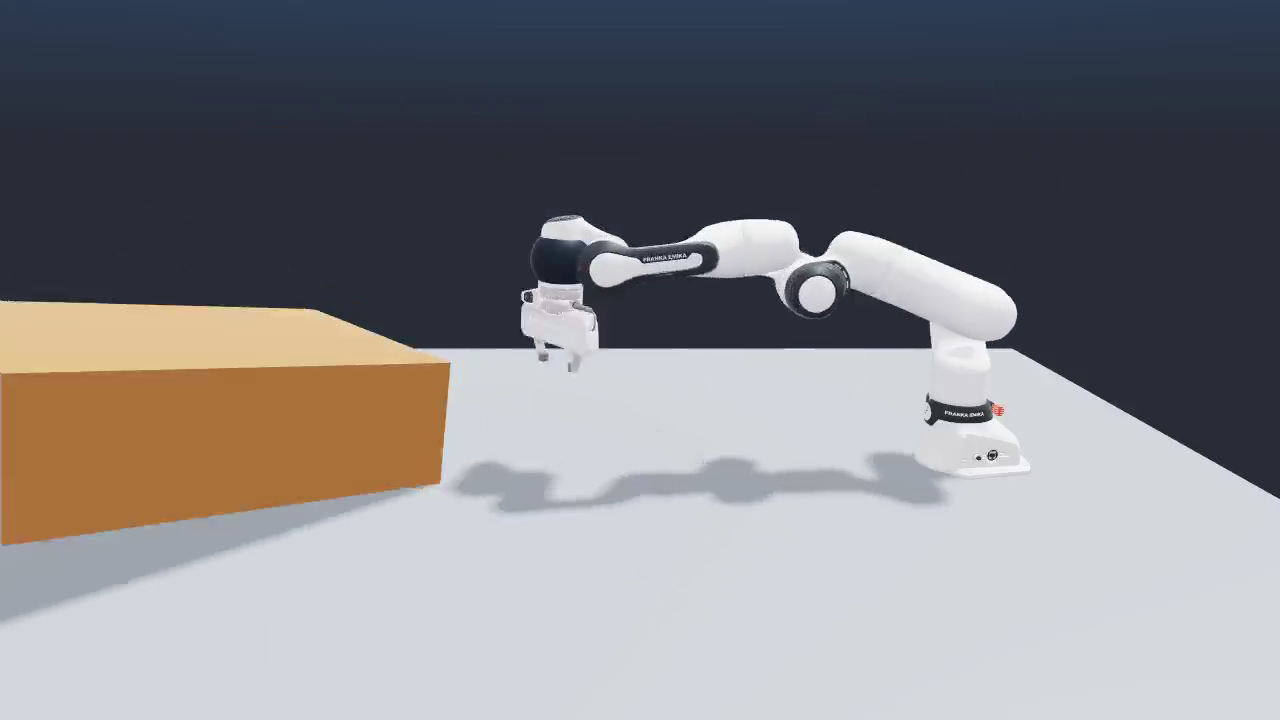}
        \includegraphics[width=0.16\textwidth]{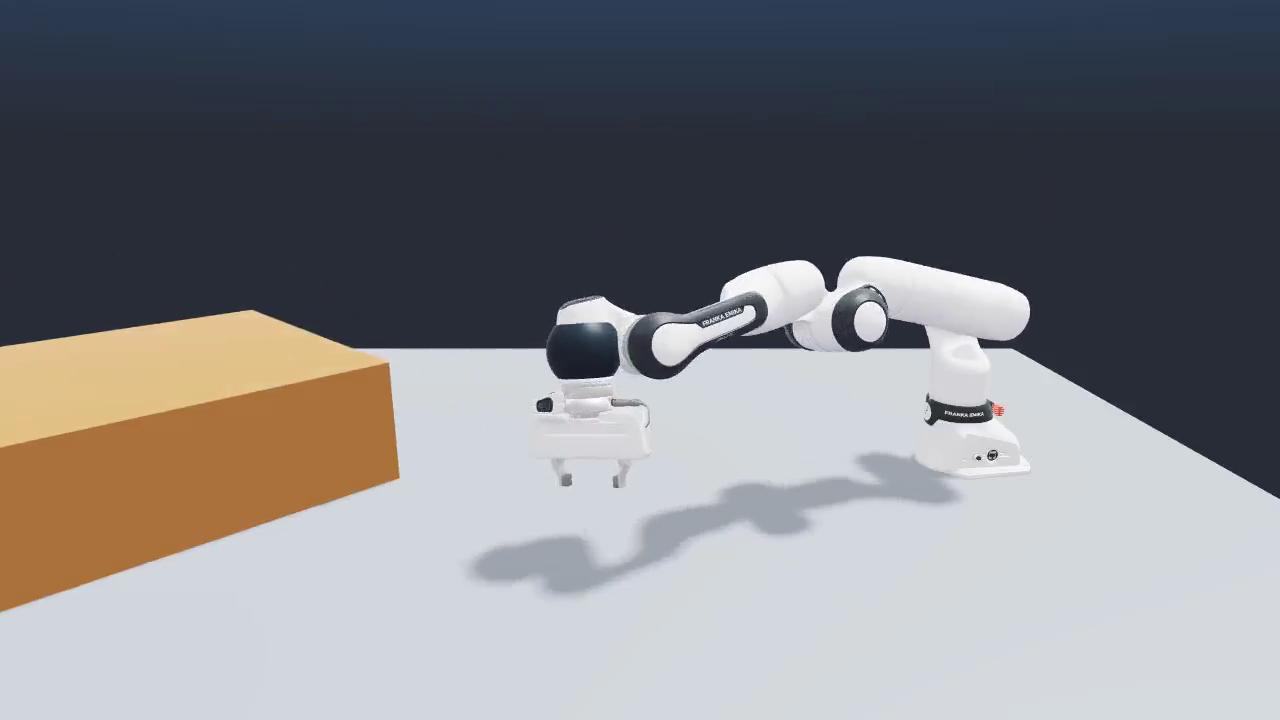}
        \caption{V-JEPA2 trajectory rendered in the simulation under the low-push setting.}
        \label{fig:vjepa_low_push}
    \end{subfigure}

    \caption{Comparison between the ground-truth physical rollout and the V-JEPA2-generated trajectory rendered in simulation under the low-push setting.}
    \label{fig:vjepa_low_push_comparison}
\end{figure}

\begin{figure}[t]
    \centering
    \includegraphics[width=\linewidth]{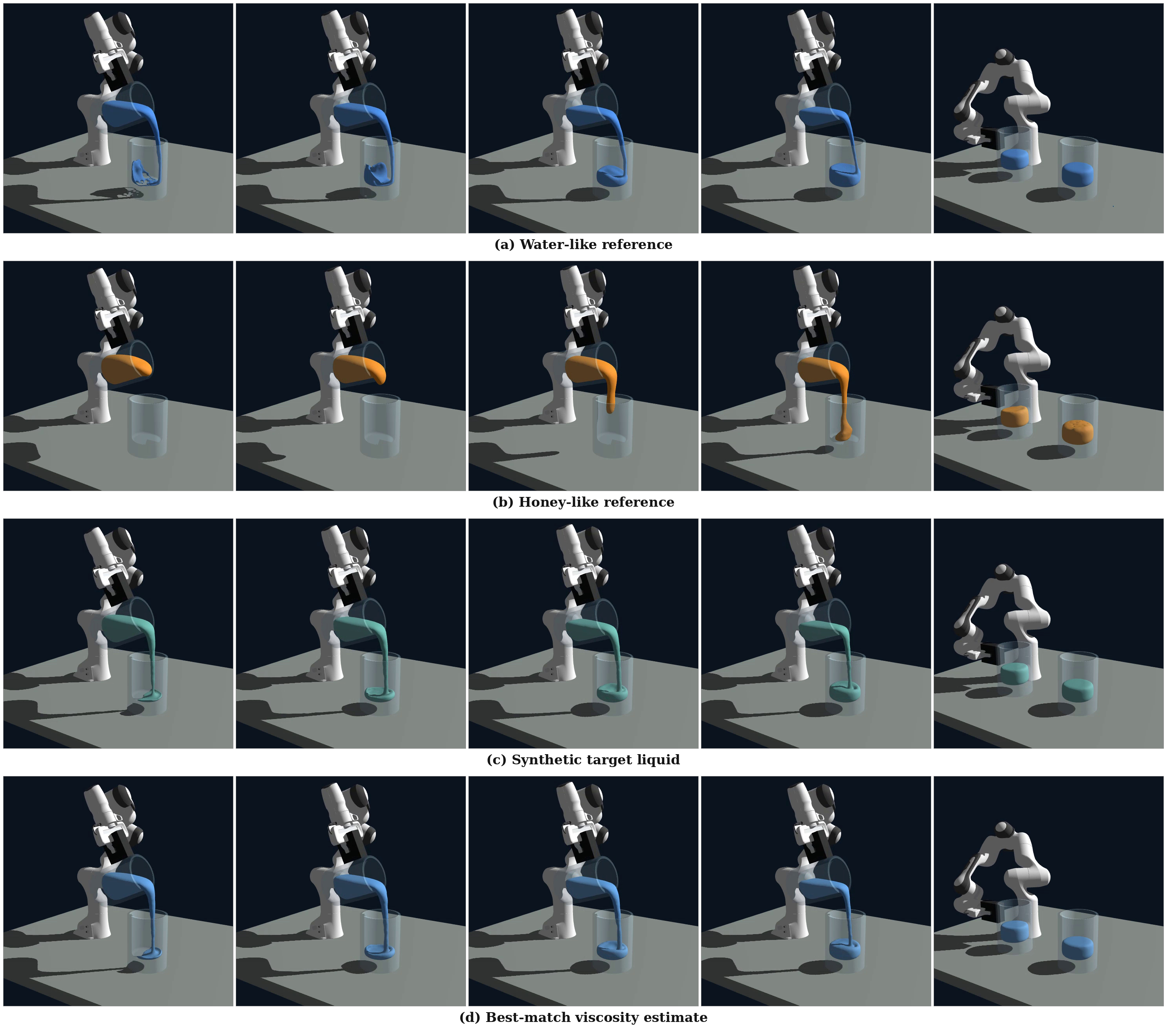}
    \caption{
    Viscosity-dependent robot-arm pouring.
    Each row shows a time sequence for a different liquid condition:
    (a) water-like reference, (b) honey-like reference, (c) synthetic target liquid, and
    (d) the best-match viscosity estimate obtained from candidate simulations.
    }
    \label{fig:appendix_pouring_viscosity_variants}
\end{figure}

We evaluate this behavior in two wall-pushing scenarios. The first varies the contact height. 
A high contact point produces a visually salient rotational response, while a low contact point requires more precise control to produce the desired interaction. 
The second varies the wall material while keeping the visual structure of the task fixed. This tests whether a vision-based model can select actions that remain valid when the same apparent goal is governed by different physical parameters.

The results are shown in~\Cref{fig:vjepa_wall_comparison} and~\Cref{fig:vjepa_low_push_comparison}. 
In the high-push case, the V-JEPA 2 planned trajectory captures the coarse behavior of the intended interaction, but it does not reliably reproduce the ground-truth physical rollout. 
This is expected from the structure of the objective: the model searches for actions that reduce latent visual discrepancy, not actions that satisfy the true contact dynamics of the wall. 
The rendered rollout therefore reveals whether the visually selected trajectory is dynamically meaningful, and in this case the correspondence is incomplete.

The material variation further exposes this limitation. 
When the wall is treated as concrete, the task changes physically even if the visual scene remains similar. 
The same end-effector displacement can lead to a different outcome because the wall's response depends on material-dependent quantities such as friction and density. 
Since the V-JEPA 2 planning objective does not explicitly infer or optimize over these physical parameters, the selected trajectory does not adapt in a reliably material-aware way. 
The model may identify a visually plausible action direction, but it has no explicit mechanism for determining whether the corresponding force interaction is feasible.
These experiments illustrate a broader limitation of vision-based world models for control. 
In contact-rich manipulation, an action is only predictive when conditioned on the relevant physical state of the scene, including geometry, contact configuration, friction, mass distribution, and material response. 
A purely visual latent model can learn correlations between observations and actions, but it does not guarantee that the learned representation preserves the physical variables required for reliable planning.

Overall, V-JEPA 2 provides a stronger baseline than direct visual generation because it produces action-conditioned latent rollouts rather than hallucinated future frames, enabling a path for downstream planning and control. 
Nevertheless, the failure mode is conceptually similar to the diffusion results: visual plausibility is not equivalent to physical validity. 
The model can optimize toward a goal representation while ignoring whether the proposed interaction is force-feasible, material-aware, or robust under execution. A glass barrier that renders a target physically inaccessible may be close to invisible visually.
With these results, we argue that vision-based predictive world models are insufficient, by themselves, for resolving the physical dynamics required by embodied robotic decision-making.
% These results suggest that vision-only training is insufficient for using foundation world models as reliable control simulators in contact-rich robotic settings.

\end{document}